\documentclass[runningheads]{llncs}
\usepackage{graphicx}
\usepackage{comment}
\usepackage{amsmath,amssymb} % define this before the line numbering.
\usepackage{color}
\usepackage{hyperref}
\usepackage{enumerate}
\usepackage{epsfig}
\usepackage{bm}
\usepackage[boxruled,linesnumbered,vlined]{algorithm2e}
\usepackage{multirow}
\usepackage{slashbox}
\usepackage{tabularx} % Automatically calculates column widths with text wrapping
\usepackage{booktabs}

\usepackage[font=small]{caption}
\usepackage{subcaption}
\let\llncssubparagraph\subparagraph
\let\subparagraph\paragraph
\usepackage[compact]{titlesec}
\let\subparagraph\llncssubparagraph

\DeclareMathOperator*{\argmax}{arg\,max}
\DeclareMathOperator*{\argmin}{arg\,min}
\DeclareMathOperator{\trace}{trace}
\DeclareMathOperator{\diag}{diag}

\newcommand{\ba}{\mathbf{a}}
\newcommand{\bb}{\mathbf{b}}
\newcommand{\bbf}{\mathbf{f}}
\newcommand{\bp}{\mathbf{p}}
\newcommand{\br}{\mathbf{r}}
\newcommand{\bs}{\mathbf{s}}
\newcommand{\bt}{\mathbf{t}}
\newcommand{\bw}{\mathbf{w}}
\newcommand{\bx}{\mathbf{x}}
\newcommand{\by}{\mathbf{y}}

\newcommand{\bphi}{\boldsymbol{\phi}}
\newcommand{\btheta}{\boldsymbol{\theta}}
\newcommand{\bA}{\mathbf{A}}
\newcommand{\bC}{\mathbf{C}}
\newcommand{\bK}{\mathbf{K}}
\newcommand{\bH}{\mathbf{H}}
\newcommand{\bR}{\mathbf{R}}
\newcommand{\bW}{\mathbf{W}}
\newcommand{\sX}{\mathcal{X}}
\newcommand{\sY}{\mathcal{Y}}
\newcommand{\bbR}{\mathbb{R}}
\newcommand{\transpose}{^\mathsf{T}}

\newcommand{\figref}[1]{Figure~\ref{#1}}
\newcommand{\eqnref}[1]{(\ref{#1})}
\newcommand{\secref}[1]{Section~\ref{#1}}
\newcommand{\tabref}[1]{Table~\ref{#1}}
\newcommand{\algoref}[1]{Algorithm~\ref{#1}}

\newcommand{\etal}{\textit{et al}.}
\newcommand{\ie}{\textit{i}.\textit{e}.}
\SetKwInput{KwData}{Inputs}
\SetKwInput{KwResult}{Output}
\SetKwComment{Comment}{$\triangleright$\ }{}

\begin{document}
% \linenumbers
\pagestyle{headings}
\mainmatter
\def\ECCVSubNumber{****}  % Insert your submission number here

% \title{A Deep Perspective-n-Point Solver Without Feature Correspondences}
\title{Learning 2D--3D Correspondences To Solve The Blind Perspective-n-Point Problem}
% \begin{comment}
% \titlerunning{ECCV-20 submission ID \ECCVSubNumber} 
% \authorrunning{ECCV-20 submission ID \ECCVSubNumber} 
% \author{Anonymous ECCV submission}
% \institute{Paper ID \ECCVSubNumber}
% \end{comment}
%******************

% \begin{comment}
\titlerunning{Deep Blind PnP}
\author{Liu Liu \inst{1,2} \and
Dylan Campbell \inst{1,2} \and
Hongdong Li \inst{1,2} \and 
Dingfu Zhou \inst{3} \and 
Xibin Song \inst{3} \and
Ruigang Yang \inst{3}
}
\authorrunning{Liu Liu et al.}
\institute{Australian National University \and
Australian Centre for Robotic Vision \and Baidu Research\\  
\email{Liu.Liu@anu.edu.au}
\\
\url{https://github.com/Liumouliu/Deep_blind_PnP}
%\and
}
% \end{comment}
%******************
\maketitle
\begin{abstract}
Conventional absolute camera pose via a Perspective-n-Point (PnP) solver often assumes that the correspondences between 2D image pixels and 3D points are given.  When the correspondences between 2D and 3D points are not known a priori, the task becomes the much more challenging {\em blind PnP} problem.  This paper proposes a deep CNN model which simultaneously solves for both the 6-DoF absolute camera pose and 2D--3D correspondences.  Our model comprises three neural modules connected in sequence.  First, a two-stream PointNet-inspired network is applied directly to both the 2D image keypoints and the 3D scene points in order to extract discriminative point-wise features harnessing both local and contextual information.
Second, a global feature matching module is employed to estimate a matchability matrix among all 2D--3D pairs. Third, the obtained matchability matrix is fed into a classification module to disambiguate inlier matches.  The entire network is trained end-to-end, followed by a robust model fitting (P3P-RANSAC) at test time only to recover the 6-DoF camera pose.
Extensive tests on both real and simulated data have shown that our method substantially outperforms existing approaches, and is capable of processing thousands of points a second with the state-of-the-art accuracy.
% \dots
\end{abstract}

\section{Introduction}

% The problem:
Solving the Perspective-n-Point (PnP) problem with unknown correspondences involves estimating the 6-DoF absolute pose (rotation $\bR$ and translation $\bt$) of the camera with respect to a reference coordinate frame, given a 2D point set from an image captured by the camera and a 3D point set of the environment in the reference frame.
Importantly, 2D--3D correspondences are unknown: any 2D point could correspond to any 3D point or to none.
This is a non-trivial chicken-and-egg problem since the estimation of correspondence and pose is coupled.
Moreover, cross-modal correspondences between image pixels and 3D points are difficult to obtain. Even if the 2D and 3D sets are known to overlap, finding the specific correspondences between 2D and 3D points is an unsolved problem.
% Even for SfM point clouds, when a new image is taken in a different season or time of day, correspondences cannot be reliably found.
%This paper solves the Perspective-n-Points (PnP) without correspondences problem, \ie, given a sparse 3D points set and a 2D points set from an image, estimate the absolute 6 DoF pose (rotation $\mathbf{R}$ and translation $\mathbf{t}$) of the image with respect to the 3D model.

\begin{figure}\centering
% \includegraphics[width=1.01\linewidth]{Figures/overall_framework2.pdf}
% \hspace*{-8.2cm}
{\normalsize\includegraphics[width=0.99\linewidth]{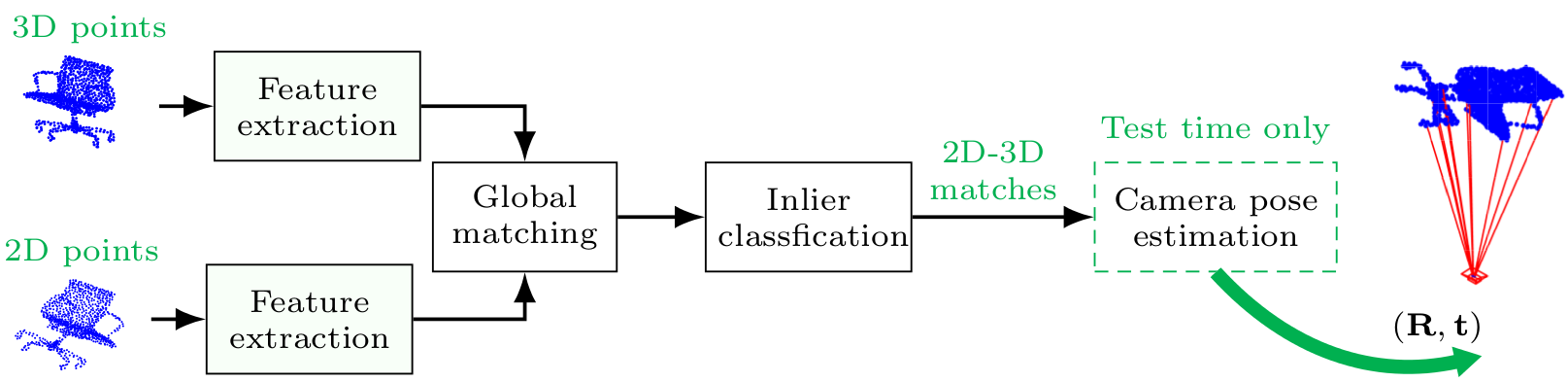}}
\caption{Overall pipeline of our method.
First, the coordinates of 2D and 3D points are passed into a two-stream network to extract point-wise deep features.
Then a global matching module estimates 2D--3D matches from these features using an optimal mass transport (OMT) technique~\cite{villani2009optimal,cuturi2013sinkhorn,courty2016optimal}. 
Finally, an inlier classification CNN is used to further separate inlier matches from those outlier matches.  At test time, apart from automatically recovering 2D--3D correspondences, the underlying 6-DoF camera pose is also recovered via standard PnP solver with RANSAC.
}
% \caption{Overall pipeline of our method.
% First, the coordinates of 2D and 3D feature points are detected in respective domains separately,  and passed into a two-stream network to extract point-wise deep features.
% Then a global matching module estimates 2D--3D matches from these features using optimal mass transport (OMT) technique \cite{villani2009optimal, cuturi2013sinkhorn, courty2016optimal}. 
% Finally, an inlier classification CNN is used to further separate inlier matches from those outlier matches.  At test time, apart from automatically recovering 2D-3D correspondences, the underlying 6-DoF camera pose is also recovered via standard PnP solver with RANSAC.
% }
\label{fig:framework}
\end{figure}
% \vspace{-10pt}

When correspondences are known, the problem reduces to the standard PnP problem \cite{grunert1841pothenotische,kneip2011novel,zheng2013revisiting,lepetit2009epnp}.
When correspondences are unknown, the problem is {\em blind PnP}, for which several traditional geometry-based methods were proposed, including SoftPoSIT \cite{david2004softposit}, BlindPnP \cite{moreno2008pose}, GOPAC \cite{campbell2017globally} and GOSMA \cite{campbell2019alignment}. 
These local methods \cite{david2004softposit,moreno2008pose} require a good pose prior to find a reasonable solution, while global methods \cite{campbell2019alignment,campbell2017globally} systematically search the space of $\bR$ and $\bt$ for a global optimum with respect to an objective function and are thus quite slow.

Instead of relying on a good prior on camera pose, or exhaustively searching over all possible camera poses, we propose to estimate the correspondence matrix directly. Once the 2D--3D correspondences have been found, the camera pose can be recovered efficiently using an off-the-shelf PnP solver inside a RANSAC \cite{fischler1981random} framework.
While a straightforward idea, finding the correspondence matrix is challenging because we need to identify inliers from a correspondence set with cardinality $M \times N$, where $M$ is the number of 3D points and $N$ is the number of 2D points.
A na\"ive RANSAC-like search of this correspondence space \cite{grimson1990object} has complexity $O(MN^3 \log N)$ \cite{david2004softposit}.
We instead estimate the correspondence matrix directly using a CNN-based method that takes only the original 2D and 3D coordinates as input.

% Motivate: Why is local information important? Patterns in 2D can be matched to patterns in 3D
The proposed method extracts discriminative feature descriptors from the point sets that encode both local geometric structure and global context at each point.
The intuition is that the local geometric structure about a point in 3D is likely to bear some resemblance to the local geometric structure of the corresponding point in 2D, modulo the effects of projection and occlusion.
We then combine the features from each point set in a novel global feature matching module to estimate the 2D--3D correspondences. This module computes a weighting (joint probability) matrix using optimal transport, where each element describes the matchability of a particular 3D point with a particular 2D point.
Sorting the 2D--3D matches in decreasing order by weight produces a prioritized match list, which can be used to recover the camera pose.
To further disambiguate inlier and outlier 2D--3D matches from the prioritized match list, we append an inlier classification CNN similar to that of Yi
\etal~\cite{yi2018learning} 
and use the filtered output to estimate the camera pose.
Our correspondence estimation CNN is trained end-to-end and the code and data will be released to facilitate future research.
The overall framework is illustrated in \figref{fig:framework}.
% We extract discriminative feature descriptor per-point purely based 3D/2D sets, and lend descriptors to estimate $\mathcal{C}$.
% In this work, 3D and 2D points are fed to a PointNet-like \cite{qi2017pointnet} CNN to extract discriminative feature descriptor per-point. Given feature descriptors, we propose a sparse feature matching method to estimate $\mathcal{C}$. Given 3D-2D matches, we use a P3P method within a RANSAC framework to estimate $\mathbf{R}$ and $\mathbf{t}$. The whole CNN is trained end-to-end by minimizing a loss function. Our overall framework is given in Figure \ref{fig:framework}.
%
Our contributions are:
% \vspace{-2pt}
\begin{enumerate}
\itemsep0em
\item a new deep method to solve the blind PnP problem with unknown 2D--3D correspondences. To the best of our knowledge, there is no existing deep method that takes unordered 2D and 3D point-sets (with unknown correspondences) as inputs, and outputs a 6-DoF camera pose;
\item a two-stream network to extract discriminative features from the point sets, which encodes both local geometric structure and global context; and
\item an original global feature matching network based on a recurrent Sinkhorn layer to find 2D--3D correspondences, with a loss function that maximizes the matching probabilities of inlier matches.
% \item state-of-the-art performance, orders of magnitude faster ($>100\times$) than existing blind PnP approaches. 
%Experiments show that the proposed method outperforms existing traditional blind PnP approaches: processing thousands of 3D and 2D points with state-of-the-art pose accuracy in one second.  Our code and data will be released to facilitate future research.
\end{enumerate}
Our method achieves state-of-the-art performance, orders of magnitude faster ($>100\times$) than existing blind PnP approaches.

% \vspace{-20pt}

%-------------------------------------------------------------------------
\section{Related Work}
When 2D--3D correspondences are known, PnP methods \cite{grunert1841pothenotische,kneip2011novel,zheng2013revisiting,lepetit2009epnp} can be used to solve the 6-DoF pose estimation problem. When correspondences are unknown, the local optimization method SoftPOSIT \cite{david2004softposit} iterates between solving for correspondences and solving for pose. The correspondences are estimated from a zero--one assignment matrix using Sinkhorn's algorithm \cite{sinkhorn1964relationship}.
This method requires a good pose prior and can only find a locally-optimal pose within the convergence basin of the prior.
BlindPnP \cite{moreno2008pose} also relies on good pose priors to restrict the number of possible 2D--3D matches.
To avoid getting trapped in local optima, the global methods GOPAC \cite{campbell2017globally} and GOSMA \cite{campbell2019alignment} were proposed. Though guaranteed to find the optimum, they can only handle a moderate number of points (often hundreds) since the time-consuming branch-and-bound \cite{landautomatic} algorithm is used. Furthermore, they are affected by geometric ambiguities, meaning that many different camera poses can be considered equivalently good.
%Though effective, global methods aim to search the optimal $\mathbf{R}$ and $\mathbf{t}$ with an objective (minimizing $L_2$ distance between mixture distributions \cite{campbell2019alignment}, maximizing the cardinality of a inlier set \cite{campbell2017globally})

When 3D points are not utilized, the PoseNet algorithms \cite{kendall2015posenet,kendall2017geometric} can directly regress a camera pose. However, the accuracy of the regressed 6-DoF poses is inferior to geometry-based methods that use 3D points.

With PointNet \cite{qi2017pointnet}, deep networks can now handle sparse and unordered point-sets. Though most PointNet-based works focus on classification and segmentation tasks \cite{wang2018dynamic,qi2017pointnet}, traditional geometry problems are ready to be addressed. For example, 3D--3D registration \cite{aoki2019pointnetlk}, 2D--2D outlier correspondence rejection \cite{yi2018learning}, and 2D--3D outlier correspondence rejection \cite{dang2018eigendecomposition}.
In contrast, this paper tackles the problem of PnP with unknown correspondences using a deep network, which has not previously been demonstrated. 
% The key challenge is encoding sufficient information in the features extracted by the PointNet-like network to ensure that 2D--3D feature matching is possible from the geometry of the points alone.
The key challenge is encoding sufficient information in point-wise features from the geometry of points alone, and matching these 2D and 3D features effectively.

% \section{Method}
\section{Learning Correspondences for Blind PnP}

\subsection{Problem Formulation}

% Notation and assumptions:
Let $\sX = \{\bx_1,\dots,\bx_M\}$ denote a 3D point set with $M$ points $\bx_i \in \bbR^3$ in the reference coordinate system, $\sY = \{\by_1,\dots,\by_N\}$ denote a 2D point set with $N$ points $\by_j \in \bbR^2$ in an image coordinate system,
and $\bC \in \bbR^{M \times N}$ denote the correspondence matrix between $\sX$ and $\sY$.
We assume that the camera is calibrated, and thus the intrinsic camera matrix $\bK$ \cite{hartley2003multiple} is known.

% Problem Definition:
Blind PnP aims to estimate a rotation matrix $\bR$ and a translation vector $\bt$ which transforms 3D points $\sX$ to align with 2D points $\sY$. Specifically,
%for corresponding points, the angle
%the angle between the transformed 3D point and the 2D bearing vector should be near zero for corresponding points 
$\angle\left(\bR\bx_i + \bt, \bK^{-1} \hat{\by}_j\right) \approx 0$ for $\bC_{ij} = 1$, where $\hat{\by} = (\by, 1)$ is the homogeneous coordinate of $\by$.
The difficult part of this problem is to estimate the correspondence matrix $\bC$. Once it is found, a traditional PnP algorithm can solve the problem.

% Approach:
We propose to estimate 
% the correspondence matrix 
$\bC$ using a deep neural network. Specifically, for each tentative 2D--3D match in $\bC$, we calculate a weight $\bW_{ij}$ for $i \in [1, M]$ and $j \in [1, N]$ describing the matchability of $\bx_i$ and $\by_j$. We can obtain a set of 2D--3D matches by taking the Top-$K$ matches according to these weights.

% Our method is outlined as follows:
% in the remainder of this section. 
We present our method for extracting point-wise discriminative features in \secref{sec:feature_extraction}. We then describe our global feature matching method for obtaining 2D--3D match probabilities in \secref{sec:feature_matching}. Finally, we provide our match refinement strategy using a classification CNN to further disambiguate inlier and outlier matches in  \secref{matches_refinement}.

% The overall framework of our method is given in Figure \ref{fig:framework}.

% In geometry part, we aim to minimize the following objective with regrad to the unknown $\mathbf{W}, \mathbf{R}, \mathbf{t}$:
% \begin{equation}\label{eq::weightReErr}
%     L = \sum_{i}^{M}\sum_{j}^{N}w_{i,j}\left \| \mathbf{\tilde{y}}_j- \mathcal{P} \left (\mathbf{K}[\mathbf{R}|\mathbf{t}]\mathbf{x}_i] \right) \right \|^2
% \end{equation}
% where $\mathcal{P}(\cdot)$ denotes project the 3-dim homogeneous coordinates to 2-dim inhomogeneous coordinates, and $\mathbf{\tilde{y}}_j$ denote the original position of 2D feature point in image coordinate system.
% The optimal solution is achieved when $w_{i,j} = 1$ for positive (inlier) 3D-2D matches and $w_{i,j} = 0$ otherwise. That is to say, the weighting matrix $\mathbf{W}$ should be as sparse as possible. This goal is achieved by PointNet-like descriptor learning CNN, as we aim to learn discriminative feature descriptor per 3D/2D point. The weighting matrix $\mathbf{W}$ is explicitly estimated based on the $L_2$ distances between feature descriptors.

% \vspace{-15pt}
\begin{figure}[!t]\centering
% \includegraphics[width=0.9\linewidth]{framework.pdf}
% \includegraphics[width=0.9\linewidth]{Figures/feature_extract.pdf}
% \hspace*{-3.2cm}
\resizebox{0.99\textwidth}{!}{%
\includegraphics[]{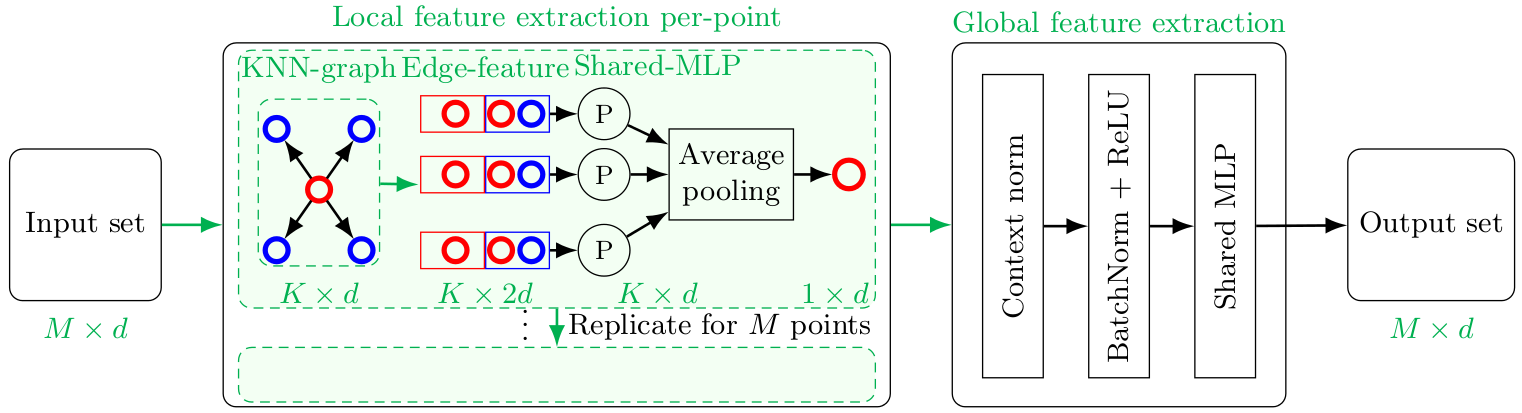}
}%
% \normalsize\includegraphics[]{Figures/pipeline.tikz}
\caption{Our feature extraction pipeline. Given an input set with size $M \times d$, we first perform nearest neighbor search and build a point-wise KNN graph. For each point (anchor), we extract $K \times d$ anchor--neighbor edge features, and concatenate these with the anchor features. This is then passed through a shared MLP block and an average pooling block to aggregate local features. Hence, this feature vector encodes local geometric information from the point-wise KNN graph.
Next, the local features at each point are passed to a context normalization module to encode global contextual information, followed by batch normalization, a ReLU non-linearity, and a shared MLP to output the final point-wise features.}
\label{fig:feature_extraction}
\end{figure}

% \vspace{9pt}
\subsection{Feature Extraction}
\label{sec:feature_extraction}

To learn discriminative features $\bbf_{\bx_i}$ and $\bbf_{\by_j}$ for each 3D point $\bx_i$ and 2D point $\by_j$ respectively, we propose a two-stream network. One branch takes 3D points from $\sX$ as inputs and the other takes 2D points from $\sY$. The two branches do not share weights. Both aim to encode information about the local geometric structure at each point as well as global contextual information. The detailed structure for a single branch is given in \figref{fig:feature_extraction}.

\vspace{3pt}
\noindent\emph{Pre-processing:}
% \paragraph{Pre-processing:}
The 2D points are transformed to normalized coordinates using the camera intrinsic matrix $\bK$ to improve numerical stability \cite{hartley2003multiple}.
% transformed to homogeneous coordinates, the inverse camera matrix is applied, and then transformed back
% $\mathbf{K}^{-1} \tilde{\mathbf{y}}$ where $\tilde{\mathbf{y}} = (\mathbf{y}, 1)$
The 3D points are aligned to a canonical direction, similarly to PointNet \cite{qi2017pointnet}, which is beneficial for extracting features. Specifically, a $3 \times 3$ transformation matrix is learned and applied to the original coordinates of 3D points.

\vspace{3pt}
\noindent\emph{Encoding Local Geometry:}
% \paragraph{Encoding Local Geometry:}
To extract point-wise local features, we first perform an $L_2$ nearest neighbor search and build a point-wise KNN graph. For a KNN graph around the anchor point indexed by $q$, the edges from the anchor to its neighbors capture the local geometric structure. Similar to EdgeConv \cite{wang2018dynamic}, we concatenate the anchor and edge features, and then pass them through an MLP module to extract local features around the $q$-th point. Specifically, the operation is defined by
\begin{equation}
    E\left(\mathbf{o}_{q}\right) = \text{avg}_{k^*, k^*\in\mho(q)} \left( \btheta \left( \mathbf{o}_{k^*}-\mathbf{o}_q \right) + \bphi \mathbf{o}_q \right),
\end{equation}
where $k^*\in\mho(q)$ denotes that point $k^*$ is in the neighborhood $\mho(q)$ of point $q$, $\btheta$ and $\bphi$ are MLP weights performed on the edge $\left ( \mathbf{o}_{k^*}-\mathbf{o}_q \right )$ and anchor point $\mathbf{o}_q$ respectively, and $\text{avg()}$ denotes that we perform average pooling in the neighborhood $\mho(q)$ after the MLP to extract a single feature for point $q$. We detail the above operations in \figref{fig:feature_extraction}. Note that for the first layer of our CNN, the MLP module lifts the dimensions of 3D and 2D points to $d=128$.

\vspace{3pt}
\noindent\emph{Encoding Global Context:}
% \paragraph{Encoding Global Context:}
After extracting point-wise local features, we aim to also embed global contextual information within them.
We use Context Normalization \cite{yi2018learning} to share global information while remaining permutation invariant. This layer normalizes the feature distribution across the point set, applying the non-parametric operation $CN(\mathbf{o}_q) = (\mathbf{o}_q-\bm{\mu})/\bm{\sigma}$, where
$\mathbf{o}_q$ is the $q$-th feature descriptor, and $\bm{\mu}$ and $\bm{\sigma}$ are the mean and standard deviation across the point set.
%Since the order of local features is arbitrary, our method to encode global context should take individual feature as input and be invariant to the order. Inspired by \cite{yi2018learning}, we exploit Context Normalization (CN) to globally normalize feature distributions.
Context normalized features are then passed through batch normalization, ReLU, and shared MLP layers to output the final point-wise features. 

We replicate the local and global feature extraction modules $F$ times with residual connections \cite{he2016deep} to extract deep features. Finally, we $L_2$ normalize all feature vectors 
to embed them to a metric space.

\subsection{Global Feature Matching}
\label{sec:feature_matching}

Given a learned feature descriptor per point in $\sX$ and $\sY$, we perform global feature matching to estimate the likelihood that a given 2D--3D pair matches. 
To do so, we compute the pairwise distance matrix $\bH \in \bbR_{+}^{M\times N}$, which measures the cost of assigning 3D points to 2D points.
Each element of $\bH$ is the $L_2$ distance between the features at point $\bx_i$ and $\by_j$, \ie, $\bH_{ij} = \| \bbf_{\bx_i} - \bbf_{\by_j} \|_2$.
Furthermore, to model the likelihood that a given point has a match and is not an outlier, we define unary matchability vectors, denoted by $\br$ and $\bs$ for the 3D and 2D set respectively.

\begin{figure}
\centering
% \includegraphics[width=0.99\linewidth]{Figures/feature_match.pdf}
% \caption{
% Our feature matching pipeline. Given an $M \times d$ feature set from the 3D data and an $N \times d$ feature set from the 2D data, we compute the pairwise $L_2$ distance matrix $\bH$. This is transformed to a joint probability matrix $\bW$ using Sinkhorn's algorithm. Reshaping $\bW$ and sorting the 3D--2D matches by their corresponding matching probabilities generates a prioritized 3D--2D match list. We take the top $K$ matches as our set of putative correspondences.
% }
% \includegraphics[width=0.99\linewidth]{Figures/feature_match2.pdf}

\includegraphics[width=0.99\linewidth]{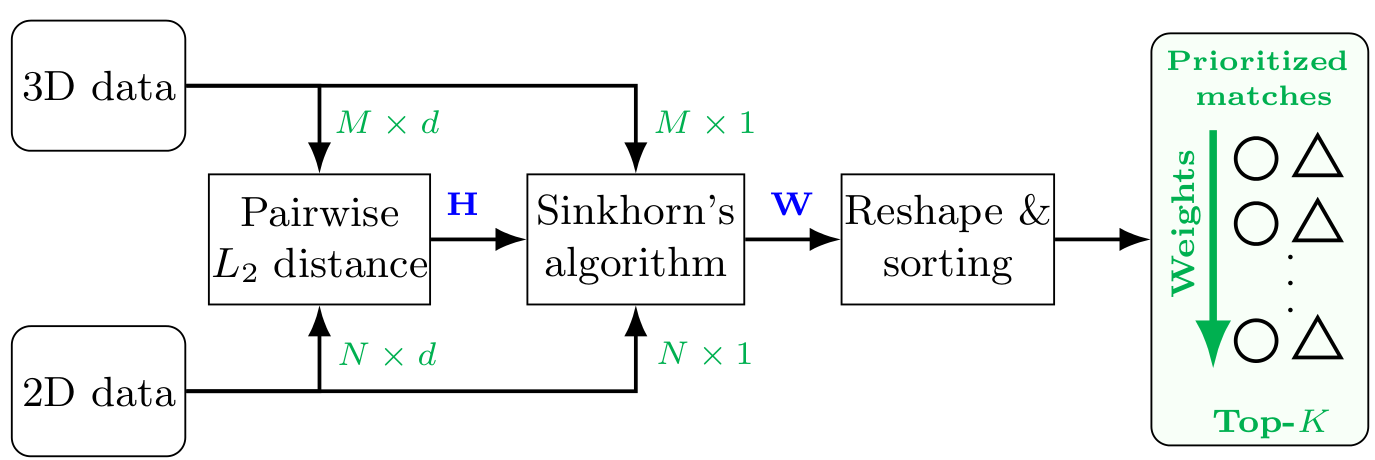}

\caption{
Our feature matching pipeline. Given an $M \times d$ feature set from the 3D data and an $N \times d$ feature set from the 2D data, we compute the pairwise $L_2$ distance matrix $\bH$. Along with a unary matchability $M$-vector from the 3D data and $N$-vector from the 2D data, the distance matrix $\bH$ is transformed to a joint probability matrix $\bW$ using Sinkhorn's algorithm. Reshaping $\bW$ and sorting the 2D--3D matches by their corresponding matching probabilities generates a prioritized 2D--3D match list. We take the Top-$K$ matches as our set of putative correspondences.
}
\label{fig:feature_matching}
\end{figure}

From $\bH$, $\br$ and $\bs$ we estimate a weighting matrix $\bW \in \bbR_{+}^{M\times N}$ where each element $\bW_{ij}$ represents the matchability of the 3D--2D pair $\{\bx_i, \by_j\}$. Note that each element $\bW_{ij}$ is estimated from the cost matrix $\bH$ and the unary matchability vectors $\br$ and $\bs$,  rather than locally from $\bH_{ij}$. In other words, the weighting matrix $\bW$ globally handles pairwise descriptor distance ambiguities in $\bH$, while respecting the unary priors. The overall pipeline is given in Figure \ref{fig:feature_matching}.
% \vspace{-30pt}

% \paragraph{Estimating the Weighting Matrix:}
% Inspired by the optimal transport theory \cite{villani2009optimal,cuturi2013sinkhorn,courty2016optimal}, we introduce a simple iterative solver to estimate $\mathbf{W}$.
% This let us globally handle descriptor matching ambiguity by taking account of pairwise descriptor distances. 
% Recall that $\bW_{ij}$ encodes the pairwise matchability of the 3D--2D pair $\{\bx_i, \by_j\}$, it should also respect unary match-abilities of 3D point $\bx_i$ and 2D point $\by_j$.

\vspace{3pt}
\noindent\emph{Prior Matchability:}
% \paragraph{Prior Matchability:}
For each point we define a prior unary matchability measuring how likely it is to have a match. Formally, let $\br_i$ and $\bs_j$ denote the unary matchabilities of points $\bx_i$ and $\by_j$ respectively. Collecting the matchabilities for all 2D or 3D points yields a matchability histogram, a 1D probability distribution, given by $\br \in \Sigma_M$ and $\bs \in \Sigma_N$, where a simplex in $\bbR^M$ is defined as
$\Sigma_{M} = \left\{ \br \in \bbR_+^M , \sum_i \br_i = 1 \right\}$.
% \begin{equation}
%     \Sigma_{M} = \left\{ \mathbf{r} \in \mathbb{R}_+^M , \sum_i r_i = 1 \right\}.
% \end{equation}

We make the assumption that the unary matchabilities are uniformly distributed, that is, $\br_i = 1/M, \bs_j = 1/N$. This means that each point has the same prior likelihood of matching.
While our model can predict non-uniform priors, we found that using learned priors led to overfitting.
%Although we can predict unary match-abilities for 3D/2D points by looking at CNN learned descriptors, it requires learned descriptors cover the descriptor space well. We found that using learned probabilities leads to overfitting in our experiments, thus we select to use fixed uniform probabilities. 
% \cite{} propose a hinge ranking loss to self-supervise the training of match-ability prediction, however, it is very complicate and introduce an auxiliary task. In our work, we can also use MLPs to predict match-abilities, however, it is left as future work.
% \vspace{-20pt}
\begin{algorithm}
\SetAlgoLined
\DontPrintSemicolon
\KwData{$\bH$, $\br$, $\bs$, $\bb = \textbf{1}^N$, $\lambda$, and iteration number $\textit{Iter}$}
\KwResult{Weighting matrix $\bW$}
% \textbf{Input:} $\bH$, $\br$, $\bs$, $\bb = \textbf{1}^N$, $\lambda$, and iterations $\textit{Iter}$\\
$\bm{\Upsilon} = \exp\left (-\textbf{H}/\lambda  \right ) $ \\
$\bm{\Upsilon} = \bm{\Upsilon} / \sum\bm{\Upsilon}$
\tcp*{normalize $\bm{\Upsilon}$ to be a joint probability matrix}
\While{$\text{it} < \text{Iter}$}{ 
  $\ba = \br \oslash (\bm{\Upsilon}\bb) $
%   \ \ \ \ \ \ \ \  \Comment{alternatively updating $\ba$ and $\bb$ }
\tcp*{alternatively updating $\ba$ and $\bb$}
  $\bb = \bs \oslash (\bm{\Upsilon}\transpose\ba) $
 }
$\bW = \diag(\ba)\bm{\Upsilon}\diag(\bb)$
\tcp*{assemble to build the weighting matrix $\bW$}
\caption{Sinkhorn's Algorithm to solve \eqref{eq::OT}. Hadamard (elementwise) division is denoted by $\oslash$. }\label{alg:sinkhorn}
\end{algorithm}
% \vspace{-20pt}

\vspace{3pt}
\noindent\emph{Solving for $\bW$:}
% \paragraph{Solving for $\bW$:}
From optimal transport theory \cite{villani2009optimal,cuturi2013sinkhorn,courty2016optimal}, the joint probability matrix $\bW$ can be obtained by solving
\begin{equation}\label{eq::OT}
  \argmin_{\bW\in\Pi\left(\br,\bs \right)} \left \langle \bH,\bW \right \rangle - \lambda E\left( \bW \right),
\end{equation}
where $\left\langle\cdot,\cdot\right\rangle$ is the Frobenius dot product and $\Pi(\br,\bs)$ is the transport polytope that couples two unary matchability vectors $\br$ and $\bs$, given by
\begin{equation}\label{eq::Pi}
\Pi\left(\br,\bs \right) = \left \{\bW \in  \bbR_+^{M\times N}, \bW\mathbf{1}^N = \br, \bW\transpose\mathbf{1}^M = \bs \right \},
\end{equation}
where $\mathbf{1}^N = [1,1,...,1]\transpose\in \bbR^N$.
The constraint on $\bW$ ensures that we assign the binary matchabilities of each 3D point to all 2D points without altering the unary matchability of the point.
The entropy regularization term $E\left(\bW \right)$
%encourages $\bW$ to have a uniform distribution
facilitates efficient computation \cite{cuturi2013sinkhorn} and is defined by
\begin{equation}
    E\left (\bW \right ) = -\sum_{i,j}{\bW}_{ij}\left ( \log \bW_{ij} - 1\right ).
\end{equation}
% Note that when $\lambda = 0$, equation \eqref{eq::OT} is the classical (discrete) optimal transport \cite{villani2009optimal}.  

To solve \eqnref{eq::OT}, we use a variant of Sinkhorn's Algorithm \cite{sinkhorn1964relationship,marshall1968scaling}, given in \algoref{alg:sinkhorn}.
Unlike the standard algorithm that generates a square, doubly-stochastic matrix, our version generates a rectangular joint probability matrix, whose existence is guaranteed (Theorem $4$ \cite{marshall1968scaling}).

\vspace{3pt}
\noindent\emph{Joint Probability Loss Function:}
% \paragraph{Joint Probability Loss Function:}
% The best argument for this loss wrt a BCE loss is that it is equivalent to the theoretically sound weighted angular reprojection loss with a hard threshold on the dot product (~angle)
To train the feature extraction and matching network, we apply a loss function to the weighting matrix $\bW$.
%inlier set weights maximization loss function.
Since $\bW$ models the joint probability distribution of $\br$ and $\bs$, we can maximize the joint probability of inlier correspondences and minimize the joint probability of outlier correspondences using
% by \eqnref{eq::Pi}
% Recall that each item in the weighting matrix $\bW$ measures the match-ability of feature point pair. If we can identify the correct/incorrect matches, a simple idea is to maximize the match-abilities for correct matches while minimizing the match-abilities for incorrect matches.
%A remark is that $\bW$ models the joint probability distribution of $\br,\bs$ since $\bW\mathbf{1}^N = \br, \bW\transpose\mathbf{1}^M = \bs$ and $\sum_i \br_i = 1, \sum_j \bs_j = 1$.
% Note that the minimization of this loss maximizes the joint probabilities of matchable and minimizes the joint probabilities of non-matchable feature pairs, respectively.
% Our inlier set weights maximization loss function is defined by:
% \begin{equation} 
%     L = \sum_{i,j}g(i,j)w_{i,j}
% \end{equation}
% where $g(i,j)$ = $-1$ for matchable feature point pair, and $1$ for non-matchable pair.
% \vspace{-10pt}
\begin{equation}\label{eq::loss}
    L = \sum_{i}^{M}\sum_{j}^{N} \left(1 - 2\bC_{ij}^\text{gt}\right){\bW}_{ij},
\end{equation}
% \vspace{-10pt}
% \begin{align}
%     L &= \sum_{i,j} L_{ij} \text{, where}\\
%     L_{ij} &= 
%     \begin{cases}
%     -{\bW}_{ij} & \text{if } \bC_{ij}^\text{gt} = 1\\
%     +{\bW}_{ij} & \text{otherwise.}
%     \end{cases}
% \end{align}
where the ground-truth correspondence matrix $\bC_{ij}^\text{gt}$ is $1$ if $\{\bx_i, \by_j\}$ is a true correspondence and $0$ otherwise.
The loss is bounded, with $L \in [-1,1)$, since $\sum_{i=1}^{M}\sum_{j=1}^{N}{\bW}_{ij}=1$.
% This loss function treats all correspondences as equally important
% We also treat all matchable feature pairs as equally important since the summation of their probabilities are maximized.

If ground-truth correspondence labels $\bC_{ij}^\text{gt}$ are not available, they can be obtained in a weakly-supervised fashion by projecting the 3D points onto the image using the ground-truth camera pose and applying an inlier threshold.

\emph{Remark 1:}
A common objective in geometry optimization is minimizing re-projection error. With estimated weighting matrix $\bW$, the weighted angular reprojection error is defined by:
% This loss function \eqnref{eq::loss} may arise the weighted angular reprojection error
% \vspace{-10pt}
\begin{equation}\label{eq::weightReErr}
    L_\text{rep} = \sum_{i}^{M}\sum_{j}^{N}{\bW}_{ij}\left ( 1 -  \mathcal{N}({\bR_\text{gt}}{\bx}_i + \bt_\text{gt})\transpose \mathcal{N}({\bK^{-1} \hat{\by}}_j) \right ),
\end{equation}
where $\bR_\text{gt}$ and $\bt_\text{gt}$ are the ground-truth rotation and translation and $\mathcal{N}(\cdot)$ denotes $L_2$ normalization.
This loss minimizes the sum of weighted angular distances between image rays $\mathcal{N}({\bK^{-1} \hat{\by}}_j)$ and rays connecting the camera center and the transformed 3D points $\mathcal{N}(\bR_\text{gt}\bx_i+\bt_\text{gt})$.
% Equivalent to our loss function if a hard inlier threshold is applied on the dot product: if dot product > 1-eps, then dot product = 1, else dot product = -1.
While this loss is geometrically meaningful, we will show in the experiments that it is inferior to our loss.

\emph{Remark 2:}
A common technique for learning discriminative cross-modal features is deep metric learning.
We tested a triplet loss \cite{hu2018cvm} that minimizes the feature distance between matchable 2D--3D pairs and maximizes the feature distance between non-matchable 2D--3D pairs, given by
% Apparently, the quality of weighting matrix $\mathbf{W}$ depends on the discriminativeness of regressed 3D/2D feature descriptors. To learn discriminative feature descriptors, we impose a triplet loss on the $M$ and $N$ regressed feature descriptors $\{\mathbf{f}_i, i=1,2,...,M\}$, $\{\mathbf{f}_j, j=1,2,...,N\}$. We aim to minimize the feature distance between matchable 3D-2D correspondences, while maximizing the feature distance between non-matchable 3D-2D correspondences. This directly leads to our loss function:
% \vspace*{-10pt}

% \setlength{\abovedisplayskip}{0pt}
% \setlength{\belowdisplayskip}{0pt}
\begin{equation}
L_\text{tri} = \sum_{i}^{M} \log \left( 1 + e^{\alpha\left ( \| \bbf_{\bx_i} - \bbf_{\by}^+  \|_2^2 - \| \bbf_{\bx_i} - \bbf_{\by}^- \|_2^2 \right )} \right),
\end{equation}
where $\alpha = 10$ is chosen empirically.
% $\{\bbf_{\bx_i}, \bbf_{\by}^+\}$ are matchable and $\{\bbf_{\bx_i}, \bbf_{\by}^-\}$ are not
We use the ground-truth labels to find the positive and negative anchors, that is $\bbf_{\by}^+ = \bbf_{\by_j}$ for $\bC_{ij}^\text{gt} = 1$ and $\bbf_{\by}^- = \bbf_{\by_j}$ for $\bC_{ij}^\text{gt} = 0$, selected at random.
% To use this loss function, we also need ground-truth labels $\bC_{ij}^\text{gt}$ to obtain $\bbf_{\by_j}^+$ for $\bbf_{\bx_i}$.
$\bbf_{\by_j}^-$ is randomly selected from non-matchable features.
While this loss is effective, it also performs worse than our loss.

\vspace{3pt}
\noindent\emph{Retrieving Correspondences:}
% \paragraph{Retrieving Correspondences:}
To retrieve the 2D--3D correspondences from the weighting matrix $\bW$, we test the following methods.

\emph{1. Top-$K$ Prioritized Matches:}
To obtain a list of prioritized matches, we have (i) reshape $\bW$ into a 1D correspondence probability vector, sort by decreasing probability; or (ii) reshape $\bH$ into a 1D correspondence distance vector, sort by increasing distance, and then retrieve the associated 2D and 3D point indices. 
%We simply reshape $\bW$ into a 1D weight vector, and sort weights in decreasing order. For each weight $\bW_{ij}$, we retrieve its corresponding 3D--2D matches $(\bx_i, \by_j)$, with ${\bW}_{ij}$ describe the match priority.
Given this list of matches $(\bx_i, \by_j, \bW_{ij})$ or $(\bx_i, \by_j, \bH_{ij})$ prioritized by weight or distance, respectively, we truncate it to obtain the Top-$K$ matches. We denote the former by Top-K\_w and the latter by Top-K\_f.
Instead of enforcing one-to-one 2D--3D matches, we defer disambiguation to the match refinement stage.
%Note that one 3D point may be matched to multiple 2D points, and vice versa. Instead of enforcing the one-to-one 3D-2D matching constraint which reject positive matches, we postpone this disambiguation step to the following classification refinement and PnP RANSAC stages.

\emph{2. Nearest Neighbors (NNs):} For each 2D point $\by_j$, we find its nearest 3D neighbor $\bx_{i^\star}$ with respect to (i) the probability matrix, that is $i^\star = \argmax_{i} \bW_{ij}$; or (ii) the regressed descriptor, that is $i^\star = \argmin_{i} \| \bbf_{\bx_i} - \bbf_{\by_j} \|_2$.
We denote the former by NN\_w and the latter by NN\_f.
This approach retrieves $N$ matches.
%We have two options: 1) \textbf{NN\_f}: nearest neighbor search using the regressed descriptors, \ie, for each 2D point $\by_j$, its best match $\bx_{i^+}$ is determined by $\bx_{i^+} = \mathop{\arg\min}_{i \in 1,2,...,M} \left \| \mathbf{f}_{\by_j}-\mathbf{f}_{\bx_i} \right \|$; 2) \textbf{NN\_w}: best matchable 3D point in the weighting matrix $\mathbf{W}$, \ie, for each 2D point $\by_j$, its best match $\bx_{i^+}$ is determined by $\bx_{i^+} = \mathop{\arg\max}_{ i \in 1,2,...,M} {\bW}_{i,j}$. Note that by using nearest neighbor search, the number of founded 3D-2D matches is the same as the number of 2D feature points $N$.
% Since not all 2D feature points are matchable to 3D points, there could be many outliers in the obtained 3D-2D matches.

\emph{3. Mutual Nearest Neighbors (MNNs):}
We extend the previous approach by also enforcing a one-to-one constraint, that is, only keeping the correspondence $(\bx, \by)$ if $\bx$ is the nearest neighbor of $\by$ and $\by$ is the nearest neighbor of $\bx$.
We again compute nearest neighbors with respect to (i) the probability matrix (MNN\_w); or (ii) the regressed descriptors (MNN\_f).
This approach retrieves fewer than $\min\{M,N\}$ correspondences.
\subsection{Correspondence Set Refinement} \label{matches_refinement}

We now have a set of putative 2D--3D correspondences, some of which are outliers, and want to estimate the camera pose.
This is the standard PnP problem with outlier correspondences, and may be solved using RANSAC \cite{fischler1981random} with a minimal P3P solver \cite{grunert1841pothenotische}.
However, recent work \cite{yi2018learning,dang2018eigendecomposition} has shown that outliers can be filtered more efficiently using deep learning.
Therefore, we apply the 2D--2D correspondence classification network from Yi \etal~\cite{yi2018learning} to reject outliers, with a modified input dimension and loss function.

% \paragraph{Weighted DLT Loss:}
\paragraph{Regression Loss:}
We directly regress rotation $\bR$ and translation $\bt$ using the weighted Direct Linear Transform (DLT). Given at least six correspondences $(\bx_i, \by_j)$, $\bR$ and $\bt$ can be estimated by solving a SVD problem \cite{hartley2003multiple}.
We first construct the linear equation $\bA\bp=\mathbf{0}$, where each 2D--3D match supplies two rows in $\bA$, giving
\begin{equation}
\underbrace{\left [ \begin{matrix}
\mathbf{0}\transpose & -\bx_i\transpose & v_j\bx_i\transpose\\ 
\bx_i\transpose & \mathbf{0}\transpose & u_j\bx_i\transpose
\end{matrix} \right ]}_\bA
\underbrace{\left [
% \begin{matrix}
% \bR_1 & t_1 & \bR_2 & t_2 & \bR_3 & t_3
% \end{matrix}
\bR_1 \;\; t_1 \;\; \bR_2 \;\; t_2 \;\; \bR_3 \;\; t_3
\right ]\transpose}_\bp
% \left ( \begin{matrix}
% \bR_1\transpose\\ 
% t_1\\ 
% \bR_2\transpose\\ 
% t_2\\ 
% \bR_3\transpose\\ 
% t_3\\ 
% \end{matrix} \right )
= \mathbf{0},
\end{equation}
where $(u_j, v_j) = \by_j$, $(t_1, t_2, t_3) = \bt$, and $\bR_i$ is the i\textsuperscript{th} row of the rotation matrix.
The camera pose can be estimated by taking the SVD of $\bA\transpose \diag(\bw)\bA$, where $\bw$ is the vector of weights predicted by the classification network. The eigenvector associated with the smallest eigenvalue is the solution $\bp$ from which $\bR$ and $\bt$ can be assembled up to a sign ambiguity.
%$\mathbf{R}_{est}$ and $\mathbf{t}_{est}$ is further normalized using Frobenius and $L_2$ norm for numerical stability.
%
% The accuracy of above weighted DLT solution highly depends on the estimated weights of 3D-2D correspondences. Ideally, for outlier correspondences, $w_{i,j} = 0$. However, we find that $w_{i,j}$ from the weighting matrix $\mathbf{W}$ cannot reach this goal, \ie, {$w_{i,j} \to 0$} for outliers, since they are calculated based on the $L_2$ distance of descriptors.
%
Given ground-truth rotation  $\bR_\text{gt}$ and translation $\bt_\text{gt}$, we define our pose loss as
% \begin{align}
%   L_\text{p} &= \min\left \{ \left \| \bR - \bR_\text{gt} \right \|_\text{F}^2, \left \| \bR + \bR_\text{gt} \right \|_\text{F}^2 \right \} \nonumber \\
%   &+ \min\left \{ \left \| \bt - \bt_\text{gt} \right \|_2^2, \left \| \bt + \bt_\text{gt} \right \|_2^2 \right \}.
% \end{align}
\begin{equation}
      L_\text{p} = \min\left \{ \left \| \bR - \bR_\text{gt} \right \|_\text{F}^2, \left \| \bR + \bR_\text{gt} \right \|_\text{F}^2 \right \}  
  + \min\left \{ \left \| \bt - \bt_\text{gt} \right \|_2^2, \left \| \bt + \bt_\text{gt} \right \|_2^2 \right \}.
\end{equation}
%We have to compute both $\left \| \mathbf{t}_{est} - \mathbf{t}_{gt} \right \|$ and $\left \| \mathbf{t}_{est} + \mathbf{t}_{gt} \right \|$ since the sign of $\mathbf{p}_{est}$ are undefined by solving ${\mathbf{Ap}}=\mathbf{0}$.
Although we do not impose an orthogonality constraint on $\bR$, minimizing the above loss function pushes $\bR$ towards the Lie group of $\mathcal{SO}(3)$.

\paragraph{RANSAC and Nonlinear Optimization:}
At test time, we can further refine the pose by identifying inliers using RANSAC \cite{fischler1981random} with a minimal P3P solver \cite{grunert1841pothenotische}, followed by nonlinear optimization of the inlier reprojection error using the Levenberg--Marquardt \cite{more1978levenberg} algorithm.

\section{Experiments}
Our experiments are conducted on both synthetic (ModelNet40 \cite{wu20153d} and NYU-RGBD \cite{Silberman:ECCV12}) and real-world (MegaDepth \cite{MegaDepthLi18}) datasets. 
% Figure \ref{fig:sample_3d_2d_datasets} presents sample 3D and 2D point clouds from these datasets.
Sample 3D and 2D point clouds from these datasets are given in Appendix. 
We first validate the components of our pipeline and then compare it with state-of-the-art methods.
% \vspace{-15pt}
% \begin{figure}
% \begin{center}
% \includegraphics[width=0.15\textwidth]{Figures/model40_sample_3d.pdf}~~~~~~~~~~~~~~~~~~~~~
% \includegraphics[width=0.15\textwidth]{Figures/nyu_non_overlap_sample_3d.pdf}~~~~~~~~~~~~~~~~~~~~~
% \includegraphics[width=0.15\textwidth]{Figures/megaDepth_sample_3d.pdf}

% \includegraphics[width=0.15\textwidth]{Figures/model40_sample_2d.pdf}~~~~~~~~~~~~~~~~~~~~~
% \includegraphics[width=0.15\textwidth]{Figures/nyu_non_overlap_sample_2d.pdf}~~~~~~~~~~~~~~~~~~~~~
% \includegraphics[width=0.15\textwidth]{Figures/megaDepth_sample_2d.pdf}
% \end{center}
% \vspace{-10pt}
% \caption{Sample 3D (top row) and 2D (bottom row) points cloud from ModelNet40 (\textbf{Left}) and NYU-RGBD (\textbf{Middle}) and real-world MegaDepth (\textbf{Right}) datasets.
% }
% \label{fig:sample_3d_2d_datasets}
% \end{figure}

%\subsection{Datasets}

\vspace{3pt}
\noindent\emph{ModelNet40 \cite{wu20153d}:}
% \paragraph{ModelNet40 \cite{wu20153d}:}
We use the default train and test splits of $9\,843$ and $2\,468$ CAD mesh models respectively.
We uniformly sample $M = 1\,000$ 3D points from the surface of each model and generate virtual camera viewpoints as follows:
Euler rotation angles are uniformly drawn from $[0^\circ,45^\circ]$, translations are uniformly drawn from $[-0.5, 0.5]$, and a translation offset of $4.5$ is applied along the $z$ axis to ensure that all 3D points are in front of the camera.
The 3D points are projected onto a virtual image plane with size $640 \times 480$ and focal length $800$ and Gaussian pixel noise ($\sigma = 2$) is added to the $M$ 2D points.
In total, $40\,000$ training and $2\,468$ testing 2D--3D pairs are generated.
% Self-occluded points are also projected, resulting in outliers

\vspace{3pt}
\noindent\emph{NYU-RGBD \cite{Silberman:ECCV12}:}
% \paragraph{NYU-RGBD \cite{Silberman:ECCV12}:}
% We also perform our synthetic experiments using the NYU Depth Dataset V2 \cite{Silberman:ECCV12}.
We use train and test splits of $1\,100$ and $349$ aligned RGB and depth image pairs respectively from the labeled NYU Depth V2 dataset.
We uniformly sample $M = 1\,000$ 2D points from each RGB image, normalize the points using the intrinsic camera matrix, and find the corresponding 3D points in the depth image.
We transform the 3D points using virtual rotations and translations generated in the same way as for the ModelNet40 dataset, without the translation offset, and add Gaussian pixel noise ($\sigma = 2$) to the 2D points.
In total, $40\,000$ training and $10\,000$ testing 2D--3D pairs are generated. Note that the scenes in the train and test sets do not overlap.

\vspace{3pt}
\noindent\emph{MegaDepth \cite{MegaDepthLi18}:}
% \paragraph{MegaDepth \cite{MegaDepthLi18}:}
% To demonstrate the effectiveness of our method to handle real data, we use the MegaDepth \cite{MegaDepthLi18} dataset.
MegaDepth is a multi-view Internet photo dataset with multiple landmark scenes obtained from Flickr.
It has diverse scene contents, image resolutions, 2D--3D point distributions, and camera poses.
The dataset provides 3D point sets reconstructed using COLMAP \cite{schoenberger2016sfm}, and 2D SIFT keypoints detected from images.
We randomly select several landmarks, yielding a total number of $40\,828$ 2D--3D training sets and $10\,795$ testing sets.
The number of 2D--3D correspondences varies from tens to thousands.
Note that the landmarks in the train and test sets do not overlap.

\vspace{3pt}
\noindent\emph{Evaluation metrics:}
% \paragraph{Evaluation metrics:}
We report the number of inlier 2D--3D matches among all matches found, using ground-truth correspondence labels.
We also report the rotation error, given by $\epsilon = \arccos((\trace(\bR_\text{gt}\transpose \bR)-1)/{2})$, where $\bR_\text{gt}$ is the ground-truth rotation and $\bR$ is the estimated rotation, and the translation error, given by the $L_2$ distance between the estimated and ground-truth translation vectors.
%
% We also calculate the AUC scores (Area Under the Curve, equivalent to mean average precision) of rotation and translation \cite{yi2018learning}.
% % curve: normalised, cumulative precision (correctness of pose) versus threshold; 

We also calculate the recalls (percentage of poses) by varying pre-defined thresholds on rotation and translation error. For each threshold, we count the number of poses with error less than that threshold, and then normalise the number by the total number of poses.

% percentage of  AUC scores (Area Under the Curve, equivalent to mean average precision) of rotation and translation \cite{yi2018learning}.
% % curve: normalised, cumulative precision (correctness of pose) versus threshold; 

\vspace{3pt}
\noindent\emph{Implementation details:}
% \paragraph{Implementation details:}
Our 12-layer two-stream network is implemented in TensorFlow and is trained from scratch using the Adam optimizer \cite{kingma2014adam} with a learning rate of $10^{-5}$ and a batch size of $12$.
%from randomly initialized network weights.
Every layer has an output channel dimension of $128$.
The number of Sinkhorn iterations is set to $20$, $\lambda$ is set to $0.1$, and the number of neighbors in the KNN-graph is set to $10$.
We utilize a two-stage training strategy: we first train our feature extraction and matching network until convergence and then train the classification network to refine 2D--3D matches.
Our model is trained on a single NVIDIA Tesla P40 GPU in $3$ days.
Code and data will be released.

\subsection{Synthetic Data Experiments}

To validate the components of our pipeline, we perform experiments on the synthetic ModelNet40 \cite{wu20153d} and NYU-RGBD \cite{Silberman:ECCV12} datasets.

\begin{figure}
\begin{center}
\includegraphics[width=0.45\textwidth]{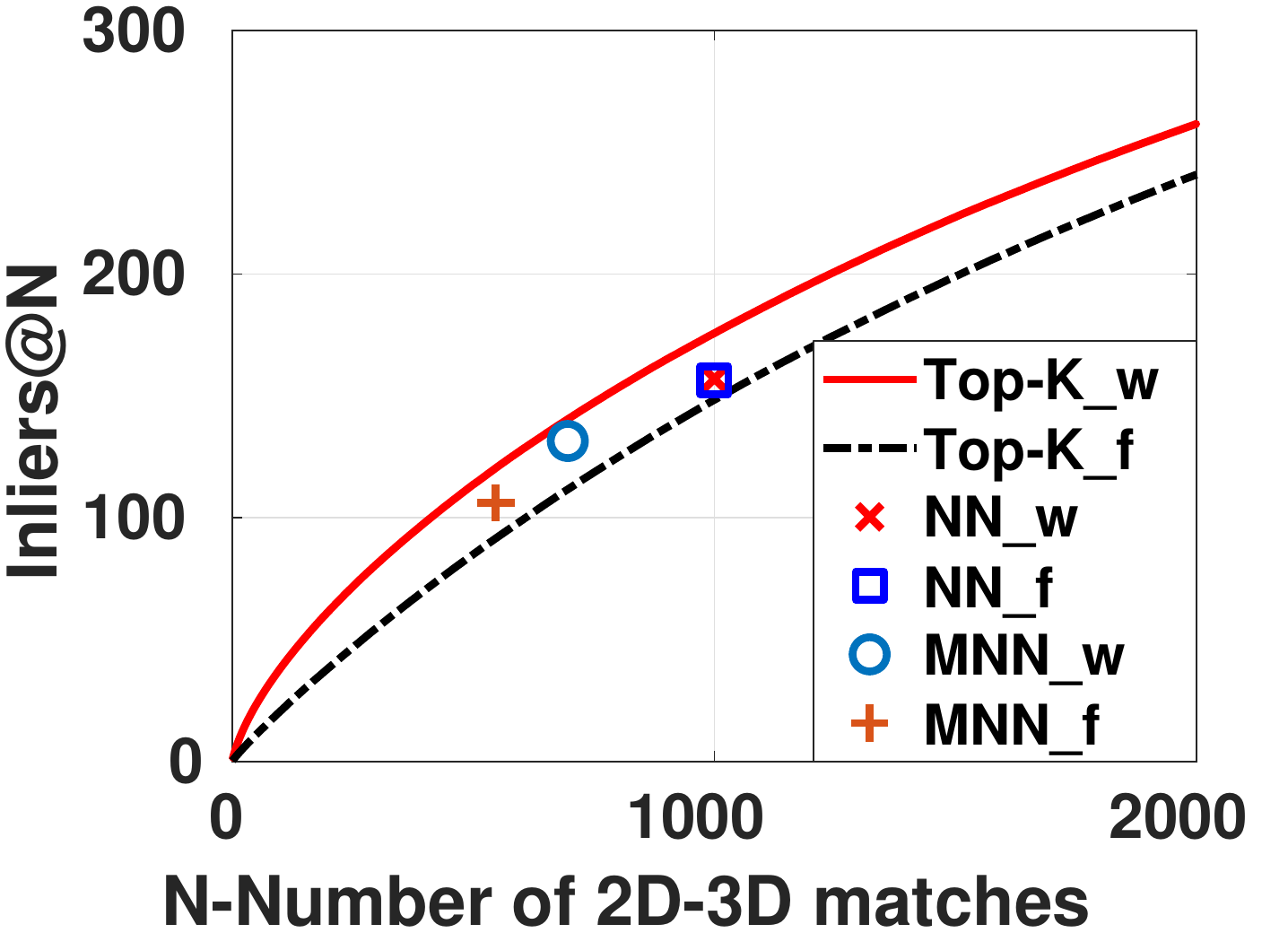}~~~~~~~~~~~~~~~~~~
\includegraphics[width=0.45\textwidth]{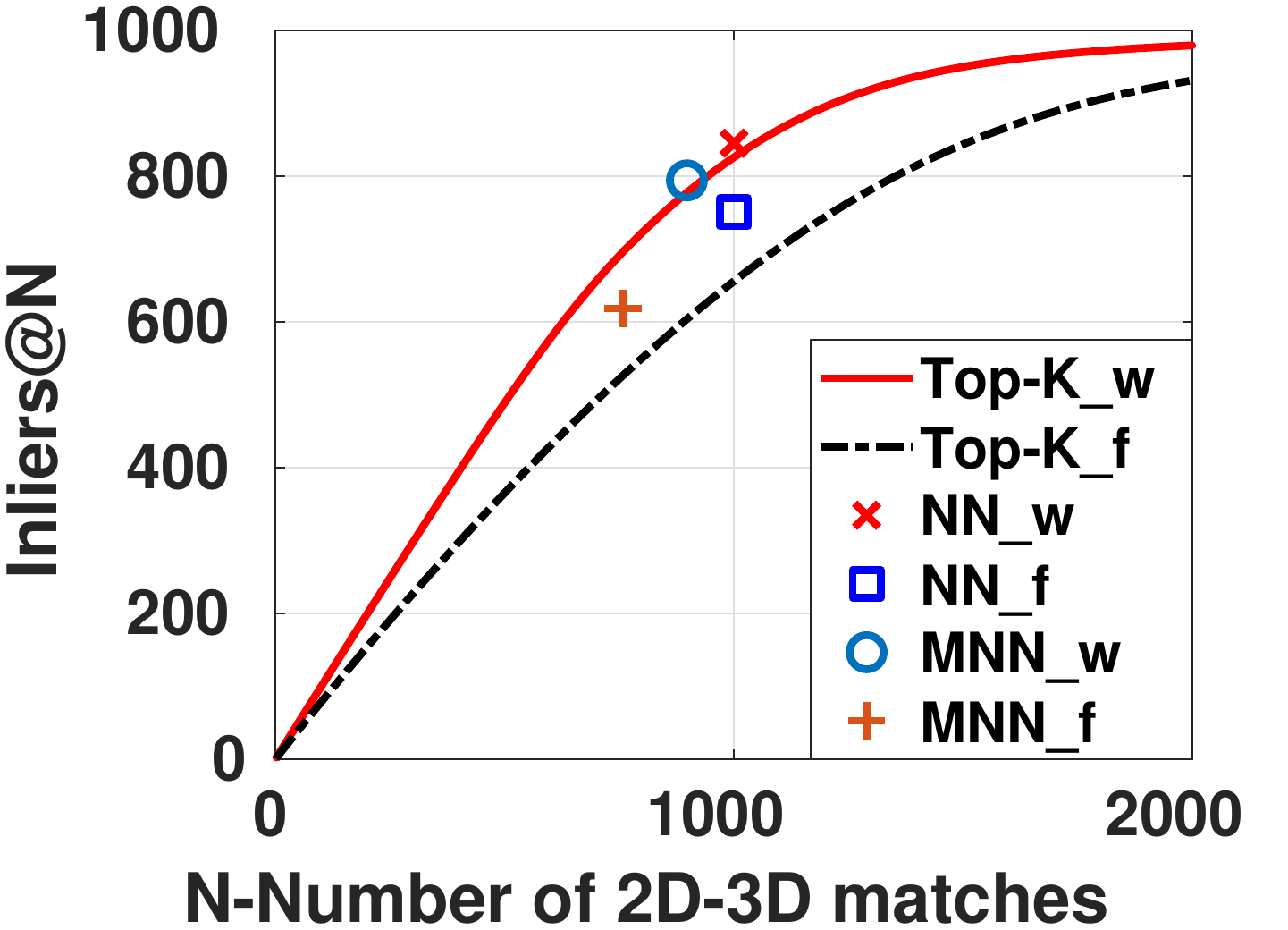}
\end{center}
% \vspace{-10pt}
\caption{Average number of inlier matches with respect to the number of found 2D--3D matches. \textbf{Left}: ModelNet40. \textbf{Right}: NYU-RGBD.
% {Top-K} denotes truncating the  prioritized matching list at the $K$-th position.  {NN\_f} denotes performing nearest neighbor search to find 3D-2D matching using learned descriptors. {NN\_w} denotes selecting the best matchable 3D point for each 2D point using weights in $\mathbf{W}$.  {MNN\_f} and {MNN\_w} denote enforcing mutual nearest neighbor constraint on {NN\_f} and {NN\_w}, respectively.
}
\label{fig:inlier_global_matching}
\end{figure}

\vspace{3pt}
\noindent\emph{The effectiveness of global matching:}
% \paragraph{The effectiveness of global matching:}
Given a point-wise regressed descriptor and a 2D--3D weighting matrix $\bW$, we have $6$ methods for retrieving 2D--3D correspondences as listed in \secref{sec:feature_matching}:  {Top-K\_w}, {Top-K\_f}, {NN\_w}, {NN\_f}, {MNN\_w} and {MNN\_f}. We calculate the number of inlier matches using ground-truth labels, and the results are shown in \figref{fig:inlier_global_matching}.
The number of inlier 2D--3D correspondences found by {NN\_w} and {MNN\_w} is consistently greater than the number found by {NN\_f} and {MNN\_f} respectively. This demonstrates that retrieving 2D--3D matches from the weighting matrix $\bW$ is better than performing nearest neighbor search using the regressed descriptors.

For the methods {Top-K\_w} and {Top-K\_f}, since we can truncate the prioritized 2D--3D matching list at the $K$\textsuperscript{th} ($K\leq MN$) position, we plot the curve showing the number of inliers with respect to the number of found 2D--3D correspondences for $K$ up to $2000$. Again, method {Top-K\_w} outperforms {Top-K\_f}.
% It shows that  the number of inliers increases with respect to the number of found matches.
Interestingly, match--inlier tuples found by {NN\_w} and {MNN\_w} lie very close to the {Top-K\_w} curve. We use the {Top-K\_w} method (omit the subscript) in the remaining experiments, since it enables us to select a sufficient number of matches while also finding large number of inliers.
%to be tailored to the dataset characteristics, although we do not fine-tune it in these experiments.
% If fixed number of matches are needed, method {NN\_w} can be used.
Our method finds fewer inlier correspondences on the ModelNet40 dataset than the NYU-RGBD dataset, since the virtual camera can only view part of the whole 3D model with the remainder being occluded.
% \vspace{-15pt}

% \vspace{-40pt}

% \vspace{-10pt}

\begin{figure}
\begin{center}
\includegraphics[width=0.45\textwidth]{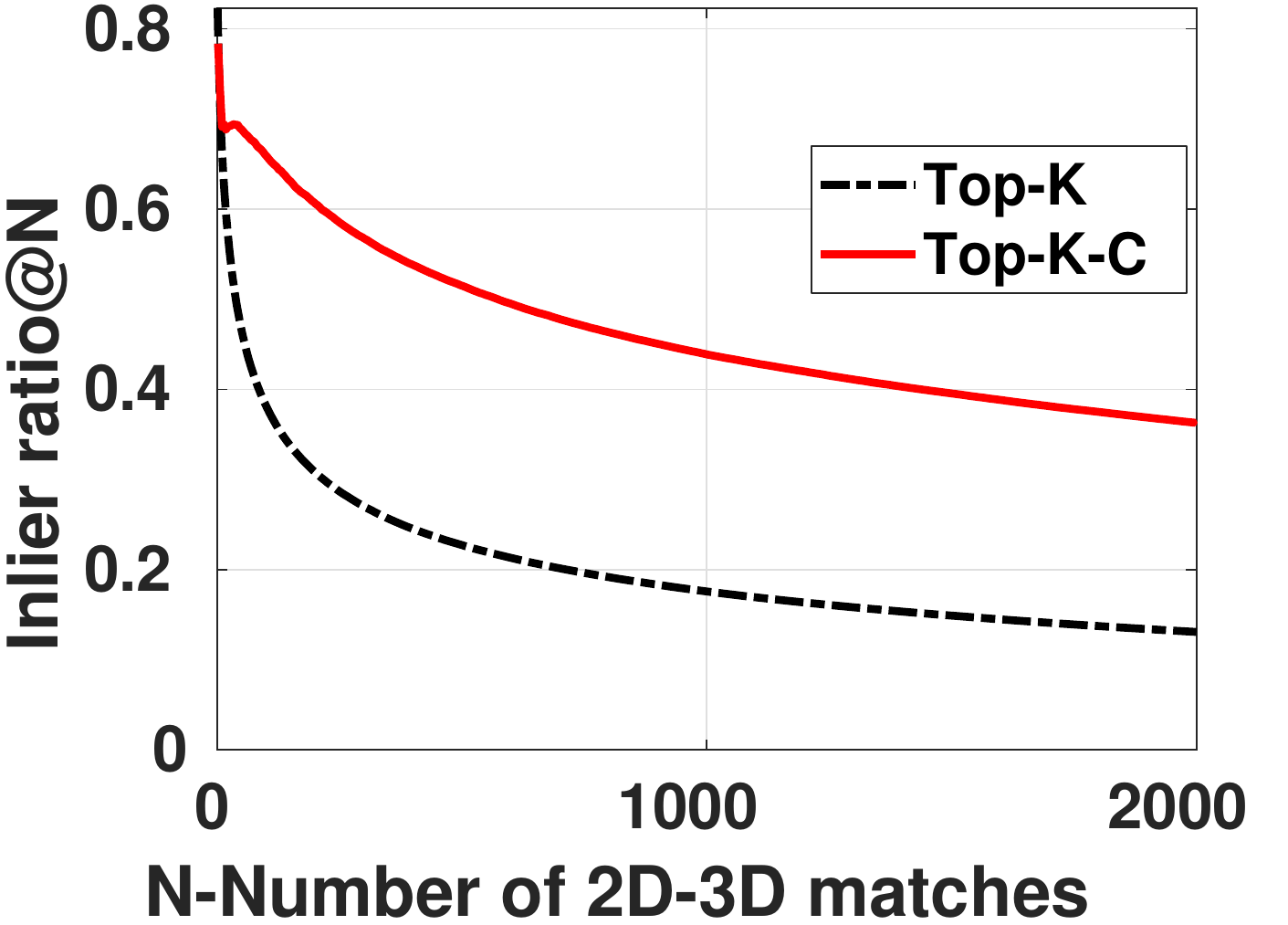}~~~~~~~~~~~~~~~~~~
\includegraphics[width=0.45\textwidth]{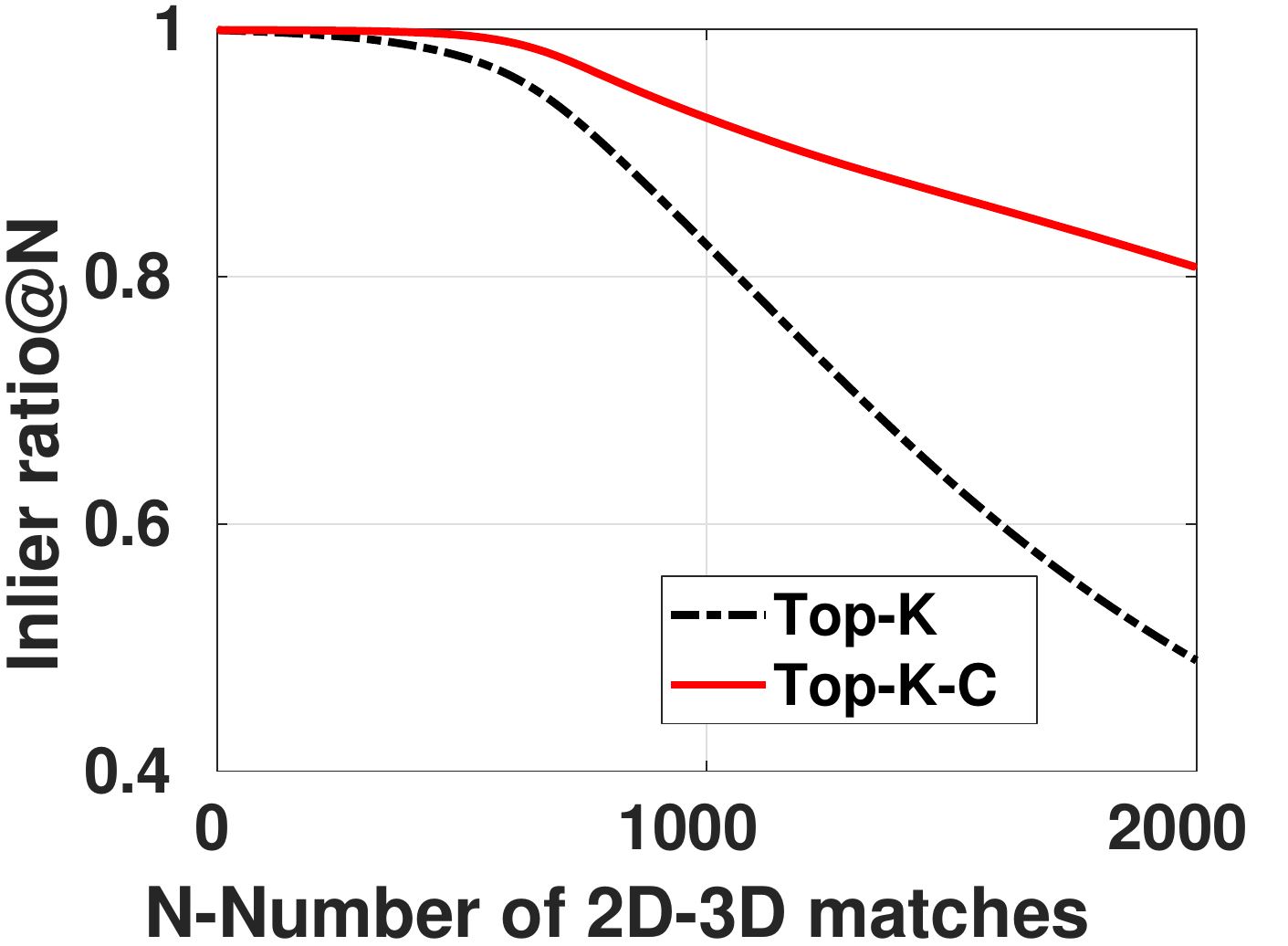}
\end{center}
% \vspace{-10pt}
\caption{Average inlier ratio with respect to the number of found 2D--3D matches. \textbf{Left}: ModelNet40. \textbf{Right}: NYU-RGBD. {Top-K} denotes truncating the prioritized matching list at the $K$\textsuperscript{th} position. {Top-K-C} denotes additionally using the classification network to disambiguate inliers from outliers.
}
% \vspace{10pt}
\label{fig:inlier_classif}
\end{figure}

% \vspace{-3pt}
\noindent\emph{The effectiveness of 2D--3D classification:}
% \paragraph{The effectiveness of 2D--3D classification:}
We evaluate the ability of the 2D--3D correspondence classification module to disambiguate inlier and outlier correspondences by running this network with the Top-$K$ ($K\in[1,2000]$) matches from the prioritized 2D--3D match list. We calculate the average inlier ratio (\#inlier/\#matches) with respect to the number of found 2D--3D matches, as shown in \figref{fig:inlier_classif}.
The results demonstrate that the classification network significantly improves the average inlier ratio, which improves the pose estimation considerably, as shown in the next experiment.

% \vspace{-4pt}
\begin{table}[]
\setlength{\tabcolsep}{3pt}
% \footnotesize
\scriptsize
\caption{Comparison of rotation and translation errors on the ModelNet40 and NYU-RGBD datasets. Q1 denotes the first quartile, Med. denotes the median, and Q3 denotes the third quartile.}
% \begin{tabular}{llllllll}
\begin{tabularx}{\linewidth}{llXXXXXX}
\hline
\multicolumn{2}{l}{\multirow{2}{*}{\backslashbox{Method}{Dataset}}}      & \multicolumn{3}{c}{ModelNet40} & \multicolumn{3}{c}{NYU-RGBD} \\
\cline{3-8}
% \cmidrule(l{3pt}r{3pt}){3-5}
% \cmidrule(l{3pt}r{3pt}){6-8}
\multicolumn{2}{l}{}                       & Q1     & Med.     & Q3     & Q1      & Med.     & Q3     \\ \hline
\multirow{2}{*}{Top-K}    & Rot. err. ($^{\circ}$)    & 4.316  & 10.85  & 19.30 & 0.303   & 0.448   & 0.645  \\
% \cline{2-8} 
                         & Trans. err. & 0.041  & 0.088   & 0.196  & 0.014   & 0.022   & 0.033  \\ \hline
\multirow{2}{*}{Top-K-C} & Rot. err. ($^{\circ}$)    & \textbf{1.349}  & \textbf{2.356}   & \textbf{5.260}  & \textbf{0.202}   & \textbf{0.291}   & \textbf{0.407}  \\
% \cline{2-8} 
                         & Trans. err. & \textbf{0.018} & \textbf{0.037}   & \textbf{0.070}  & \textbf{0.009}   & \textbf{0.014}   & \textbf{0.020}  \\ \hline
\end{tabularx}
\label{tab::rotation_trans_error_synthetic}
\end{table}

% Though we lost small amount of inliers with the classification CNN, 
% \vspace{-15pt}
% \vspace{-10pt}

% \vspace{-13pt}
\begin{table}[]
\setlength{\tabcolsep}{3.8pt}
\scriptsize
% \vspace{-4pt}
\caption{Comparison of rotation and translation errors on the ModelNet40 dataset.}
\begin{tabularx}{\linewidth}{llXXXXXX}
\hline
\multicolumn{2}{l}{\multirow{2}{*}{\backslashbox{Method}{Error}}} & \multicolumn{3}{c}{Rotation ($^{\circ}$) }  & \multicolumn{3}{c}{Translation} \\ 
\cline{3-8} 
% \cmidrule(l{3pt}r{3pt}){3-5}
% \cmidrule(l{3pt}r{3pt}){6-8}
\multicolumn{2}{l}{}                  & Q1    & Med.    & Q3     & Q1     & Med.    & Q3     \\ \hline
\multicolumn{2}{l}{P3P-RANSAC}             & 90.82 & 138.5 & 164.8 & 0.433  & 1.147  & 3.077  \\ 
\multicolumn{2}{l}{SoftPOSIT \cite{david2004softposit}}             & 16.10 & 21.75 & 28.00 & 0.332  & 0.488  & 0.719  \\ 
\multicolumn{2}{l}{GOSMA \cite{campbell2019alignment}}             & 10.08 & 22.06 & 52.01 & 0.254  & 0.464  & 0.746  \\ \hline
\multicolumn{2}{l}{Ours}               & \textbf{1.349} & \textbf{2.356}  & \textbf{5.260}  & \textbf{0.018}  & \textbf{0.037}  & \textbf{0.070}  \\ \hline
\end{tabularx}
\label{tab::rotation_trans_error_stoa_modelnet}
\end{table}

\begin{figure}
\begin{center}
\includegraphics[width=0.9\textwidth]{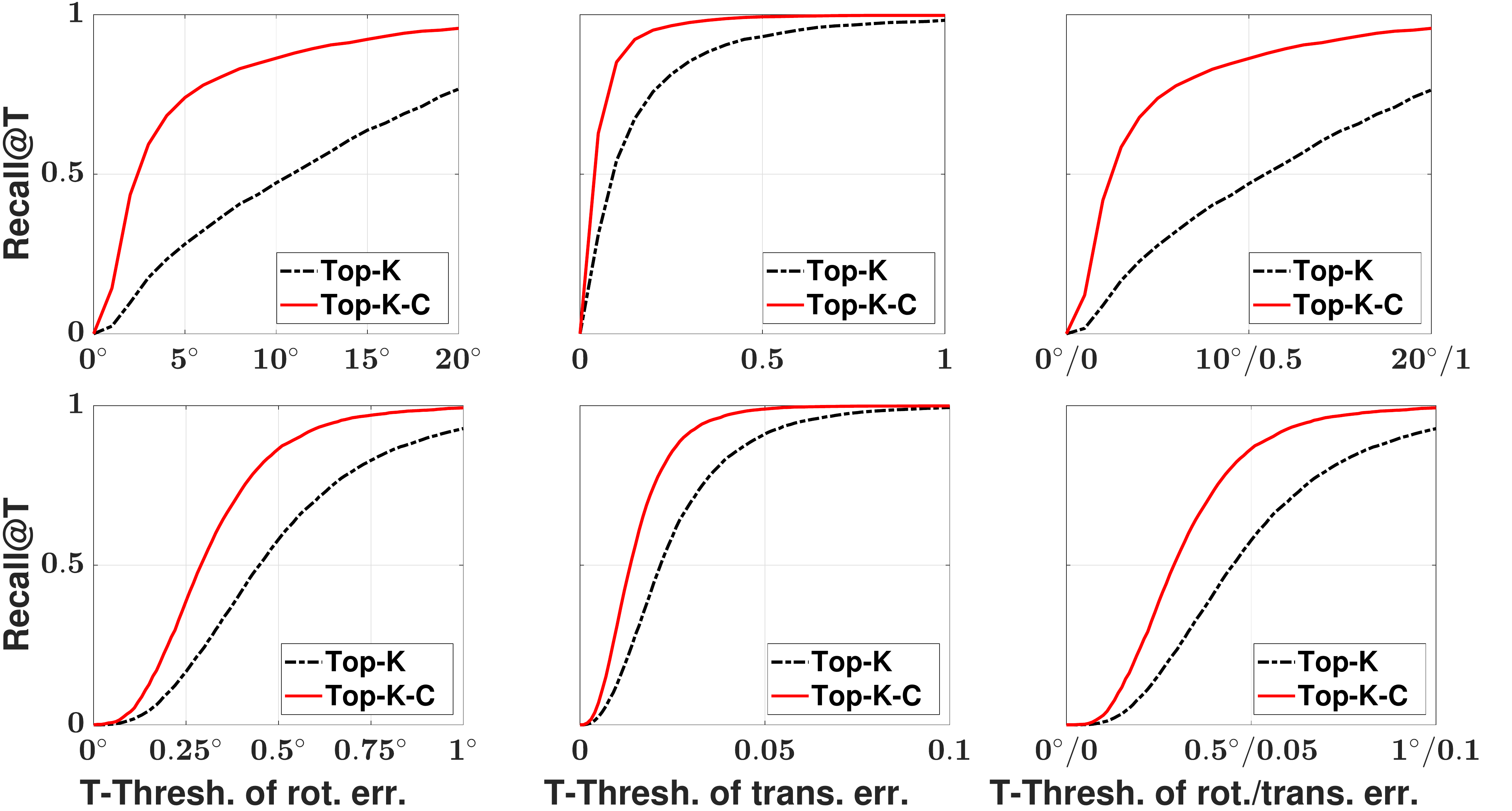}
\end{center}
% \vspace{-10pt}
\caption{Comparison of recall with respect to error thresholds on rotation and translation. \textbf{Top}: ModelNet40. \textbf{Bottom}: NYU-RGBD. 
% \vspace{10pt}
}
\label{fig:poses_AUCclassif}
\end{figure}

\vspace{3pt}
\noindent\emph{Estimating 6-DoF pose:}
% \paragraph{Estimating 6-DoF pose:}
Once the 2D--3D matches have been established, we apply the P3P algorithm in a RANSAC framework to estimate the 6-DoF camera pose. We compare two methods: (a) P3P with Top-K matches ($K=2000$); and (b) P3P with Top-K matches and classification network filtering. The results are shown in \figref{fig:poses_AUCclassif}. Observe that after classification filtering we consistently obtain larger recalls at each error thresholds. The same trend is visible in the rotation and translation error statistics, shown in \tabref{tab::rotation_trans_error_synthetic}.
Due to these conclusive results, we thus use the Top-K + Classification ({Top-K-C}) method as our default configuration.
The performance of state-of-the-art methods on ModelNet40 is presented in \tabref{tab::rotation_trans_error_stoa_modelnet}; see the Appendix for a full comparison on both synthetic datasets. Our method outperforms all others by a large margin, in addition to a $>100\times$ speed-up.
% \vspace{-15pt}

% \vspace{-34pt}

% \vspace{-20pt}
% \begin{figure}
% \begin{center}
% \includegraphics[width=0.23\textwidth]{Figures/model40rotation_classif.pdf}
% \includegraphics[width=0.23\textwidth]{Figures/nyu_non_overlaprotation_classif.pdf}
% \includegraphics[width=0.23\textwidth]{Figures/model40trans_classif.pdf}
% \includegraphics[width=0.23\textwidth]{Figures/nyu_non_overlaptrans_classif.pdf}
% \end{center}
% \caption{Comparison of AUC scores with respect to pre-defined error thresholds on rotation and translation. \textbf{Left}: ModelNet40; \textbf{Right}: NYU-RGBD. 
% }
% \label{fig:poses_AUCclassif}
% \end{figure}

% Furthermore, to demonstrate the effectiveness of the propose network, we replace our network with PointNet \cite{}, CNNet \cite{}, and EdgeNet \cite{}.
% \vspace{-30pt}
\begin{table}[]
\setlength{\tabcolsep}{3.8pt}
\scriptsize
% \vspace{-4pt}
\caption{Comparison of rotation and translation errors on the MegaDepth dataset.}
\begin{tabularx}{\linewidth}{llXXXXXX}
\hline
\multicolumn{2}{l}{\multirow{2}{*}{\backslashbox{Method}{Error}}} & \multicolumn{3}{c}{Rotation ($^{\circ}$) }  & \multicolumn{3}{c}{Translation} \\
\cline{3-8} 
% \cmidrule(l{3pt}r{3pt}){3-5}
% \cmidrule(l{3pt}r{3pt}){6-8}
\multicolumn{2}{l}{}                  & Q1    & Med.    & Q3     & Q1     & Med.    & Q3     \\ \hline
\multicolumn{2}{l}{P3P-RANSAC}             & 66.64 & 122.1 & 155.4 & 6.796  & 15.18  & 28.18  \\ 
\multicolumn{2}{l}{SoftPOSIT \cite{david2004softposit}}             & 1.806 & 21.39 & 165.4 & 0.242  & 1.532  & 6.101  \\ 
\multicolumn{2}{l}{GOSMA \cite{campbell2019alignment}}             & 8.685 & 86.78 & 144.5 & 1.070  & 5.670  & 9.335  \\ \hline
\multicolumn{2}{l}{Ours}               & \textbf{0.028} & \textbf{0.056}  & \textbf{0.137}  & \textbf{0.002}  & \textbf{0.005}  & \textbf{0.018}  \\ \hline
\end{tabularx}
\label{tab::rotation_trans_error_stoa}
\end{table}

% \vspace{-20pt}
\subsection{Real Data Experiments}
We further demonstrate the effectiveness of our method to handle real-world data on the MegaDepth \cite{MegaDepthLi18} dataset.

\begin{figure}
\begin{center}
\includegraphics[width=0.85\textwidth]{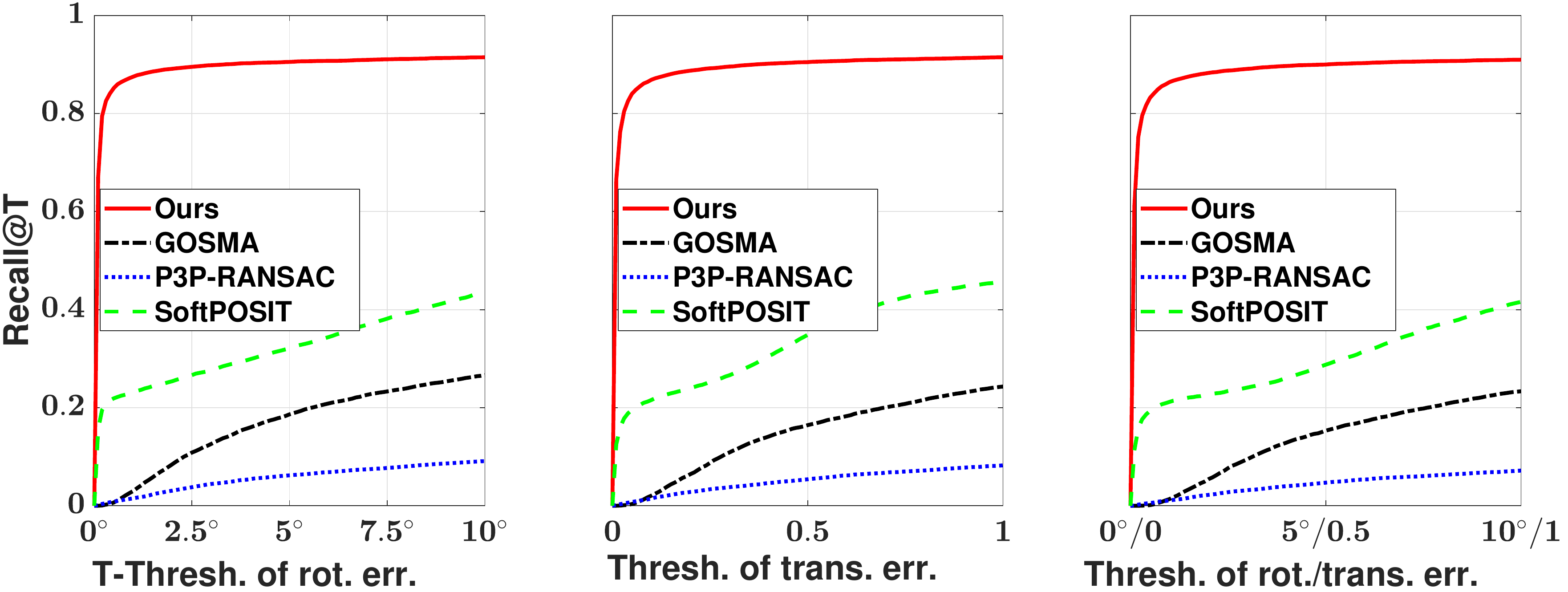}
\end{center}
% \vspace{-10pt}
\caption{Comparison of rotation and translation errors on the MegaDepth dataset. Recall with respect to error thresholds on rotation (\textbf{Left}), translation (\textbf{Middle}), and both (\textbf{Right}) are plotted.
% \vspace{14pt}
}
\label{fig:poses_stoa}
\end{figure}

\begin{figure}
\begin{center}

\begin{subfigure}[t]{0.19\textwidth}
\includegraphics[width=\textwidth]{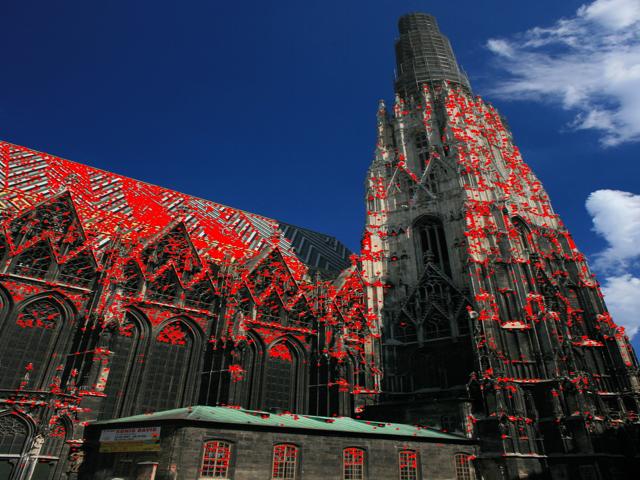}
\caption{GT}
\end{subfigure}
\hfill
\begin{subfigure}[t]{0.19\textwidth}
\includegraphics[width=\textwidth]{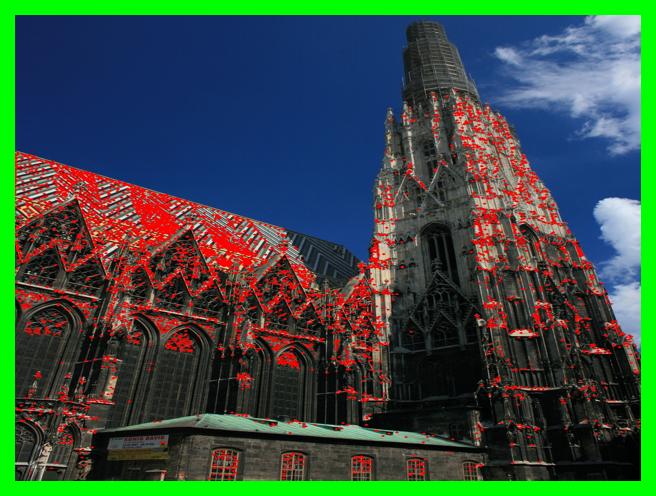}
\caption{Our}
\end{subfigure}
\hfill
\begin{subfigure}[t]{0.19\textwidth}
\includegraphics[width=\textwidth]{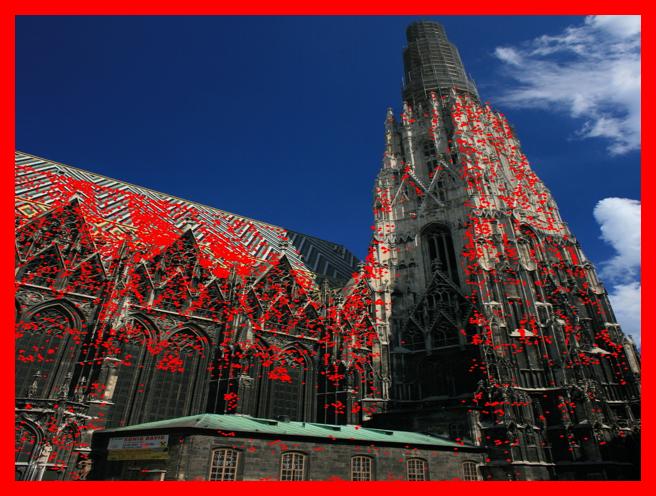}
\caption{GOSMA}
\end{subfigure}
\hfill
\begin{subfigure}[t]{0.19\textwidth}
\includegraphics[width=\textwidth]{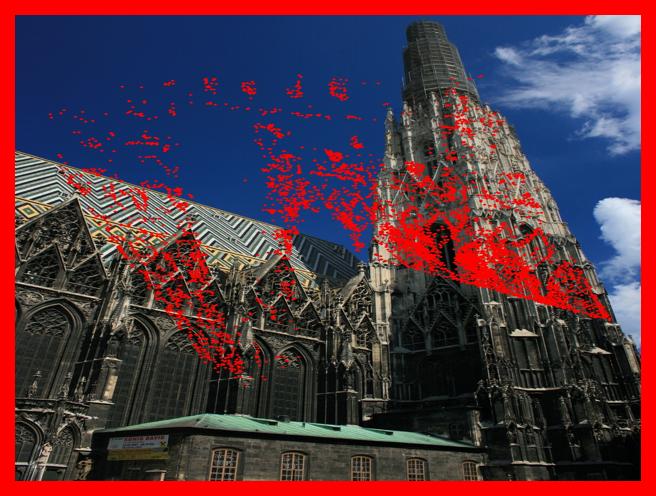}
\caption{SoftPOSIT}
\end{subfigure}
\hfill
\begin{subfigure}[t]{0.19\textwidth}
\includegraphics[width=\textwidth]{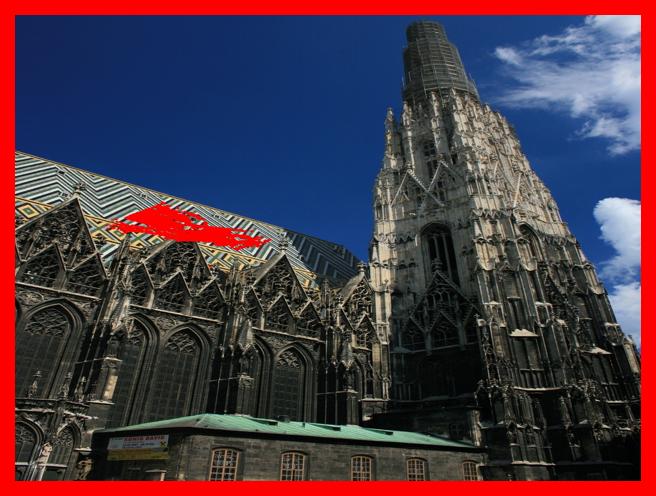}
\caption{\scriptsize P3P-RANSAC}
\end{subfigure}

\end{center}
% \vspace{4pt}
\caption{Qualitative comparison with state-of-the-art methods on the MegaDepth dataset, showing the projection of 3D points onto images using poses estimated by different methods. {\color{green} Green} border indicates the rotation/translation error of the estimated pose is less than  $5^\circ$/$0.5$ while {\color{red} red} border indicates the rotation/translation error is larger than  $5^\circ$/$0.5$. Only our method found the correct 6-DoF pose. (best viewed in color). More comparisons are given in the Appendix.
}
% \vspace{15pt}
\label{fig:Qualitative_cmp}
\end{figure}

\vspace{3pt}
\noindent\emph{Comparison with state-of-the-art methods:}
% \paragraph{Comparison with state-of-the-art methods:}
The unavailability of Gaussian pose priors and the sheer number ($\sim \!\! 1000$) of 2D and 3D points precludes the use of the methods BlindPnP \cite{moreno2008pose} and GOPAC \cite{campbell2017globally}.
We compare our method against P3P-RANSAC with randomly-sampled 2D--3D correspondences and the state-of-the-art local solver SoftPOSIT \cite{david2004softposit} and global solver GOSMA \cite{campbell2019alignment}. All methods are terminated at $\sim\!\!30$s per alignment for time considerations, returning the best value found so far.
Note that with this approach, GOSMA's guarantee of global optimality is traded off against its runtime. For randomly-sampling 2D--3D correspondences in P3P-RANSAC, the probability of finding minimal inlier correspondences set  approximates zero (see Appendix).
Since SoftPOSIT requires a good prior pose, we simulate it by adding a small perturbation to the ground-truth pose, with the angular perturbation drawn uniformly from $[-10^{\circ}, 10^{\circ}]$ and the translation perturbation drawn uniformly from $[-0.5, 0.5]$. The configuration details of these methods are given in the Appendix. The performance for 6-DoF pose estimation is shown in \figref{fig:poses_stoa} and \tabref{tab::rotation_trans_error_stoa}. It shows that our method outperforms the second-best method GOSMA,\footnote{SoftPOSIT is initialized with a good prior pose from the ground-truth, and so cannot be compared directly.}
% thus it possesses privilege at the beginning. It's unfair to compare it with other methods.}
with median rotation and translation errors of $0.056^{\circ}$ and $0.005$ for our method, and $86.784^{\circ}$ and $5.670$ for GOSMA.
The qualitative comparisons in \figref{fig:Qualitative_cmp} show that the projection of 3D points using our method's pose aligns very well with the images.
% It shows that using 6-DoF poses estimated by our method, the projections align well with images, and are very similar to projections obtained using ground-truth 6-DoF poses.
% The average runtime of our method, SoftPOSIT, P3P-RANSAC and GOSMA is $0.15$s, $18.19$s, $29.09$s and $35.11$s respectively. The small fluctuation of runtime of GOSMA is for that we do not put restrictions on its mixture generation part which handles irregular number of 3D points for MegaDepth.
The average runtime of our method, SoftPOSIT, P3P-RANSAC and GOSMA is $0.15s$, $18s$, $30s$ and $30s$ respectively, where the last three algorithms were run for a maximum runtime of $30s$.
% \vspace{-20pt}

% % \vspace{-30pt}
% \begin{figure}
% \begin{center}
% \includegraphics[width=0.22\textwidth]{Figures/gt_img_72.jpg}~~~~~~
% \includegraphics[width=0.22\textwidth]{Figures/our_img_72.jpg}~~~~~~
% \includegraphics[width=0.22\textwidth]{Figures/gosma_img_72.jpg}

% % \includegraphics[width=0.22\textwidth]{Figures/gt_img_8252.jpg}~~~~~~
% % \includegraphics[width=0.22\textwidth]{Figures/our_img_8252.jpg}~~~~~~
% % \includegraphics[width=0.22\textwidth]{Figures/gosma_img_8252.jpg}
% % \includegraphics[width=0.155\textwidth]{Figures/gt_img_5719.jpg}
% % \includegraphics[width=0.155\textwidth]{Figures/our_img_5719.jpg}
% % \includegraphics[width=0.155\textwidth]{Figures/gosma_img_5719.jpg}
% \end{center}
% \vspace{-10pt}
% \caption{Qualitative comparison with the state-of-the-art GOSMA algorithm \cite{campbell2019alignment} on the MegaDepth dataset, showing the projection of 3D points onto images. Only our method found the correct 6-DoF pose. \textbf{Left}: projection using the ground-truth poses; \textbf{Middle}: using our method's estimated poses; \textbf{Right}: using GOSMA's estimated poses (best viewed in color). More comparisons are given in the supplementary material.
% \vspace{15pt}
% }
% \label{fig:Qualitative_cmp}
% \end{figure}

% \vspace{-30pt}

% \subsection{Ablation Study}
% We give detailed analysis of our method using the real-world MegaDepth \cite{MegaDepthLi18} dataset.

\begin{figure}
\begin{center}
\includegraphics[width=0.45\textwidth]{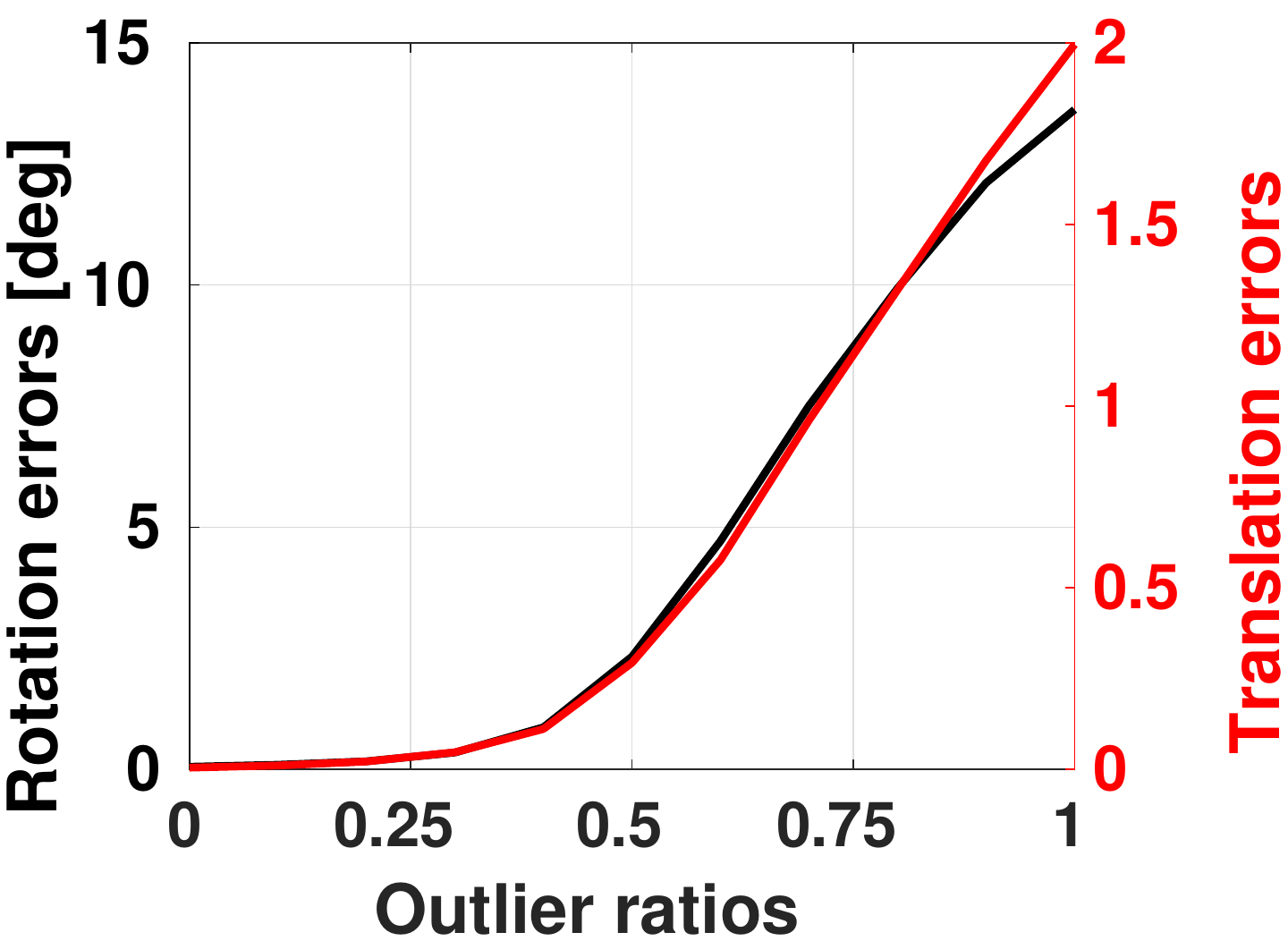}~~~~~~~~~~~~~~~~~~
\includegraphics[width=0.45\textwidth]{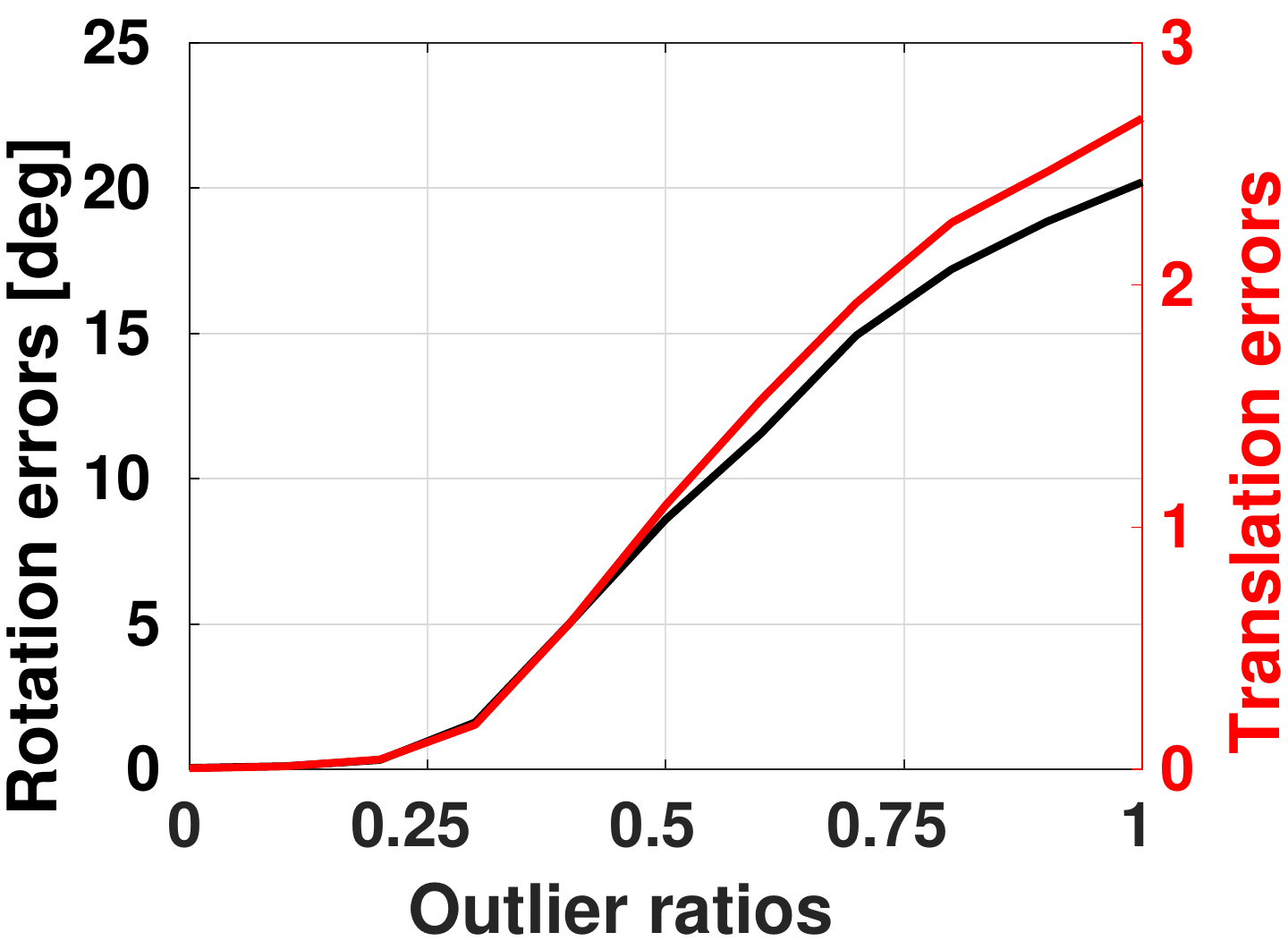}
\end{center}
% \vspace{-10pt}
\caption{Robustness to outliers on the MegaDepth dataset. Median rotation and translation errors with respect to the outlier ratio. \textbf{Left}: synthetic. \textbf{Right}: real-world outliers. 
}
\label{fig:poses_outlier_ratios_synthetic}
\end{figure}

\vspace{3pt}
\noindent\emph{Robustness to outliers:}
% \paragraph{Robustness to outliers:}
To demonstrate the effectiveness of our method at handling outliers, we add outliers to both the 3D and 2D point-sets. Specifically, for original 3D and 2D point-sets with cardinality $M$ and $N$, we add $\nu M$ and $\nu N$ outliers to the 3D and 2D point-sets, respectively, for an outlier ratio $\nu \in [0,1]$.
We add two types of outliers: synthetic and real. For synthetic outliers, they are generated uniformly within the bounding box enclosing the 3D and 2D point-sets. The rotation and translation errors with respect to the outlier ratio are given in \figref{fig:poses_outlier_ratios_synthetic} (Left). For real outliers, 2D outliers are added from detected SIFT keypoints that do not have a matchable 3D point, and 3D outliers are added from 3D model points that do not have a matchable 2D point. The rotation and translation errors with respect to the outlier ratio are given in \figref{fig:poses_outlier_ratios_synthetic} (Right).
It shows that the performance of our method degrades gracefully with respect to an increasing outlier ratio. See Appendix for more comparisons.
% \vspace{-20pt}

% \vspace{-20pt}
% \paragraph{Robustness to outliers (Real)}
% To demonstrate the effectiveness of our method to handle outliers, we add outliers to both 3D and 2D points sets. Specifically, for 3D and 2D points set with cardinality at $M$ and $N$, we add $\nu M$ and $\nu N$ outliers to 3D and 2D points set, respectively. Outlier ratio $\nu \in [0,1]$. For 2D points set, outliers are added with detected image feature points which do not have matchable 3D points. For 3D points set, outliers are added with 3D points from the model that don't have matchable 2D points. The histogram of number of inliers, rotation and translation errors with respect to outlier ratios are given in Figure \ref{fig:poses_outlier_ratios_real}. It shows that the performance of our method degrades gracefully with respect to increasing outlier ratios.

% \begin{figure}
% \begin{center}
% \includegraphics[width=0.23\textwidth]{Figures/megaDepthreal_outlierinliers_outlierRatio.pdf}
% \includegraphics[width=0.24\textwidth]{Figures/megaDepthreal_outlierrotation_trans_outlier.pdf}
% % \includegraphics[width=0.23\textwidth]{Figures/model40trans_classif.pdf}
% % \includegraphics[width=0.23\textwidth]{Figures/nyu_non_overlaptrans_classif.pdf}
% \end{center}
% \caption{Robustness to outliers on MegaDepth dataset. Left: histogram of number of inliers  with respect to outlier ratios. Right: median rotation and translation errors with respect to outlier ratios. (Best viewed in color with zoom in)
% }
% \label{fig:poses_outlier_ratios_real}
% \end{figure}

\begin{figure}
\begin{center}
\includegraphics[width=0.45\textwidth]{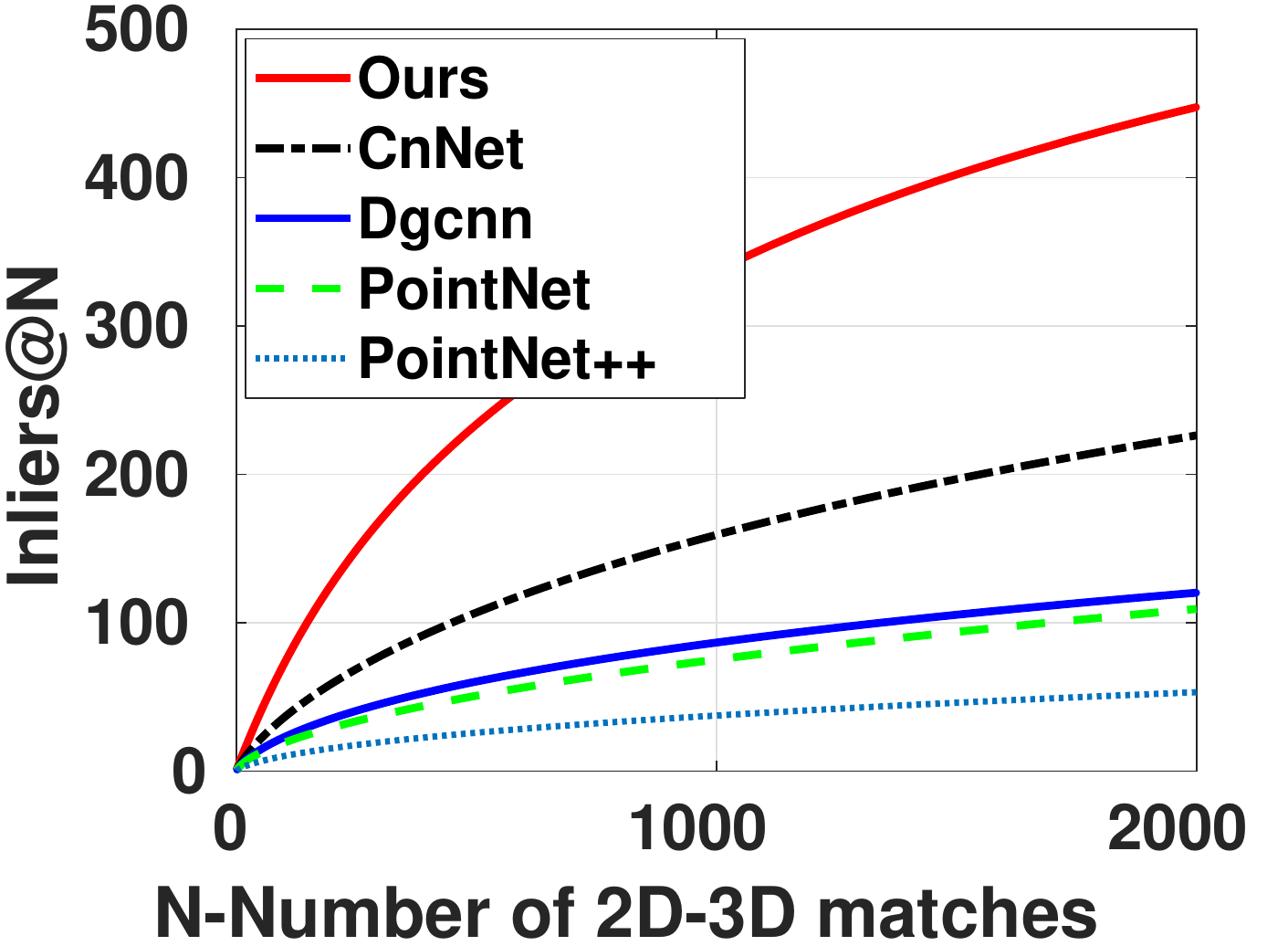}~~~~~~~~~~~~~~~~~~
\includegraphics[width=0.45\textwidth]{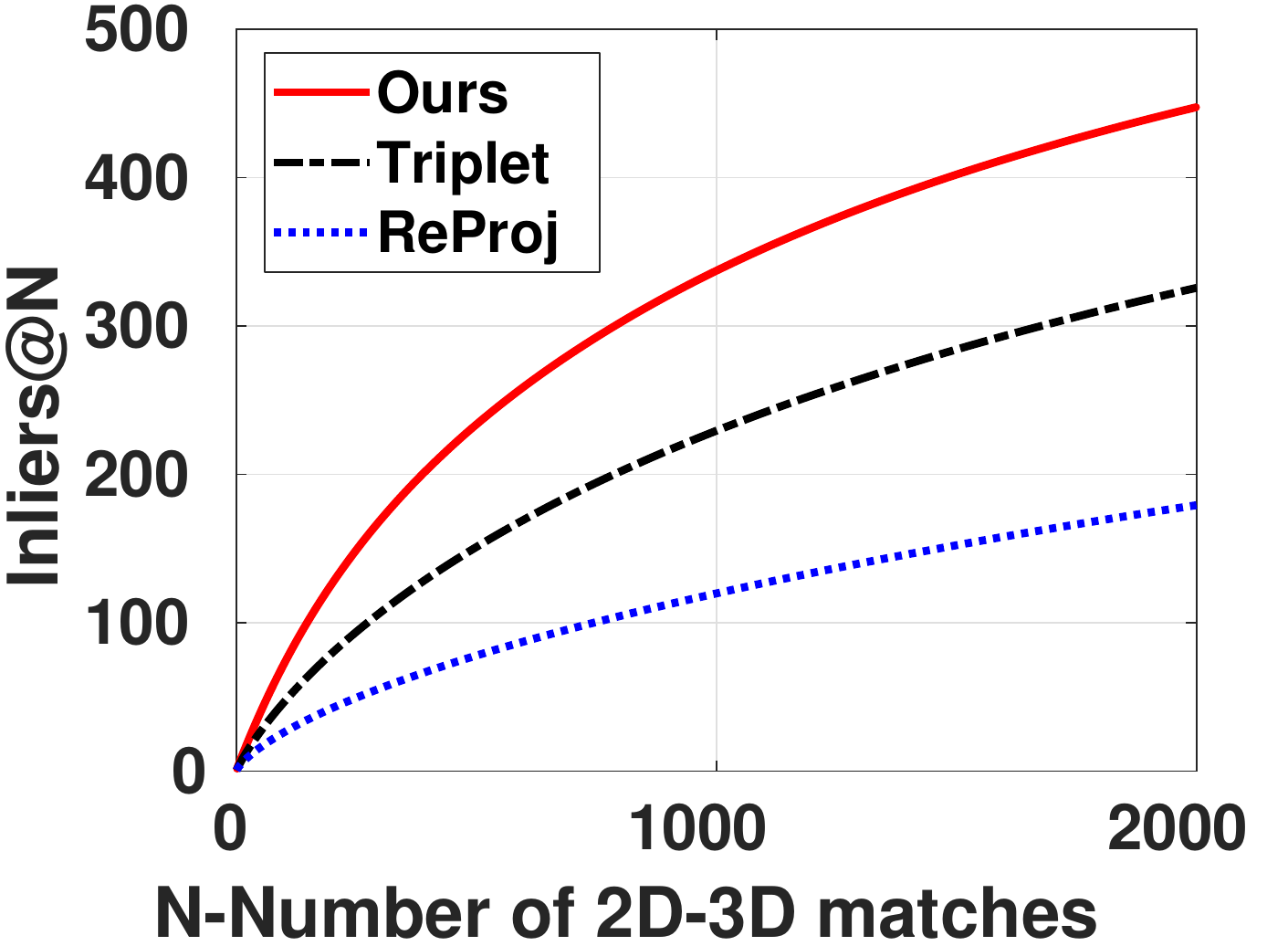}
\end{center}
% \vspace{-10pt}
\caption{Comparison of the average number of inliers with respect to the number of found 2D--3D matches for different backbone networks (\textbf{Left}) and loss functions (\textbf{Right}) on the MegaDepth dataset.
% \vspace{15pt}
}
\label{fig:megaDepthinliers_backbones}
\end{figure}

\vspace{3pt}
\noindent\emph{Backbone networks:}
% \paragraph{Backbone networks:}
To demonstrate the effectiveness of our method at regressing point-wise descriptors, we compare four networks: PointNet \cite{qi2017pointnet}, PointNet++ \cite{qi2017pointnetplusplus},  CnNet \cite{yi2018learning} and Dgcnn \cite{wang2018dynamic}. The features from PointNet \cite{qi2017pointnet} and Dgcnn \cite{wang2018dynamic} are taken before global pooling and have dimension $1024$. CnNet \cite{yi2018learning} and PointNet++ \cite{qi2017pointnetplusplus} generate features with dimension $128$. For all networks, the output point-wise feature vectors are $L_2$ normalized to embed them in a metric space. The configuration details are given in the Appendix. 
We compute the average number of inliers with respect to the number of found 2D--3D matches using each backbone network, as shown in \figref{fig:megaDepthinliers_backbones} (Left). It demonstrates that our feature extraction network significantly outperforms all other networks, finding more inlier 2D--3D matches. 

% The poor performance of PointNet++ \cite{qi2017pointnetplusplus} may rise from using pre-defined thresholds to sample and group points. 
% \vspace{-20pt}

% \vspace{-20pt}

\vspace{3pt}
\noindent\emph{Loss functions:}
% \paragraph{Loss functions:}
To show the effectiveness of our proposed inlier set probability maximization loss, we compare it with two others: (a) a reprojection loss; and (b) a triplet loss. For the triplet loss, we use an exhaustive mini-batch strategy to maximize the number of triplets.
% and build $2B(B-1)$ triplets within each batch. 
Using the weighting matrix $\bW$ learned using the different losses, we calculate the average number of inliers with respect to the number of found 2D--3D matches, shown in \figref{fig:megaDepthinliers_backbones} (Right). It shows that the proposed loss outperforms other losses, finding more inlier matches.

\section{Conclusion}
We have proposed a new deep network to solve the blind PnP problem of simultaneously estimating the 2D--3D correspondences and 6-DoF camera pose from 2D and 3D points.  The key idea is to extract discriminative point-wise feature descriptors that encode local geometric structure and global context, and use these to establish 2D--3D matches via a global feature matching module.
The high-quality correspondences found by our method facilitates the direct application of traditional PnP methods to recover the camera pose.  Our experiments demonstrate that our method significantly outperforms traditional geometry-based methods with respect to speed and accuracy.

\clearpage
\appendix
In this Appendix, we first give additional experimental results and then describe implementation details.

\section{Comparison with State-of-the-Art Methods}
In this section, we provide the full set of results for the synthetic ModelNet40 \cite{wu20153d} and NYU-RGBD \cite{Silberman:ECCV12} datasets.
We provide the 6 DoF pose estimation results in Table \ref{tab::rotation_trans_error_synthetic}. It shows that our method outperforms all state-of-the-art methods by a large margin.

\begin{table}[]
\setlength{\tabcolsep}{10pt}
% \footnotesize
\centering
\caption{Comparison of rotation and translation errors on the ModelNet40 and NYU-RGBD datasets.}
\begin{tabularx}{\linewidth}{llXXXXXX}
\hline
\multicolumn{2}{l}{\multirow{2}{*}{\backslashbox[50mm]{Method}{Dataset}}}      & \multicolumn{3}{c}{ModelNet40} & \multicolumn{3}{c}{NYU-RGBD} \\ \cline{3-8} 
\multicolumn{2}{l}{}                       & Q1     & Med.     & Q3     & Q1      & Med.     & Q3     \\ \hline

\multirow{2}{*}{P3P-RANSAC}    & Rot. err. ($^{\circ}$)    & 90.82  & 138.6  & 164.8 & 40.11   & 99.26   & 154.0  \\ \
                         & Trans. err. & 0.433  & 1.147  & 3.077  & 0.827   & 1.295   & 2.023  \\ \hline
\multirow{2}{*}{SoftPOSIT \cite{david2004softposit}}    & Rot. err. ($^{\circ}$)    & 16.10  & 21.75  & 28.00 & 12.88   & 20.61   & 31.32  \\  
                         & Trans. err. & 0.332  & 0.488  & 0.719  & 0.646   & 0.935   & 1.299  \\ \hline
                         
\multirow{2}{*}{GOSMA \cite{campbell2019alignment}}    & Rot. err. ($^{\circ}$)    & 10.08  & 22.06  & 52.01 & 1.364   & 3.184   & 21.98  \\ 
                         & Trans. err. & 0.254  & 0.464  & 0.746  & 0.126   & 0.212   & 0.688  \\ \hline
                         
\multirow{2}{*}{\textbf{Our}} & Rot. err. ($^{\circ}$)    & \textbf{1.349}  & \textbf{2.356}   & \textbf{5.260}  & \textbf{0.202}   & \textbf{0.291}   & \textbf{0.407}  \\ 
                         & Trans. err. & \textbf{0.018} & \textbf{0.037}   & \textbf{0.070}  & \textbf{0.009}   & \textbf{0.014}   & \textbf{0.020}  \\ \hline

\end{tabularx}
\label{tab::rotation_trans_error_synthetic}
\end{table}

% The runtime of all compared methods is given in Table \ref{tab::runtime_stoa}. It shows that our method  is very fast, with a greater than $100\times$ speed-up compared with the other methods.

% \begin{table}[!h]
% \setlength{\tabcolsep}{2.8pt}
% \footnotesize
% \caption{Comparison of runtime (in seconds) on the ModelNet40 and NYU-RGBD datasets.}
% \begin{tabularx}{\linewidth}{llXXXXXX}
% \hline
% \multicolumn{2}{l}{\multirow{2}{*}{\backslashbox{Method}{Dataset}}} & \multicolumn{3}{c}{ModelNet40 }  & \multicolumn{3}{c}{NYU-RGBD} \\ \cline{3-8} 
% \multicolumn{2}{l}{}                  & Q1    & Med.    & Q3     & Q1     & Med.    & Q3     \\ \hline
% \multicolumn{2}{l}{P3P-RANSAC}             & 30.12 & 30.22 & 30.28 & 30.07  & 30.17  & 30.25  \\ 
% \multicolumn{2}{l}{SoftPOSIT \cite{david2004softposit}}             & 24.77 & 25.90 & 29.39 & 30.26  & 30.42  & 30.64  \\ 
% \multicolumn{2}{l}{GOSMA \cite{campbell2019alignment}}             & 32.03 & 34.06 & 36.58 & 39.26  & 42.10  & 45.52  \\ 
% \multicolumn{2}{l}{Our}               & \textbf{0.146} & \textbf{0.151}  & \textbf{0.160}  & \textbf{0.161}  & \textbf{0.165}  & \textbf{0.170}  \\ \hline
% \end{tabularx}
% \label{tab::runtime_stoa}
% \end{table}

\section{Visualization of the Weighting Matrix}

\begin{figure}[!h]
\centering
\includegraphics[width=0.44\columnwidth]{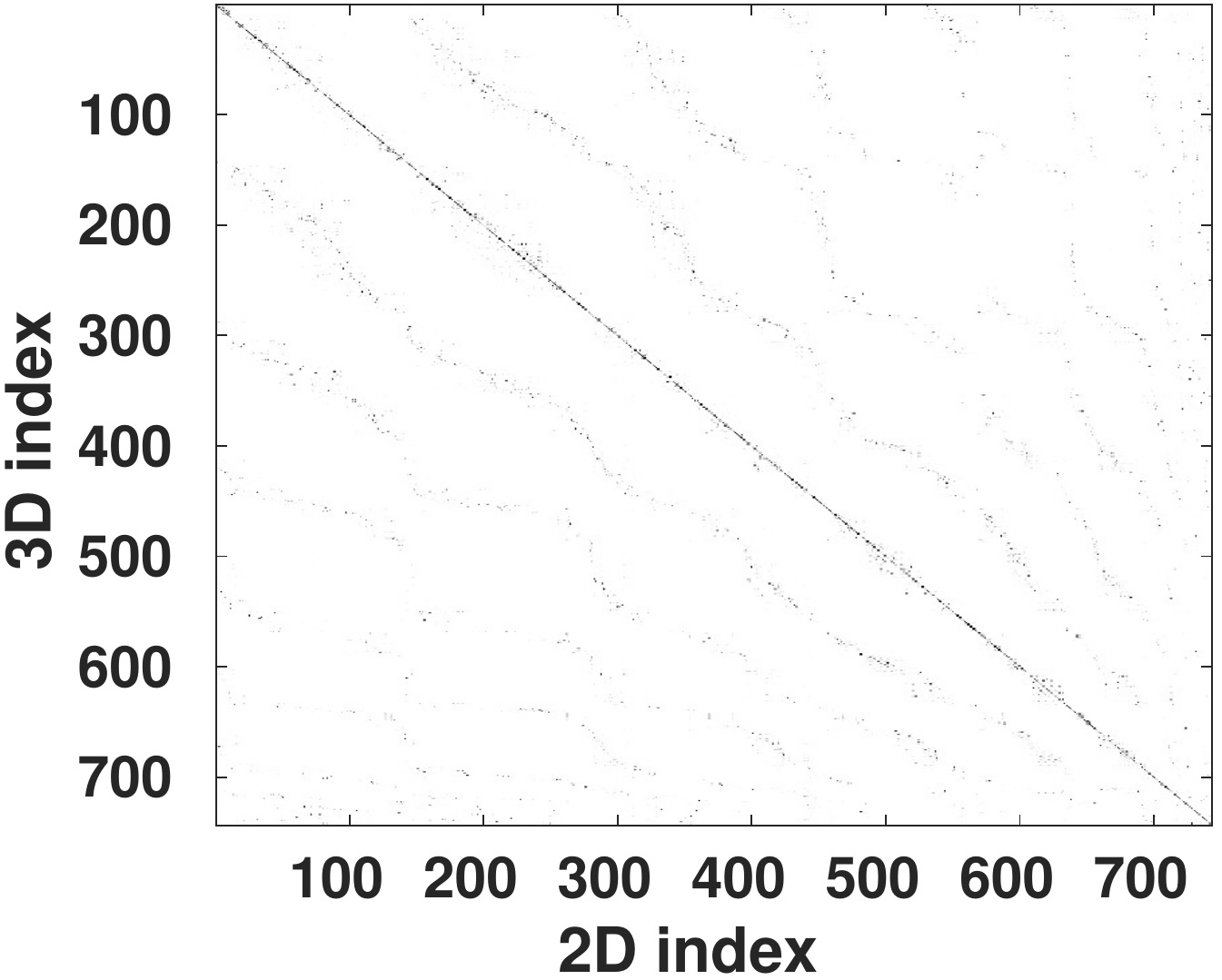}\hfill
\includegraphics[width=0.49\columnwidth]{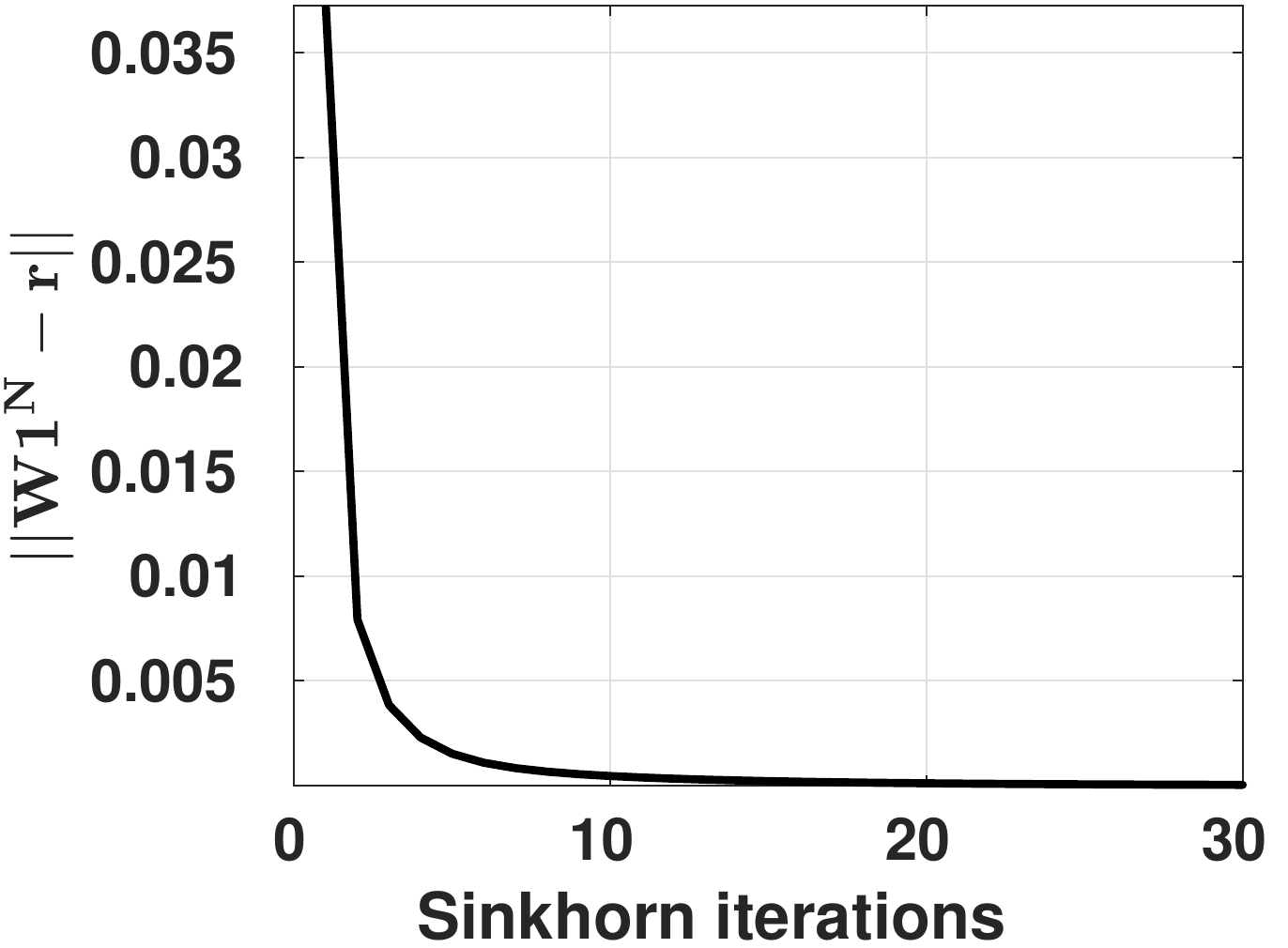}
\caption{Left: A sample weighting matrix $\bW$ from the test set of the MegaDepth dataset \cite{MegaDepthLi18}. For visualization, we arrange ground-truth 2D--3D matches in a regularized sequential order (\ie, the $k$-th 2D point matches the $k$-th 3D point). Ground-truth 2D--3D matches thus lie on the diagonal. Right: The convergence of $||\bW\mathbf{1}^N - \br||$ with respect to the Sinkhorn iteration number.
}
\label{fig:sinkhorn_interations}
\end{figure}

We provide a sample visualization of the weighting matrix $\bW$ in \figref{fig:sinkhorn_interations} (left) with ground-truth 2D--3D matches on the diagonal. The corresponding convergence curve of the Sinkhorn algorithm is given in \figref{fig:sinkhorn_interations} (right).

\section{Sample 3D and 2D Point Cloud}

\begin{figure}
\begin{center}
\includegraphics[width=0.25\textwidth]{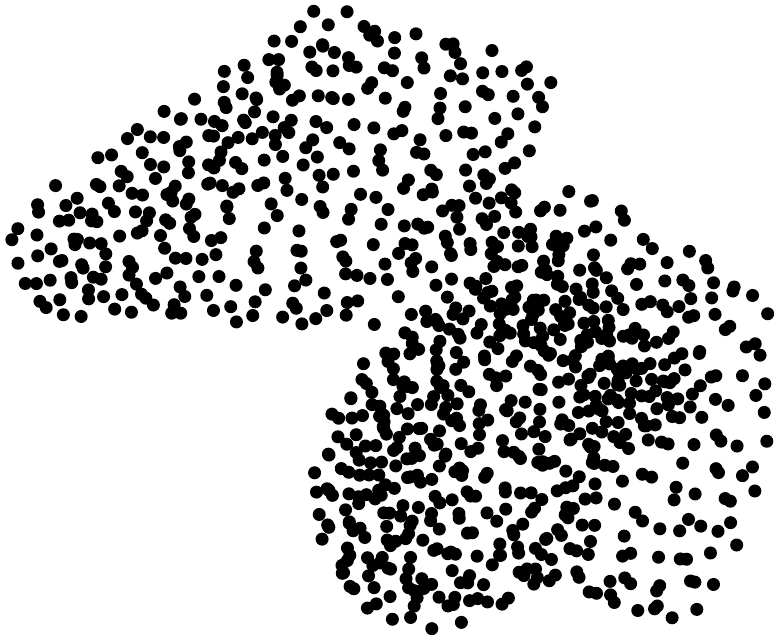}\hfill
\includegraphics[width=0.25\textwidth]{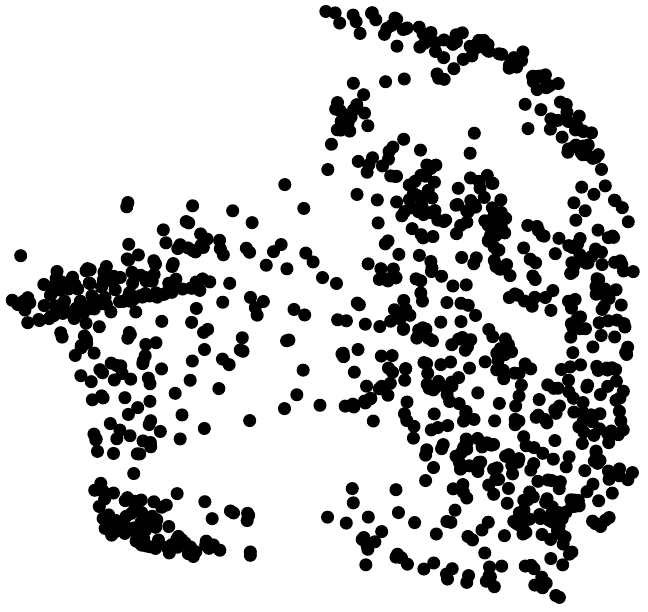}\hfill
\includegraphics[width=0.25\textwidth]{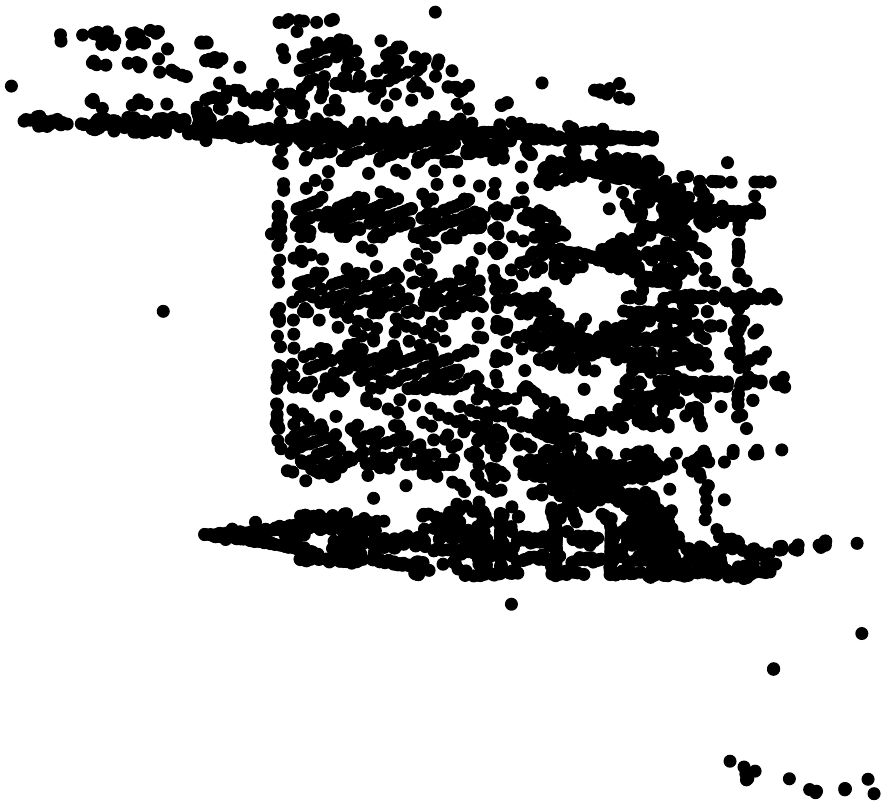}

\includegraphics[width=0.25\textwidth]{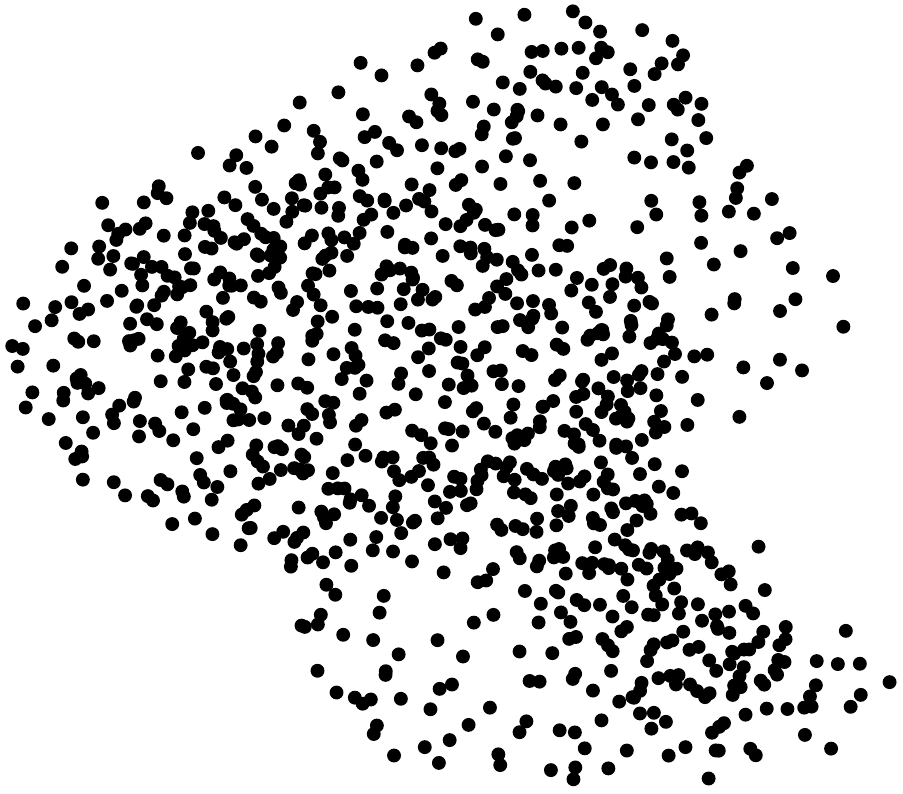}\hfill
\includegraphics[width=0.25\textwidth]{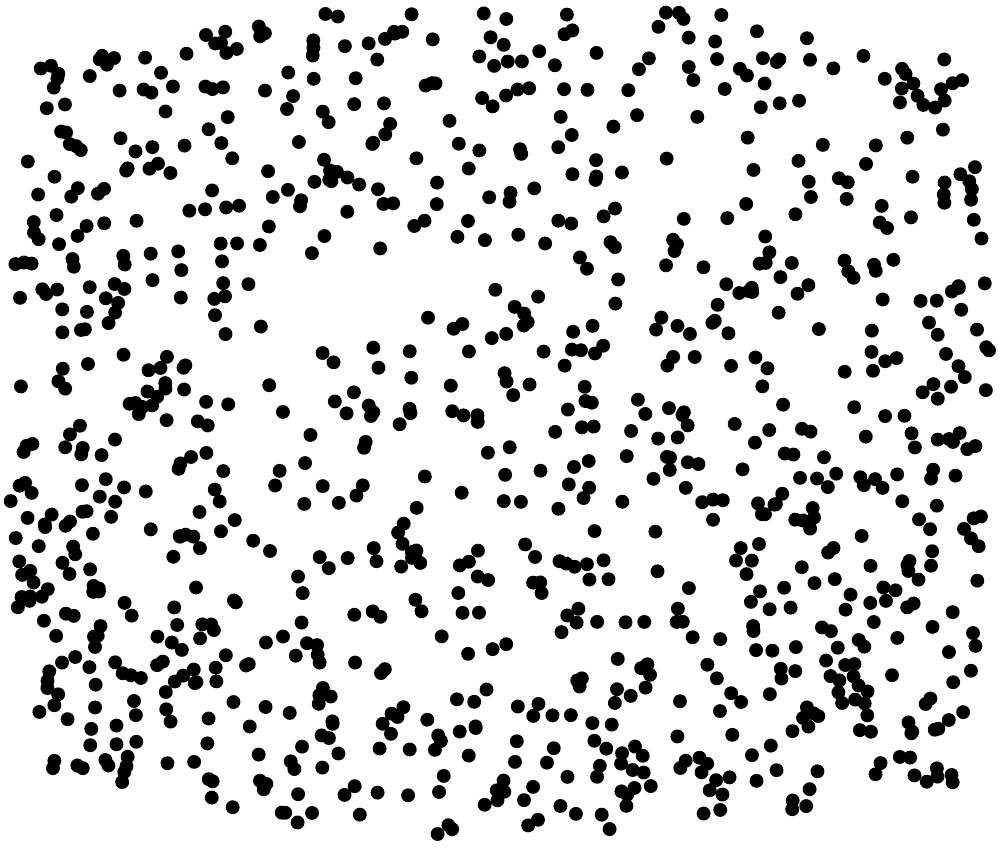}\hfill
\includegraphics[width=0.25\textwidth]{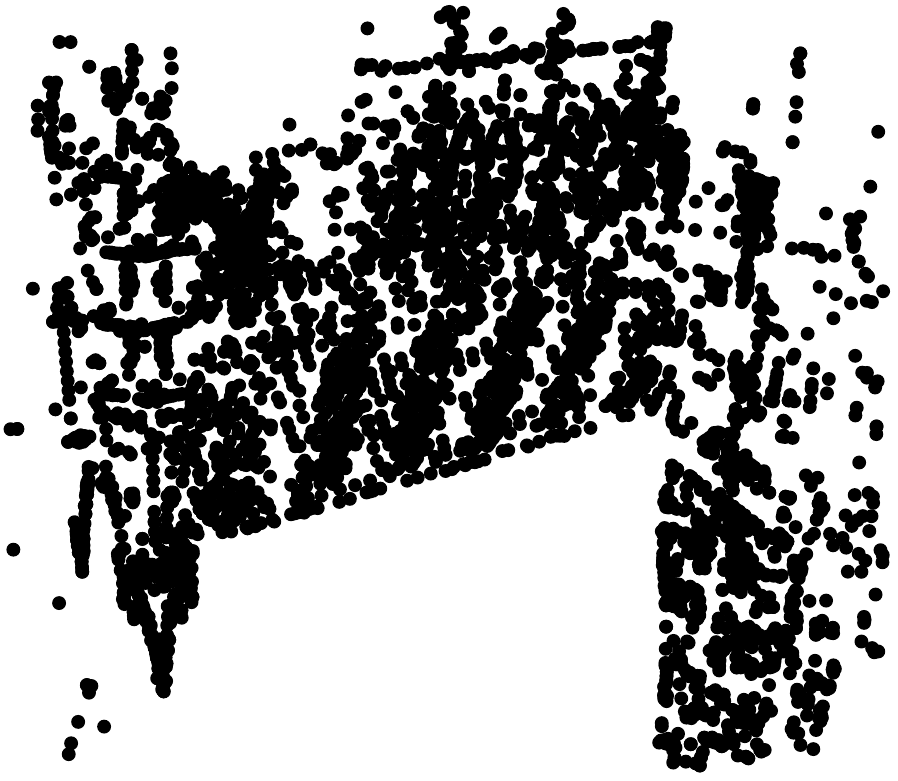}
\end{center}
\vspace{-10pt}
\caption{Sample 3D (top row) and 2D (bottom row) points cloud from ModelNet40 (\textbf{Left}) and NYU-RGBD (\textbf{Middle}) and real-world MegaDepth (\textbf{Right}) datasets.
}
\label{fig:sample_3d_2d_datasets}
\end{figure}

Our experiments are conducted on both synthetic (ModelNet40 \cite{wu20153d} and NYU-RGBD \cite{Silberman:ECCV12}) and real-world (MegaDepth \cite{MegaDepthLi18}) datasets. 
Figure \ref{fig:sample_3d_2d_datasets} presents sample 3D and 2D point clouds from these datasets.

\section{More Qualitative Results}
\begin{figure}
\begin{center}
\includegraphics[width=0.19\textwidth]{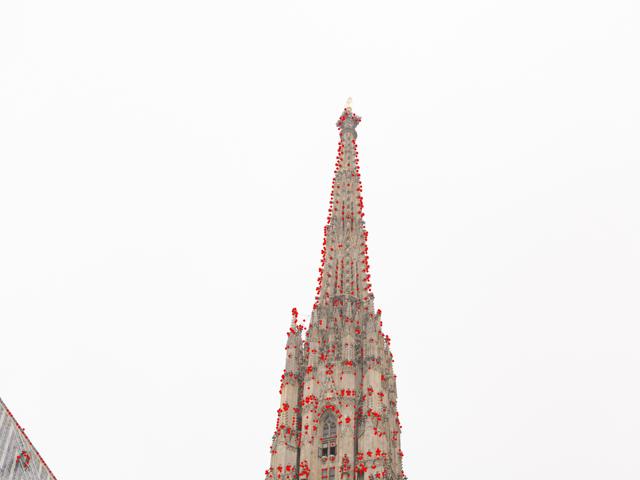}
\hfill
\includegraphics[width=0.19\textwidth]{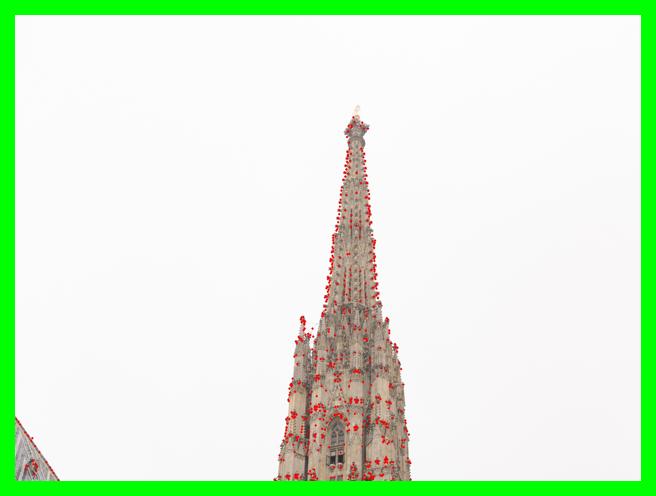}
\hfill
\includegraphics[width=0.19\textwidth]{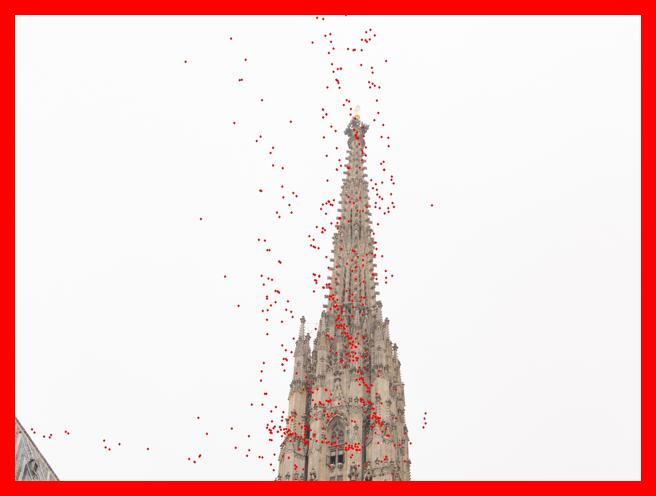}
\hfill
\includegraphics[width=0.19\textwidth]{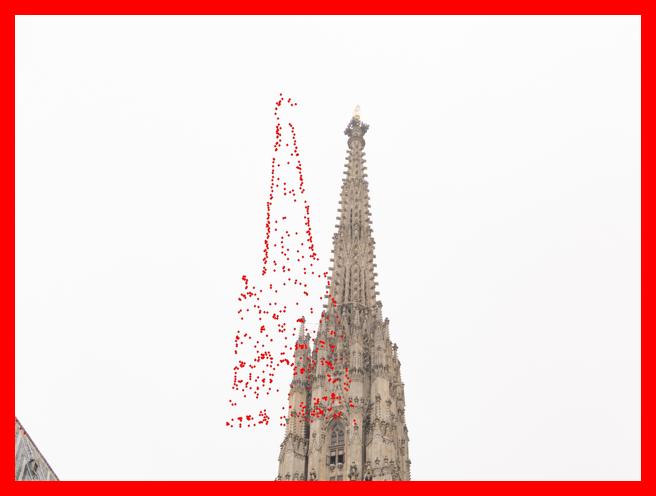}
\hfill
\includegraphics[width=0.19\textwidth]{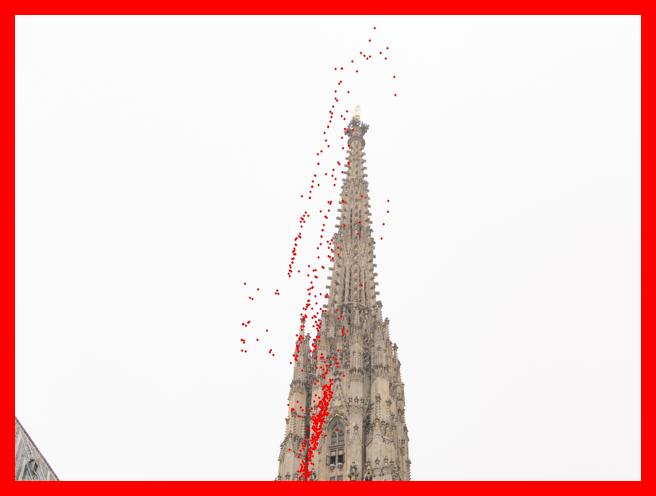}

\includegraphics[width=0.19\textwidth]{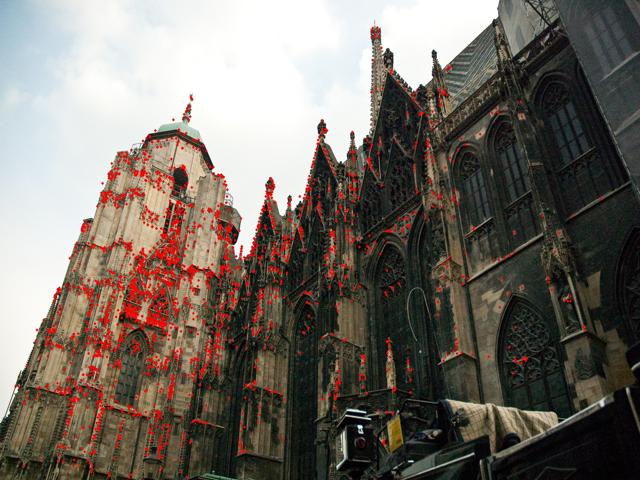}
\hfill
\includegraphics[width=0.19\textwidth]{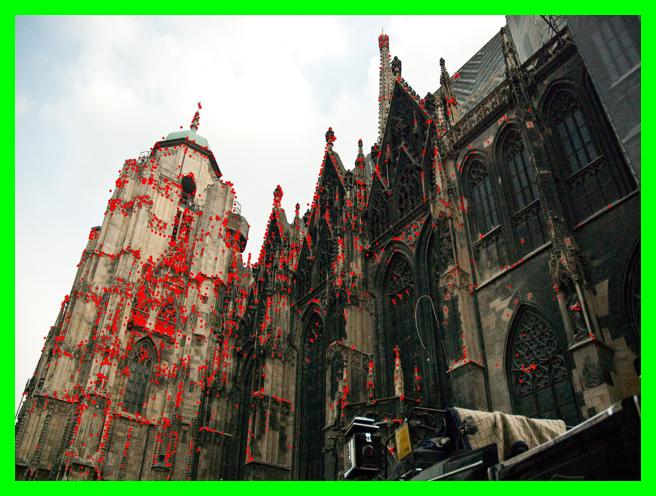}
\hfill
\includegraphics[width=0.19\textwidth]{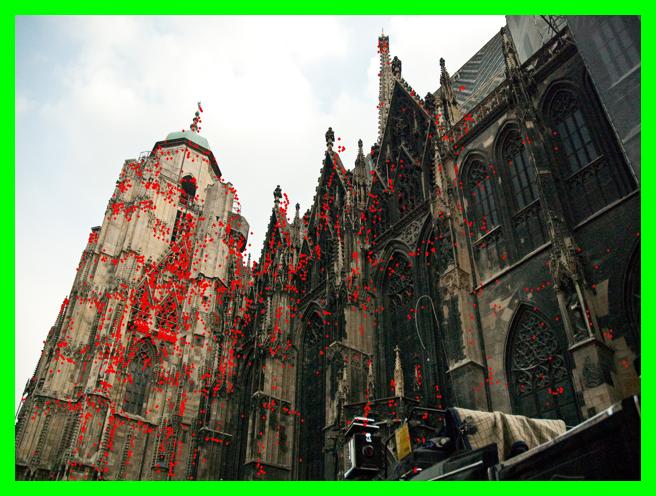}
\hfill
\includegraphics[width=0.19\textwidth]{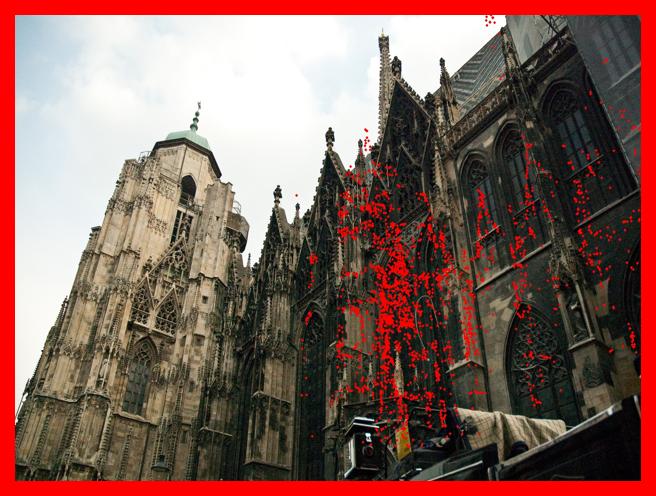}
\hfill
\includegraphics[width=0.19\textwidth]{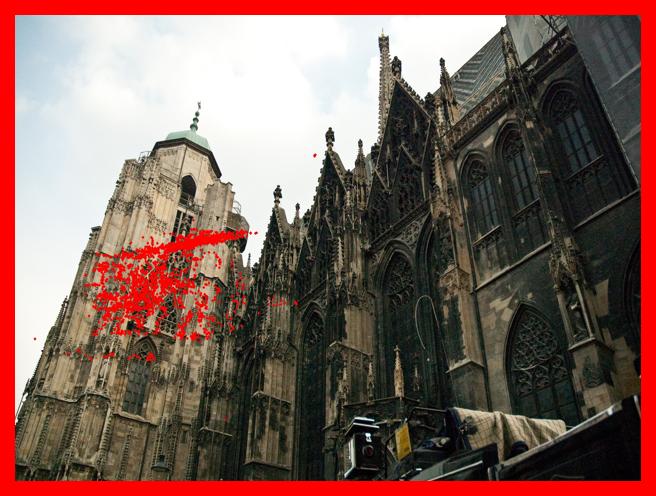}

\includegraphics[width=0.19\textwidth]{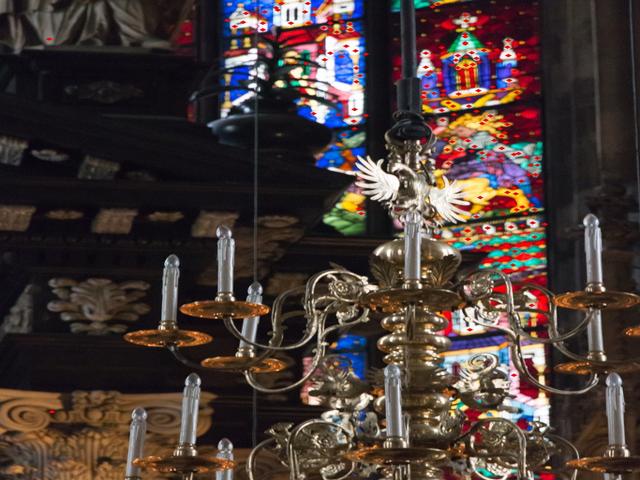}
\hfill
\includegraphics[width=0.19\textwidth]{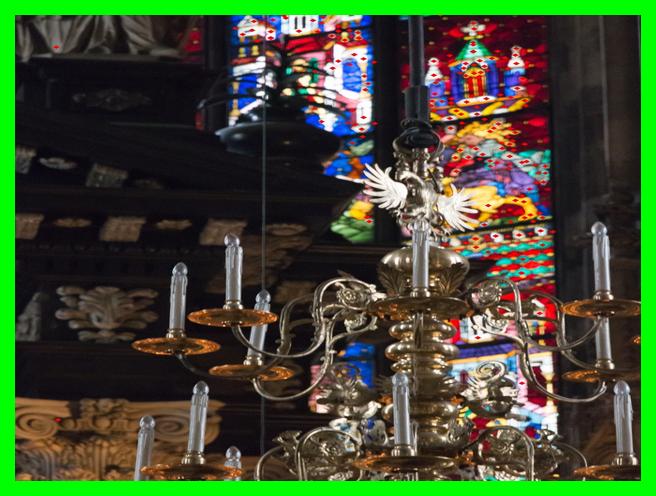}
\hfill
\includegraphics[width=0.19\textwidth]{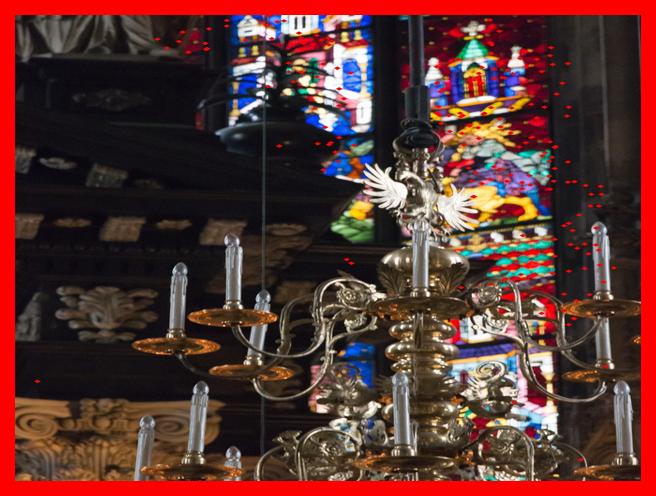}
\hfill
\includegraphics[width=0.19\textwidth]{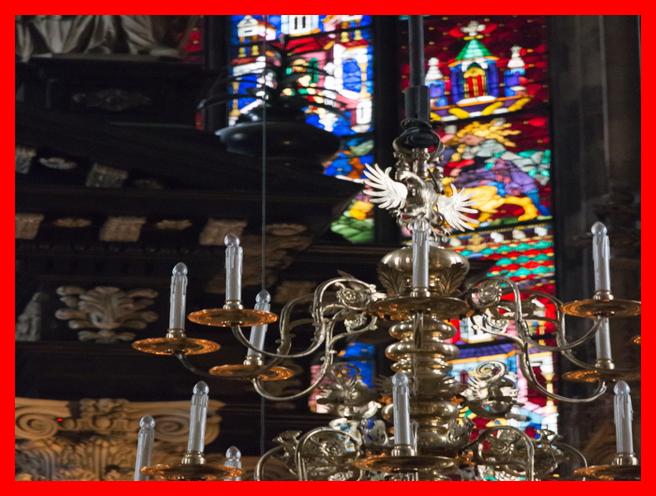}
\hfill
\includegraphics[width=0.19\textwidth]{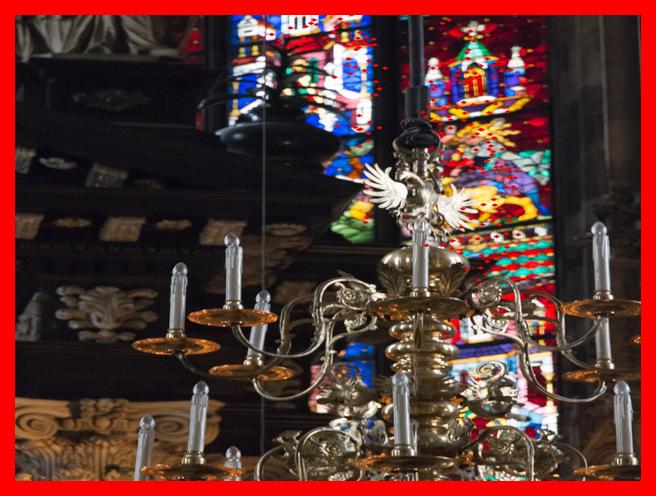}

\includegraphics[width=0.19\textwidth]{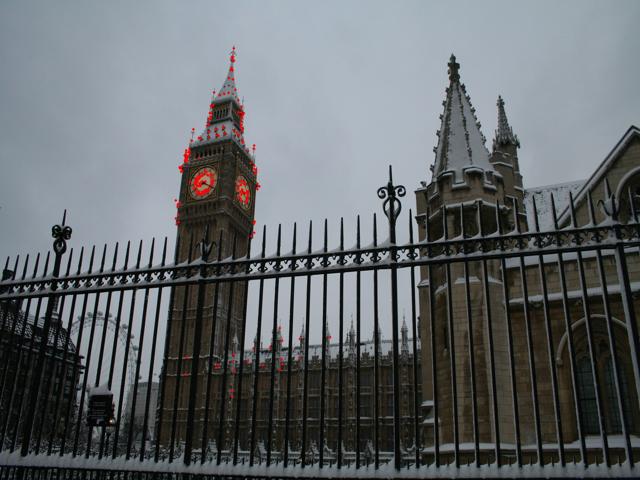}
\hfill
\includegraphics[width=0.19\textwidth]{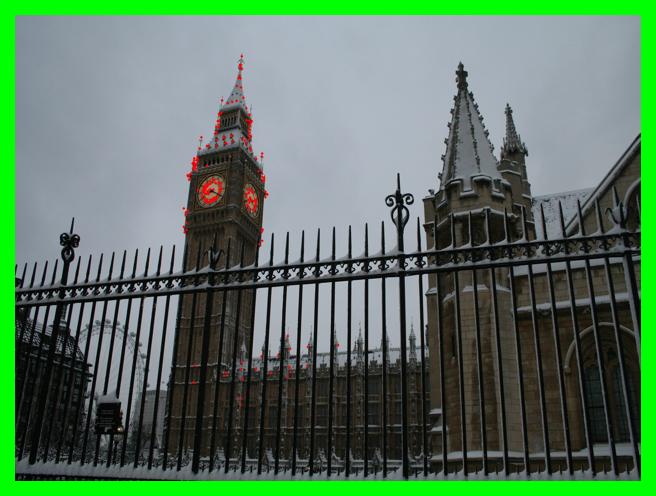}
\hfill
\includegraphics[width=0.19\textwidth]{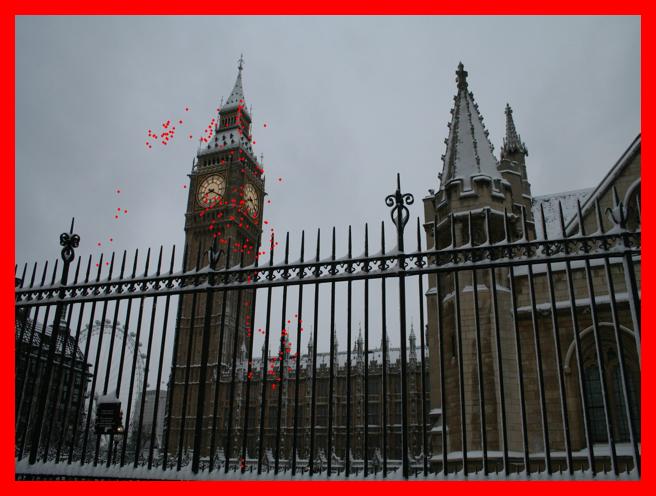}
\hfill
\includegraphics[width=0.19\textwidth]{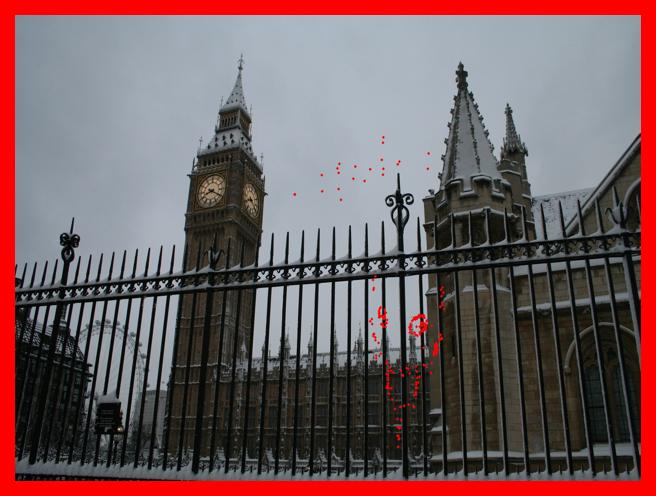}
\hfill
\includegraphics[width=0.19\textwidth]{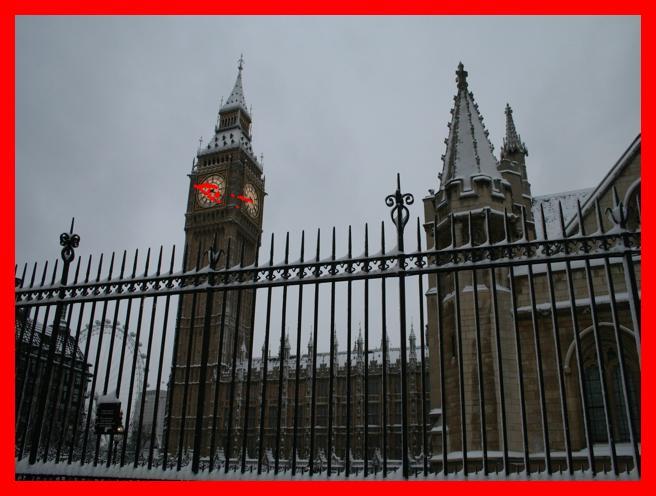}

\includegraphics[width=0.19\textwidth]{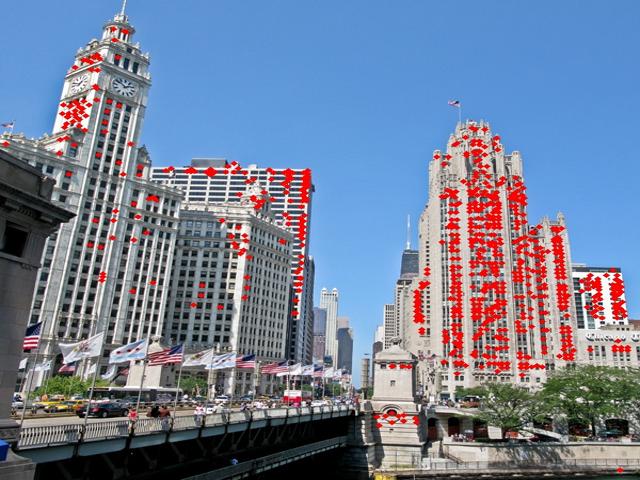}
\hfill
\includegraphics[width=0.19\textwidth]{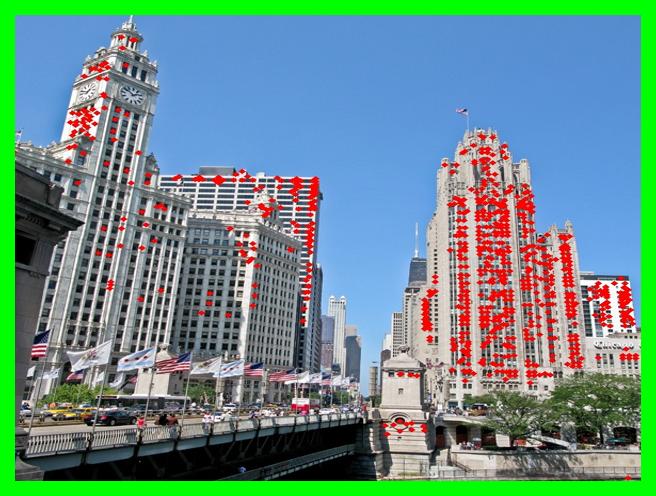}
\hfill
\includegraphics[width=0.19\textwidth]{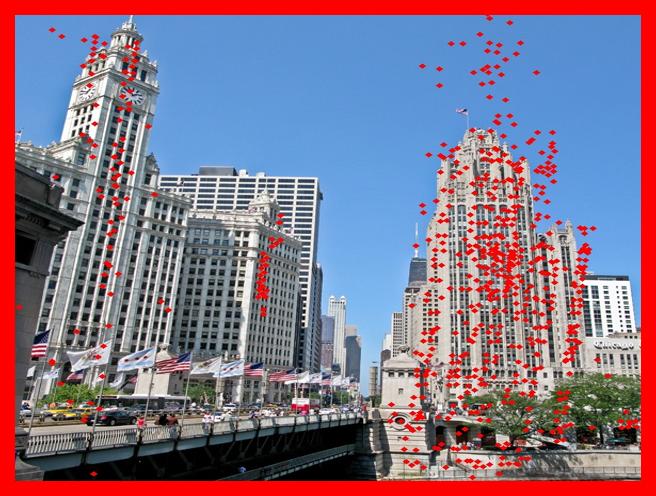}
\hfill
\includegraphics[width=0.19\textwidth]{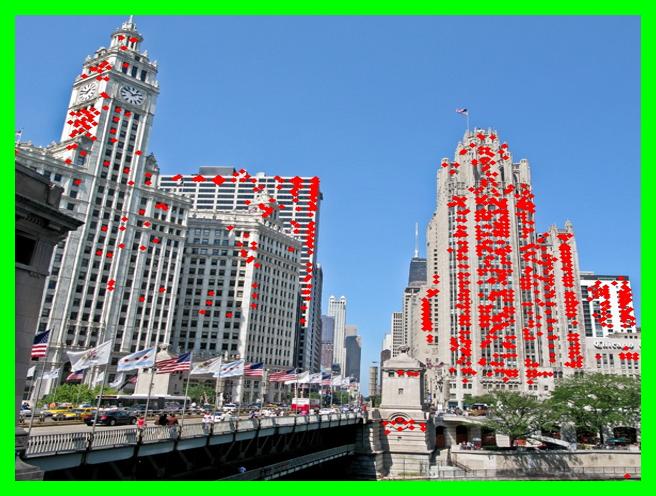}
\hfill
\includegraphics[width=0.19\textwidth]{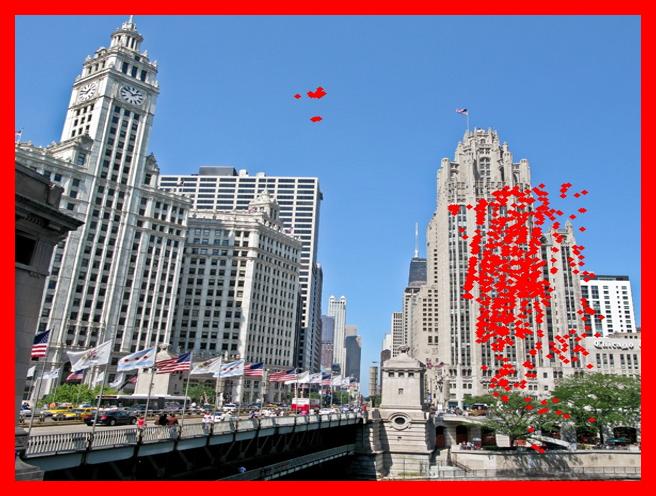}

\includegraphics[width=0.19\textwidth]{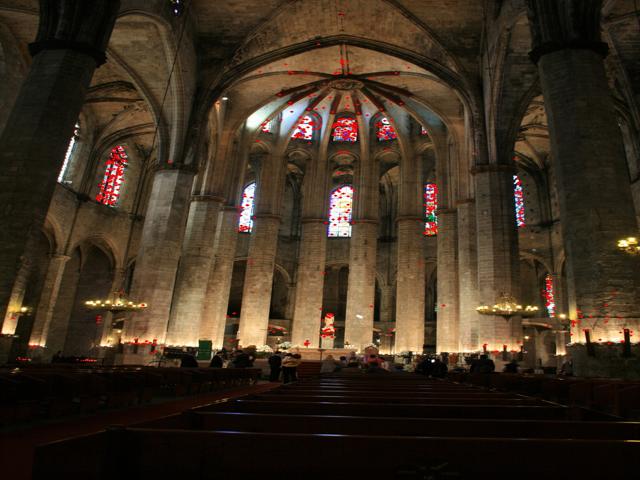}
\hfill
\includegraphics[width=0.19\textwidth]{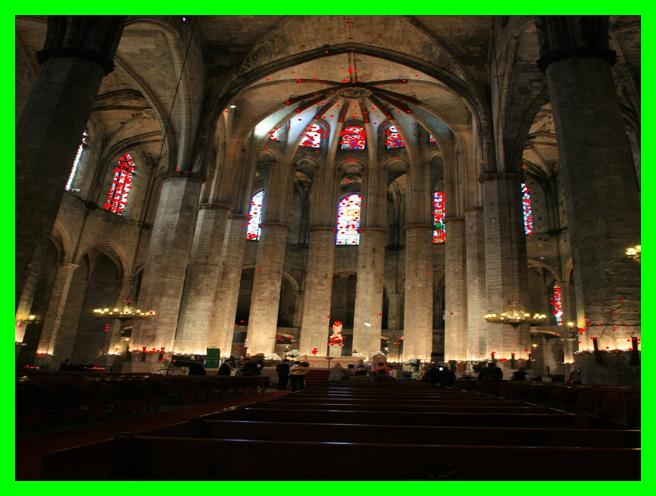}
\hfill
\includegraphics[width=0.19\textwidth]{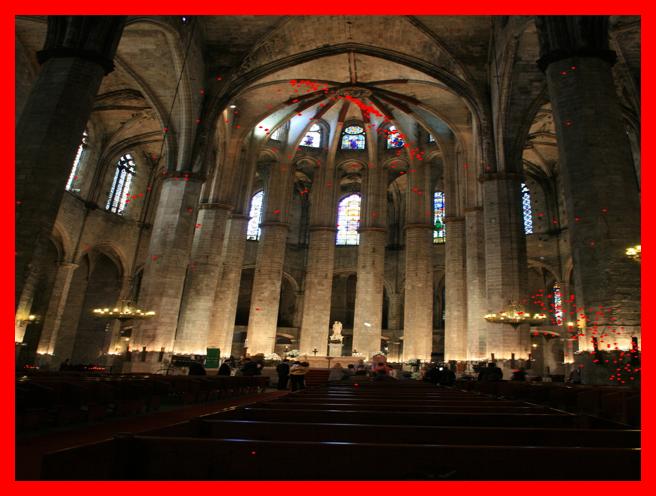}
\hfill
\includegraphics[width=0.19\textwidth]{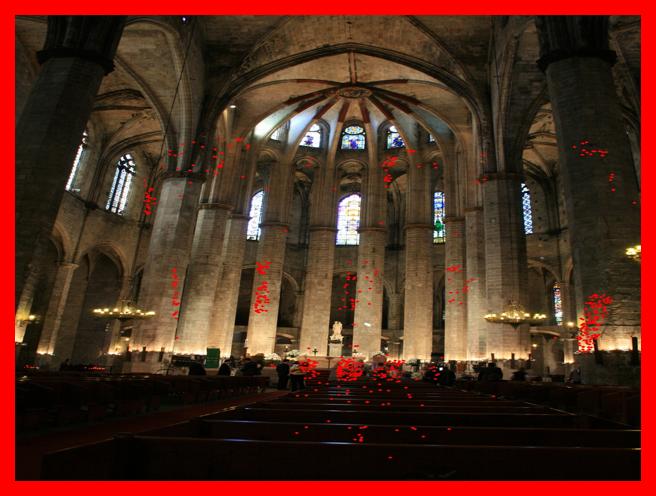}
\hfill
\includegraphics[width=0.19\textwidth]{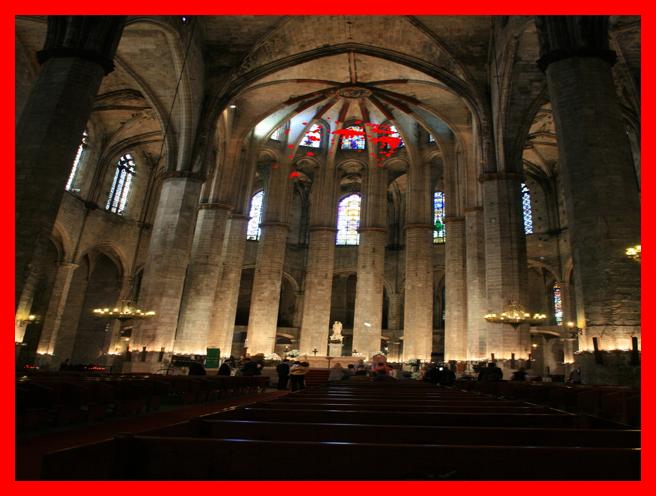}

\includegraphics[width=0.19\textwidth]{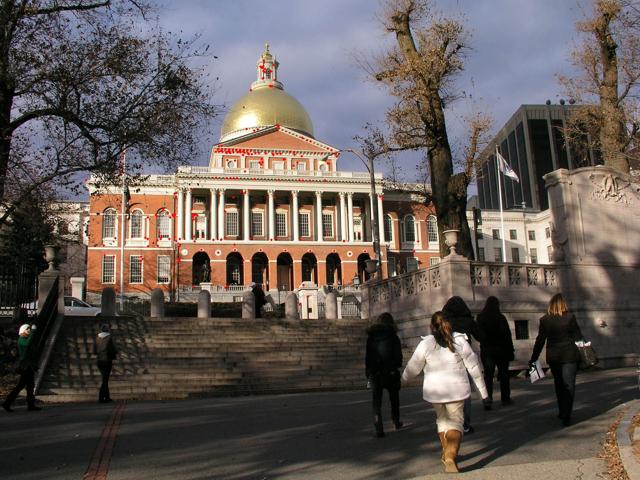}
\hfill
\includegraphics[width=0.19\textwidth]{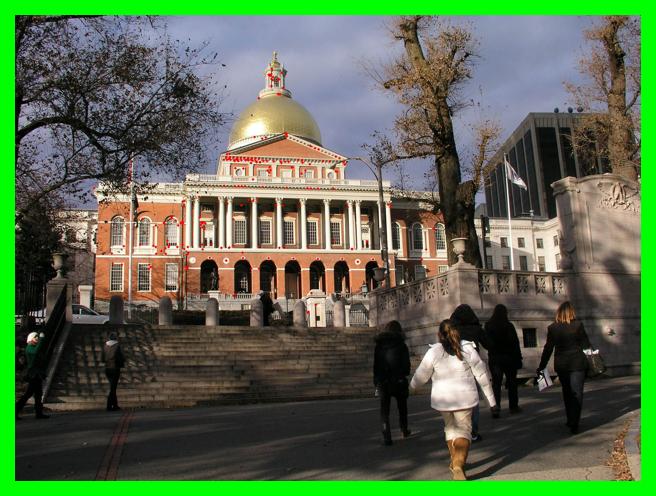}
\hfill
\includegraphics[width=0.19\textwidth]{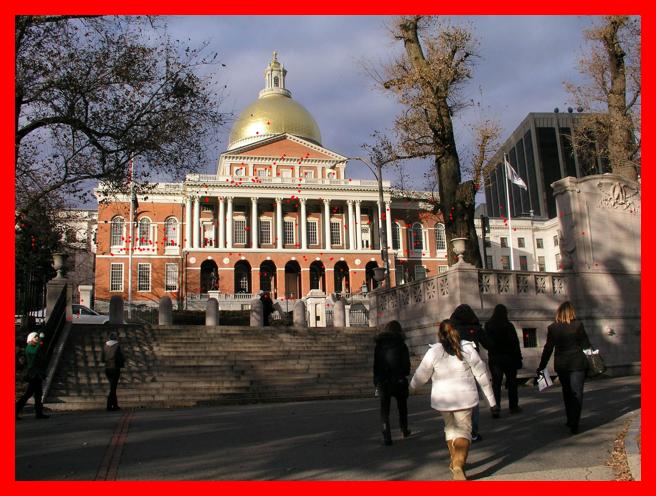}
\hfill
\includegraphics[width=0.19\textwidth]{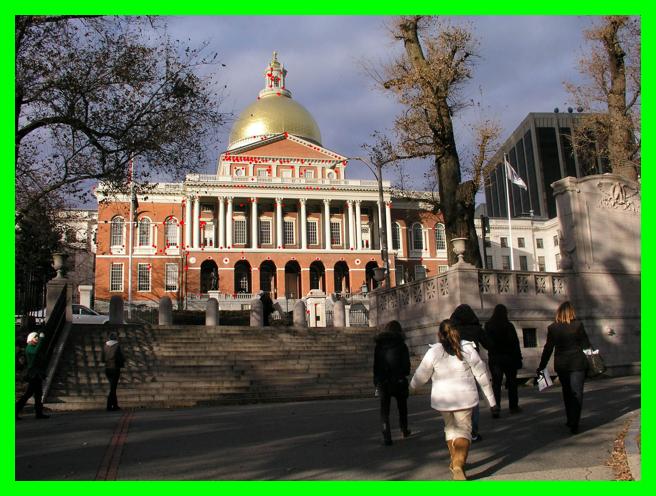}
\hfill
\includegraphics[width=0.19\textwidth]{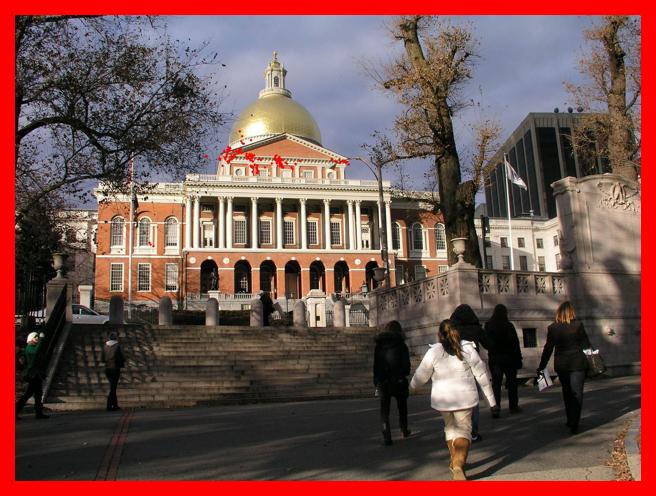}

\begin{subfigure}[t]{0.19\textwidth}
\includegraphics[width=\textwidth]{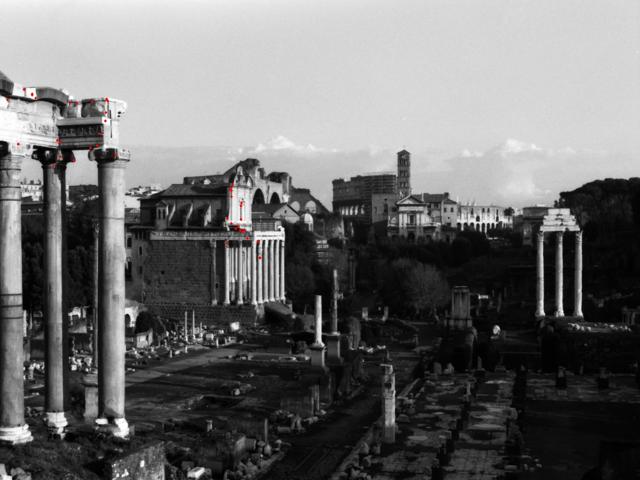}
\caption{GT}
\end{subfigure}
\hfill
\begin{subfigure}[t]{0.19\textwidth}
\includegraphics[width=\textwidth]{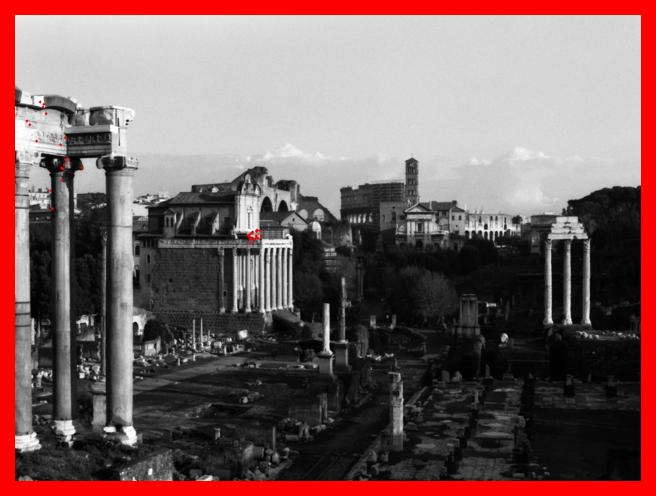}
\caption{Our}
\end{subfigure}
\hfill
\begin{subfigure}[t]{0.19\textwidth}
\includegraphics[width=\textwidth]{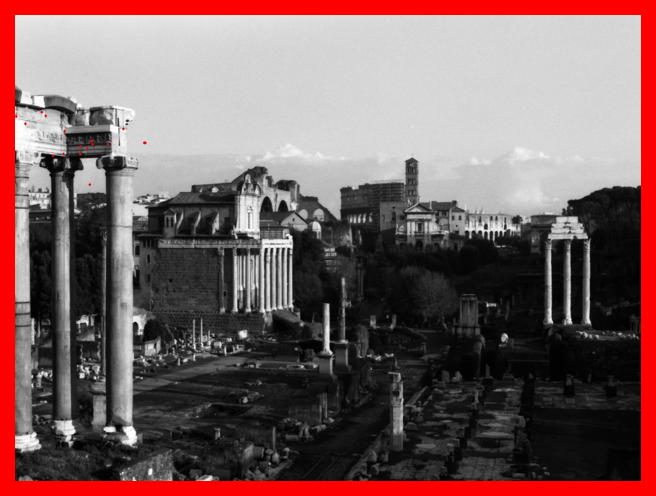}
\caption{GOSMA}
\end{subfigure}
\hfill
\begin{subfigure}[t]{0.19\textwidth}
\includegraphics[width=\textwidth]{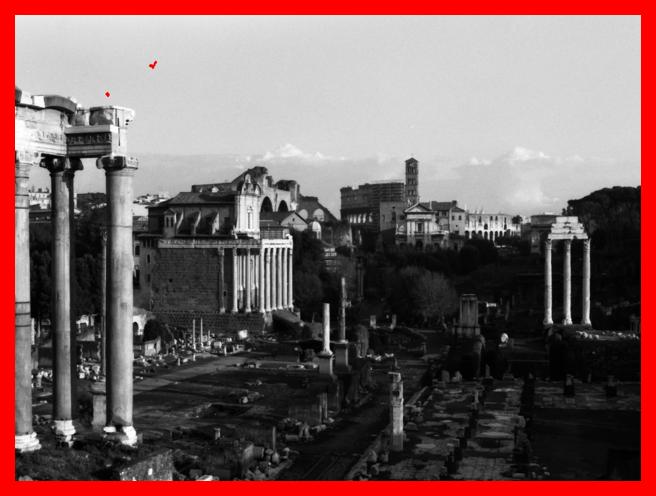}
\caption{SoftPOSIT}
\end{subfigure}
\hfill
\begin{subfigure}[t]{0.19\textwidth}
\includegraphics[width=\textwidth]{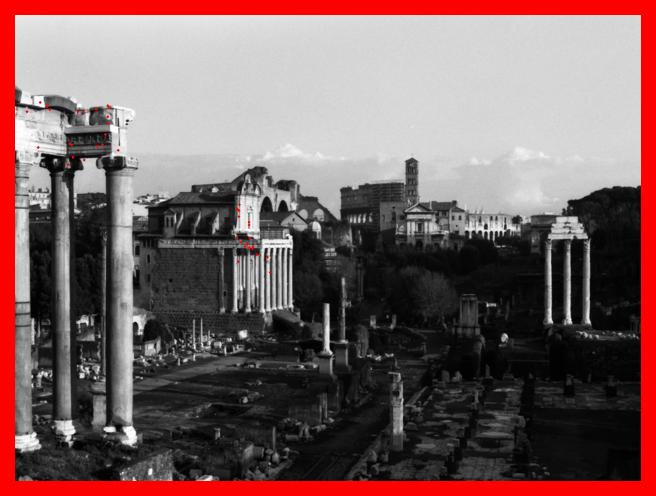}
\caption{ \scriptsize P3P-RANSAC}
\end{subfigure}

% \includegraphics[width=0.18\textwidth]{Figures/supply_figures/fgt_img_7000.jpg}

% \includegraphics[width=0.18\textwidth]{Figures/supply_figures/four_img_7000.jpg}
% \hfill
% \includegraphics[width=0.18\textwidth]{Figures/supply_figures/fgosma_img_7000.jpg}
% \hfill
% \includegraphics[width=0.18\textwidth]{Figures/supply_figures/fsoftposit_img_7000.jpg}
% \hfill
% \includegraphics[width=0.18\textwidth]{Figures/supply_figures/fransac_img_7000.jpg}

\end{center}
\caption{Qualitative comparison with state-of-the-art methods on the MegaDepth dataset, showing the projection of 3D points onto images using poses estimated by different methods. {\color{green} Green} border indicates the rotation/translation error of the estimated pose is less than  $5^\circ$/$0.5$ while {\color{red} red} border indicates the rotation/translation error of the estimated pose is larger than  $5^\circ$/$0.5$. Our method found more correct poses. The indices of these images on the MegaDepth testing dataset are $1$, $1000$, $2000$, $3000$, $4000$, $5000$, $6000$ and $7000$ from top to down. (best viewed in color).
}
\label{fig:Qualitative_cmp_supply}
\end{figure}

We give more qualitative results on the real-world MegaDepth \cite{MegaDepthLi18} dataset in Figure \ref{fig:Qualitative_cmp_supply}. We add green borders to images if their corresponding poses are estimated with rotation errors less than $5^\circ$ and translation errors less than 0.5. We add red borders to images if their corresponding poses are estimated with rotation errors larger than $5^\circ$ or translation errors larger than 0.5.

\section{Robustness to outliers}

In the main paper, to demonstrate the effectiveness of our method at handling outliers, we add outliers to both the 3D and 2D point-sets at the same time.
To further test the robustness of our method to real-world outliers, we add real-world outliers to the 3D and 2D point-sets separately. 

Specifically, for original 3D and 2D point-sets with cardinality $M$ and $N$, we add $\nu_{3D} M$ and $\nu_{2D} N$ outliers to the 3D and 2D point-sets, respectively, for outlier ratios $\nu_{3D} \in [0,1]$ and $\nu_{2D} \in [0,1]$. Note that the configurations of the main paper correspondences to $\nu_{3D}=\nu_{2D}$.

\begin{figure}
\begin{center}
\includegraphics[width=0.45\textwidth]{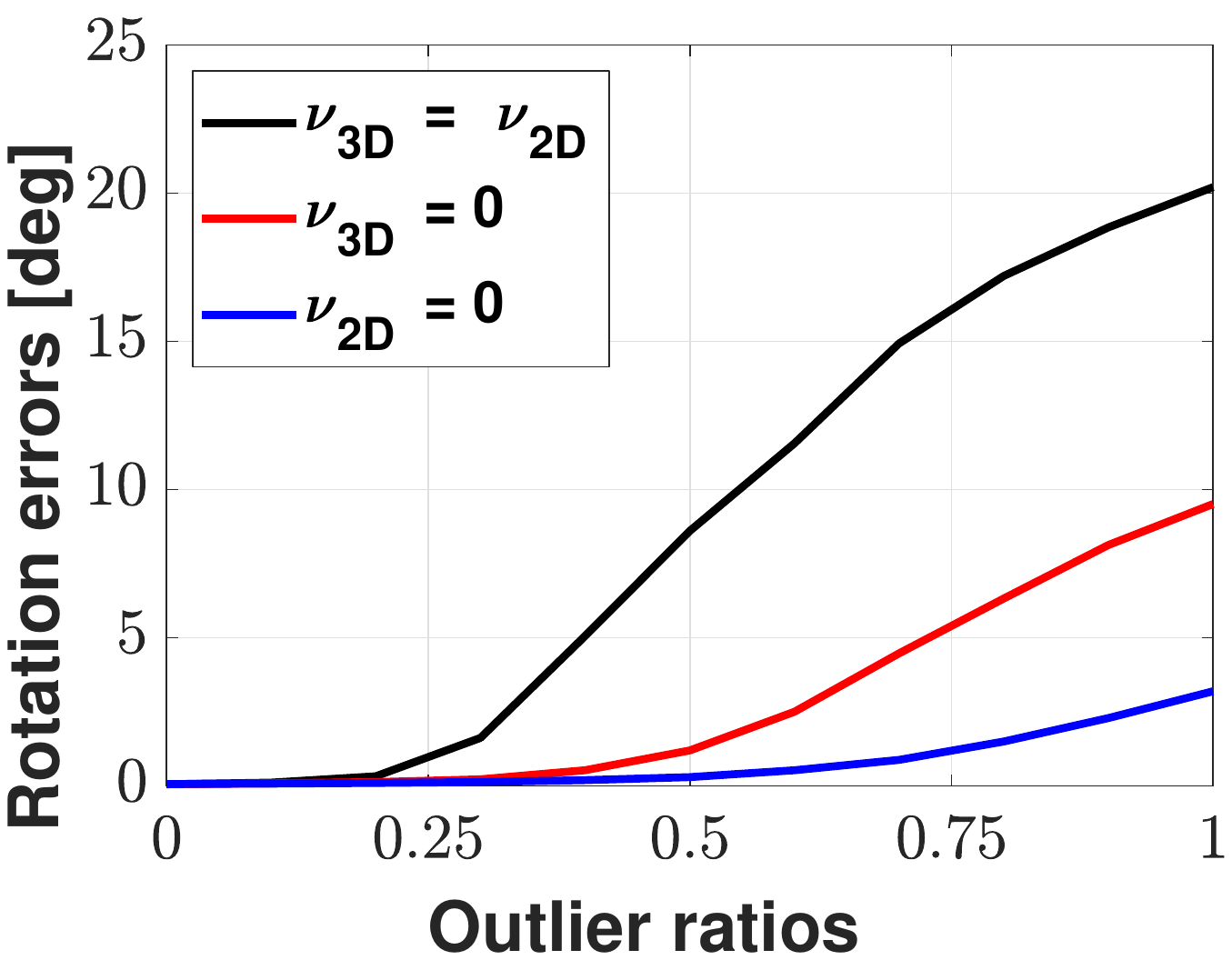}\hfill
\includegraphics[width=0.45\textwidth]{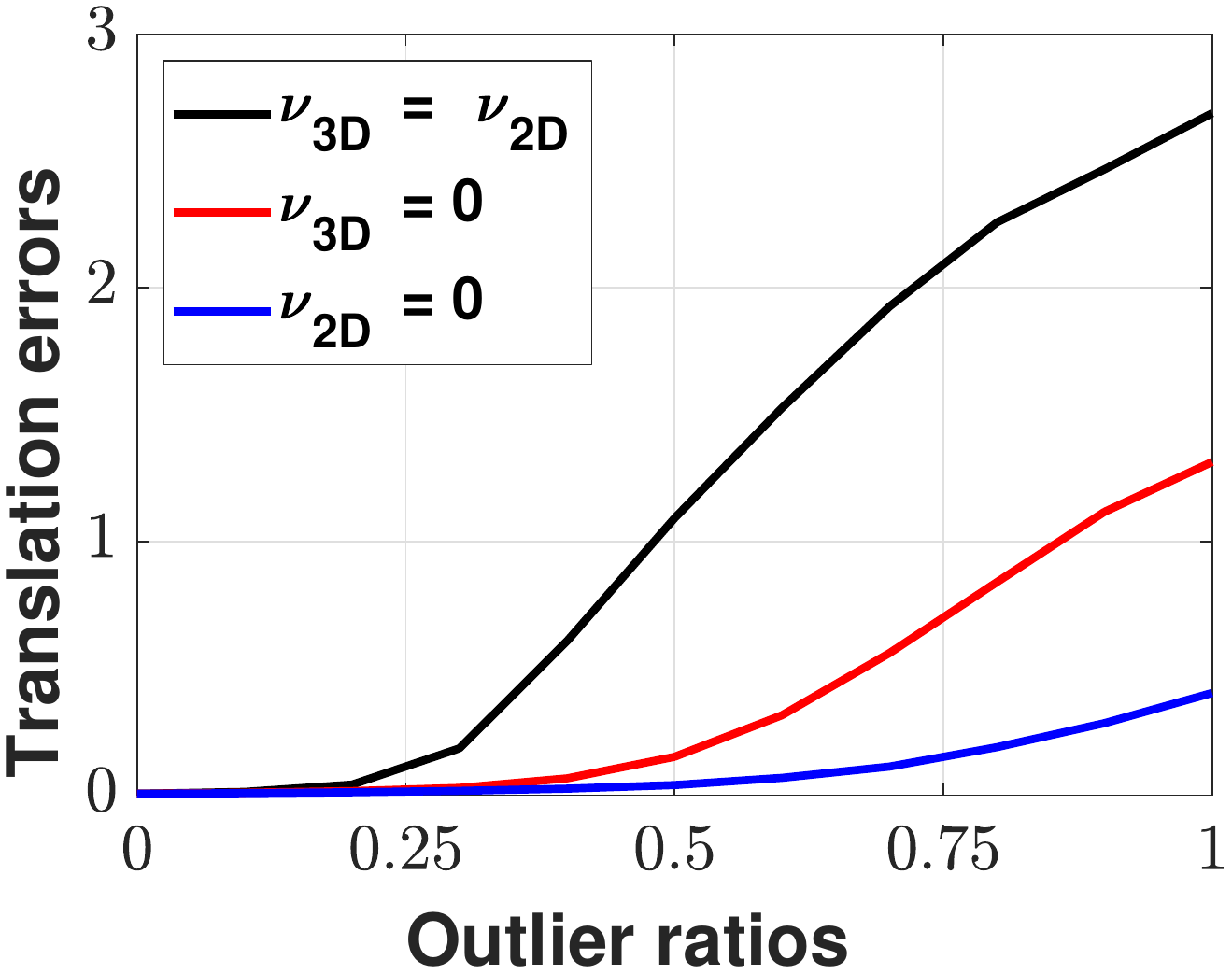}
\end{center}
\caption{Robustness to outliers on the MegaDepth dataset. Median rotation (\textbf{Left}) and translation (\textbf{Right}) errors with respect to the outlier ratio.
}
\label{fig:outlier_suppy}
\end{figure}

We first set  $\nu_{2D} = 0$ (\ie, no outliers are added to 2D points), and test the robustness of our method to real-world 3D outliers.  We then set  $\nu_{3D} = 0$ (\ie, no outliers are added to 3D points), and test the robustness of our method to real-world 2D outliers. The results of rotation and translation errors with respect to the outlier ratio are given in \figref{fig:outlier_suppy}. We also include the results of the main paper for completeness.

It shows that the performance of our method degrades gracefully with respect to an increasing outlier ratio. Interestingly, our method is more robust to 3D outliers than 2D outliers. The potential reason is that geometry structure within 2D points cloud is more easily to be destroyed by outliers than geometry structure within 3D points cloud. Note that the perspective projection of pinehole camera does not preserve the Euclidean properties of 3D geometry structure.

% We add two types of outliers: synthetic and real. For synthetic outliers, they are generated uniformly within the bounding box enclosing the 3D and 2D point-sets. The rotation and translation errors with respect to the outlier ratio are given in \figref{fig:poses_outlier_ratios_synthetic} (Left). For real outliers, 2D outliers are added from detected SIFT keypoints that do not have a matchable 3D point, and 3D outliers are added from 3D model points that do not have a matchable 2D point. The rotation and translation errors with respect to the outlier ratio are given in \figref{fig:poses_outlier_ratios_synthetic} (Right).
% It shows that the performance of our method degrades gracefully with respect to an increasing outlier ratio. More results are given in supplementary material.

\section{Implementation Details}

\subsection{State-of-the-Art Methods}

\paragraph{GOSMA \cite{campbell2019alignment}:}
We provide a translation domain to the GOSMA algorithm in the following way, in order to reduce the search space for the algorithm:
(i) find the axis-aligned bounding box that includes all points in the 3D model excluding outliers (\ie excluding the 2.5\% percentile minimum and maximum); (ii) extend the bounding box to include the ground-truth camera position; and (iii) increase the size of the resulting bounding box by 10\%.
In this way, we ensure that the search space encompasses all reasonable camera positions. The runtime is set to a maximum of $30$s per alignment for time considerations.
As a result, the algorithm does not always converge to the global optimum or provide an optimality guarantee, which can require minutes per alignment and is therefore impractical for a large dataset.

\paragraph{SoftPOSIT \cite{david2004softposit}:}
SoftPOSIT requires an initial estimate of the rotation and translation. For the synthetic datasets ModelNet40 and NYU-RGBD, we use the mean translation and the average rotation to initialise SoftPOSIT. The median initial rotation error is $21.5^\circ$ and the median initial translation error is $0.49$. For the real-world dataset MegaDepth, we set the initial pose by adding a perturbation to the ground-truth poses. The angular perturbation on rotation is uniformly drawn from $[-10, 10]$ degrees, and the perturbation on translation is uniformly drawn from $[-0.5, 0.5]$. The number of iterations is set to $25$, which corresponds to $\sim\!30$s for $\sim\!1000$ 2D/3D points in our experiments.

\paragraph{P3P-RANSAC:}
We compare our method against P3P-RANSAC with randomly-sampled 2D--3D correspondences. For randomly-sampling 2D--3D correspondences in P3P-RANSAC, the probability of finding inlier 2D--3D correspondences approximates zero within $30s$.
The number of RANSAC iterations $k$ is given by:
\begin{equation}
    k = \log(1-p) / \log(1 - w^q)
\end{equation}
where $p$ is the confidence level, $w$ is the ground-truth inlier ratio of 2D--3D correspondences, and $q=4$ is the minimal number of 2D--3D correspondences for P3P (one more 2D--3D correspondence to prune multiple solutions of P3P).
For a moderate confidence $p=90\%$, the number of 2D and 3D points at $1000$ ($w = 1/1000$), the number of RANSAC iterations $k$ approximates $2.3\times10^{12}$.

Within $30s$, we can evaluate $8.7\times10^5$ hypotheses, resulting in a RANSAC success ratio at $3.8\times10^{-7}$.

\subsection{Backbone Network Architectures}
\paragraph{PointNet \cite{qi2017pointnet}:}
The architecture is cropped from the segmentation model. The input point cloud is passed to a transformation network to regress a $3\times3$ matrix. The matrix is applied to each point.
After this alignment stage, the point cloud is passed to an MLP ($64$,$64$) network (with  layer  output  sizes  $64$,  $64$)  for each point. The output features are then passed to a transformation network to regress a $64\times64$ matrix. The matrix is applied to each feature.
After the feature-alignment, features are passed to an MLP ($64$,$128$,$1024$) network. The output feature of each point is $L_2$ normalized to embed it to a metric space.

\paragraph{PointNet++ \cite{qi2017pointnetplusplus}:}
The architecture is cropped from the segmentation model. The input point cloud is passed to $4$ set abstraction modules (SA) and $4$ feature propagation layers (FP). The configuration of SAs are: SA ($1024$, $0.1$, [$32$,$32$,$64$]),  SA ($256$, $0.2$, [$64$,$64$,$128$]),  SA ($64$, $0.4$, [$128$,$128$,$256$]) and SA ($16$, $0.8$, [$256$,$256$,$512$]). The configuration of FPs are: FP ($256$,$256$), FP ($256$,$256$), FP ($256$,$128$) and FP ($128$,$128$,$128$).
The output feature of each point is $L_2$ normalized.

\paragraph{Dgcnn \cite{wang2018dynamic}:}
The architecture is cropped from the part segmentation model. The number of nearest neighbors is set to $10$. It contains $3$ MLP blocks. For the first MLP block, points cloud is passed to MLP ($64$,$64$) network (with layer output sizes $64$,  $64$) on each point. Local features are aggregated using max-pooling.
For the second MLP block, features are passed to MLP ($64$,$64$) network. Local features are aggregated using max-pooling. For the last MLP block, features are passed to MLP ($64$) network. Local features are aggregated using max-pooling.
The outputs of $3$ MLP blocks are concatenated and passed to MLP ($1024$) network. The output feature of each point is $L_2$ normalized.

\paragraph{CnNet \cite{yi2018learning}:}
The architecture is cropped before the ReLU+Tanh operation. The output feature of each point is $L_2$ normalized.
\clearpage
% \begin{appendices}
%   \chapter{Consectetur adipiscing elit}
%   \chapter{Mauris euismod}
% \end{appendices}

\bibliographystyle{splncs04}
\bibliography{egbib}

\begin{thebibliography}{10}
\providecommand{\url}[1]{\texttt{#1}}
\providecommand{\urlprefix}{URL }
\providecommand{\doi}[1]{https://doi.org/#1}

\bibitem{aoki2019pointnetlk}
Aoki, Y., Goforth, H., Srivatsan, R.A., Lucey, S.: Pointnetlk: Robust \&
  efficient point cloud registration using pointnet. In: Proceedings of the
  IEEE Conference on Computer Vision and Pattern Recognition. pp. 7163--7172
  (2019)

\bibitem{campbell2017globally}
Campbell, D., Petersson, L., Kneip, L., Li, H.: Globally-optimal inlier set
  maximisation for simultaneous camera pose and feature correspondence. In:
  Proceedings of the IEEE International Conference on Computer Vision. pp.
  1--10 (2017)

\bibitem{campbell2019alignment}
Campbell, D., Petersson, L., Kneip, L., Li, H., Gould, S.: The alignment of the
  spheres: Globally-optimal spherical mixture alignment for camera pose
  estimation. In: Proceedings of the IEEE Conference on Computer Vision and
  Pattern Recognition. pp. 11796--11806 (2019)

\bibitem{courty2016optimal}
Courty, N., Flamary, R., Tuia, D., Rakotomamonjy, A.: Optimal transport for
  domain adaptation. IEEE transactions on pattern analysis and machine
  intelligence  \textbf{39}(9),  1853--1865 (2016)

\bibitem{cuturi2013sinkhorn}
Cuturi, M.: Sinkhorn distances: Lightspeed computation of optimal transport.
  In: Advances in neural information processing systems. pp. 2292--2300 (2013)

\bibitem{dang2018eigendecomposition}
Dang, Z., Moo~Yi, K., Hu, Y., Wang, F., Fua, P., Salzmann, M.:
  Eigendecomposition-free training of deep networks with zero eigenvalue-based
  losses. In: Proceedings of the European Conference on Computer Vision (ECCV).
  pp. 768--783 (2018)

\bibitem{david2004softposit}
David, P., Dementhon, D., Duraiswami, R., Samet, H.: Softposit: Simultaneous
  pose and correspondence determination. International Journal of Computer
  Vision  \textbf{59}(3),  259--284 (2004)

\bibitem{fischler1981random}
Fischler, M.A., Bolles, R.C.: Random sample consensus: a paradigm for model
  fitting with applications to image analysis and automated cartography.
  Communications of the ACM  \textbf{24}(6),  381--395 (1981)

\bibitem{grimson1990object}
Grimson, W.E.L., Huttenlocher, D.P., et~al.: Object recognition by computer:
  the role of geometric constraints. Mit Press (1990)

\bibitem{grunert1841pothenotische}
Grunert, J.A.: Das pothenotische problem in erweiterter gestalt nebst bber
  seine anwendungen in der geodasie. Grunerts Archiv fur Mathematik und Physik
  pp. 238--248 (1841)

\bibitem{hartley2003multiple}
Hartley, R., Zisserman, A.: Multiple view geometry in computer vision.
  Cambridge university press (2003)

\bibitem{he2016deep}
He, K., Zhang, X., Ren, S., Sun, J.: Deep residual learning for image
  recognition. In: Proceedings of the IEEE conference on computer vision and
  pattern recognition. pp. 770--778 (2016)

\bibitem{hu2018cvm}
Hu, S., Feng, M., Nguyen, R.M., Hee~Lee, G.: Cvm-net: Cross-view matching
  network for image-based ground-to-aerial geo-localization. In: Proceedings of
  the IEEE Conference on Computer Vision and Pattern Recognition. pp.
  7258--7267 (2018)

\bibitem{kendall2017geometric}
Kendall, A., Cipolla, R.: Geometric loss functions for camera pose regression
  with deep learning. In: Proceedings of the IEEE Conference on Computer Vision
  and Pattern Recognition. pp. 5974--5983 (2017)

\bibitem{kendall2015posenet}
Kendall, A., Grimes, M., Cipolla, R.: Posenet: A convolutional network for
  real-time 6-dof camera relocalization. In: Proceedings of the IEEE
  international conference on computer vision. pp. 2938--2946 (2015)

\bibitem{kingma2014adam}
Kingma, D.P., Ba, J.: Adam: A method for stochastic optimization. In:
  International Conference on Learning Representations (ICLR) (2015)

\bibitem{kneip2011novel}
Kneip, L., Scaramuzza, D., Siegwart, R.: A novel parametrization of the
  perspective-three-point problem for a direct computation of absolute camera
  position and orientation. In: CVPR 2011. pp. 2969--2976. IEEE (2011)

\bibitem{landautomatic}
Land, A.H., Doig, A.G.: An automatic method for solving discrete programming
  problems. In: 50 Years of Integer Programming 1958-2008, pp. 105--132.
  Springer (2010)

\bibitem{lepetit2009epnp}
Lepetit, V., Moreno-Noguer, F., Fua, P.: Epnp: An accurate o (n) solution to
  the pnp problem. International journal of computer vision  \textbf{81}(2),
  ~155 (2009)

\bibitem{MegaDepthLi18}
Li, Z., Snavely, N.: Megadepth: Learning single-view depth prediction from
  internet photos. In: Computer Vision and Pattern Recognition (CVPR) (2018)

\bibitem{marshall1968scaling}
Marshall, A.W., Olkin, I.: Scaling of matrices to achieve specified row and
  column sums. Numerische Mathematik  \textbf{12}(1),  83--90 (1968)

\bibitem{yi2018learning}
Moo~Yi, K., Trulls, E., Ono, Y., Lepetit, V., Salzmann, M., Fua, P.: Learning
  to find good correspondences. In: Proceedings of the IEEE Conference on
  Computer Vision and Pattern Recognition. pp. 2666--2674 (2018)

\bibitem{more1978levenberg}
Mor{\'e}, J.J.: The levenberg-marquardt algorithm: implementation and theory.
  In: Numerical analysis, pp. 105--116. Springer (1978)

\bibitem{moreno2008pose}
Moreno-Noguer, F., Lepetit, V., Fua, P.: Pose priors for simultaneously solving
  alignment and correspondence. In: European Conference on Computer Vision. pp.
  405--418. Springer (2008)

\bibitem{Silberman:ECCV12}
Nathan~Silberman, Derek~Hoiem, P.K., Fergus, R.: Indoor segmentation and
  support inference from rgbd images. In: ECCV (2012)

\bibitem{qi2017pointnet}
Qi, C.R., Su, H., Mo, K., Guibas, L.J.: Pointnet: Deep learning on point sets
  for 3d classification and segmentation. In: Proceedings of the IEEE
  Conference on Computer Vision and Pattern Recognition. pp. 652--660 (2017)

\bibitem{qi2017pointnetplusplus}
Qi, C.R., Yi, L., Su, H., Guibas, L.J.: Pointnet++: Deep hierarchical feature
  learning on point sets in a metric space. arXiv preprint arXiv:1706.02413
  (2017)

\bibitem{schoenberger2016sfm}
Sch\"{o}nberger, J.L., Frahm, J.M.: Structure-from-motion revisited. In:
  Conference on Computer Vision and Pattern Recognition (CVPR) (2016)

\bibitem{sinkhorn1964relationship}
Sinkhorn, R.: A relationship between arbitrary positive matrices and doubly
  stochastic matrices. The annals of mathematical statistics  \textbf{35}(2),
  876--879 (1964)

\bibitem{villani2009optimal}
Villani, C.: Optimal transport--old and new, volume 338 of a series of
  comprehensive studies in mathematics (2009)

\bibitem{wang2018dynamic}
Wang, Y., Sun, Y., Liu, Z., Sarma, S.E., Bronstein, M.M., Solomon, J.M.:
  Dynamic graph cnn for learning on point clouds. arXiv preprint
  arXiv:1801.07829  (2018)

\bibitem{wu20153d}
Wu, Z., Song, S., Khosla, A., Yu, F., Zhang, L., Tang, X., Xiao, J.: 3d
  shapenets: A deep representation for volumetric shapes. In: Proceedings of
  the IEEE conference on computer vision and pattern recognition. pp.
  1912--1920 (2015)

\bibitem{zheng2013revisiting}
Zheng, Y., Kuang, Y., Sugimoto, S., Astrom, K., Okutomi, M.: Revisiting the pnp
  problem: A fast, general and optimal solution. In: Proceedings of the IEEE
  International Conference on Computer Vision. pp. 2344--2351 (2013)

\end{thebibliography}

\end{document}